\documentclass{article} 
\usepackage{iclr2027_conference,times}

\usepackage{amsmath,amsfonts,bm}

\def\eqref#1{equation~\ref{#1}}

\def\1{\bm{1}}

\DeclareMathAlphabet{\mathsfit}{\encodingdefault}{\sfdefault}{m}{sl}
\SetMathAlphabet{\mathsfit}{bold}{\encodingdefault}{\sfdefault}{bx}{n}

\newcommand{\KL}{D_{\mathrm{KL}}}

\usepackage[hidelinks]{hyperref}
\usepackage{url}

\title{Neighborhood Smoothing for Calibration}

\author{Idan Horowitz \& Avigdor Gal \\
Faculty of Data and Decision Sciences\\
Technion\\
Haifa, 3200003, IL \\
\texttt{idanhorowitz@campus.technion.ac.il,
        avigal@technion.ac.il} \\
}

\newtheorem{theorem}{Theorem}
\newtheorem{lemma}{Lemma}
\newtheorem{proposition}{Proposition}
\newtheorem{corollary}{Corollary}
\newtheorem{definition}{Definition}

\newcommand{\model}{NeSCal }
\newcommand{\modelNoSpace}{NeSCal}

\usepackage{amssymb}
\usepackage{booktabs}
\usepackage{multirow}
\usepackage{xcolor}
\usepackage{algorithm}
\usepackage{algpseudocode}
\usepackage{graphicx}  
\providecommand{\nescalResultCell}[2]{\mbox{#1{\tiny\,$\pm$\,#2}}}
\newcommand{\nrc}{\nescalResultCell}

\definecolor{zeroth}{RGB}{0,0,180}
\definecolor{first}{RGB}{180,0,0}
\definecolor{second}{RGB}{0,120,0}
\definecolor{higher}{RGB}{120,120,120}

\iclrfinalcopy
\begin{document}

\maketitle
\fancyhead{}
\renewcommand{\headrulewidth}{0pt}

\begin{abstract}
Modern neural networks are often miscalibrated, with a tendency to overconfidence. Existing train-time calibration methods largely modify task losses or calibration penalties, leaving neighborhood structure in learned representations underexploited. We introduce graph smoothing as a general principle for train-time calibration, which encourages similar predictive distributions across neighboring samples in representation space. We analyze the effects of graph smoothing, deriving bounds that connect predictive divergence between neighboring samples to local confidence variation and to the propagation of pointwise calibration error, and characterize the conditions under which smoothing can or cannot improve calibration.  In light of this analysis, we propose \modelNoSpace, a graph-based train-time regularizer that penalizes the Jensen--Shannon divergence between predictive distributions of neighboring samples. We present a thorough empirical analysis, showing that across standard calibration benchmarks, \model improves predictive quality, and the improvement is complementary to post-hoc calibration: after temperature scaling, \model attains the lowest NLL of all evaluated train-time methods in seven of the eight image and tabular settings. These findings demonstrate the value of graph smoothing over learned representations for neural network calibration. 
\end{abstract}

\section{Introduction}\label{sec:intro}

Confidence calibration in classification refers to the alignment between the empirical correctness of a model's predictions and its confidence in these predictions. A well-calibrated confidence serves as a reliable indication of prediction correctness and an essential component of downstream decisions.
Modern deep neural networks tend to be over-confident~\citep{Guo_2107_calibration}. Approaches to addressing miscalibration are typically either post-hoc, learning a function on a withheld data set to reshape confidence values~\citep{Guo_2107_calibration, kull_dirichlet_cwece, platt__posthoc_platt_scaling, zadrozny_posthoc_isotonic_regression},  or train-time, modifying the learning objective to encourage the model to be better calibrated, either by some explicit, calibration-aware penalty term~\citep{Hebbalaguppe_mdca, karandikar_sAVuC_sbECE, krishnan_avuc, Kumar_2018_MMCE}, or implicitly, through a loss shown to improve calibration~\citep{mukhoti_focal, muller_label_smoothing}. 

Existing train-time calibration methods largely modify the learning objective based primarily on model outputs and labels, without explicitly exploiting the geometry of the representations learned by the model. Yet these representations encode information about which samples the model regards as similar. Nearby samples in representation space should, when they correspond to similar underlying inputs, receive similar predictions, providing a natural structure over which predictive behavior can be regularized. 

In this work, we investigate model regularization to encourage adjacent samples in a neighborhood graph over its representation space to receive similar predictive distributions, a principle we term \emph{graph smoothing}. Penalizing the variation of the learned function in such a graph is the basis of manifold regularization~\citep{Belkin_2006_manifold_learning}, where it has been used to improve classification accuracy under limited supervision. Applied to calibration, suppressing predictive disagreement constrains local confidence variation: a well-calibrated prediction can constrain neighboring calibration errors when the connected samples have similar true probabilities of correctness. By exploiting this additional structure during training, graph smoothing can also improve the learned predictor in ways that remain beneficial when combined with subsequent post-hoc recalibration. 

\begin{figure}[!t]
\vspace{-3.75em}
\centering
\includegraphics[width=\linewidth,trim={0 40 0 40}, clip]{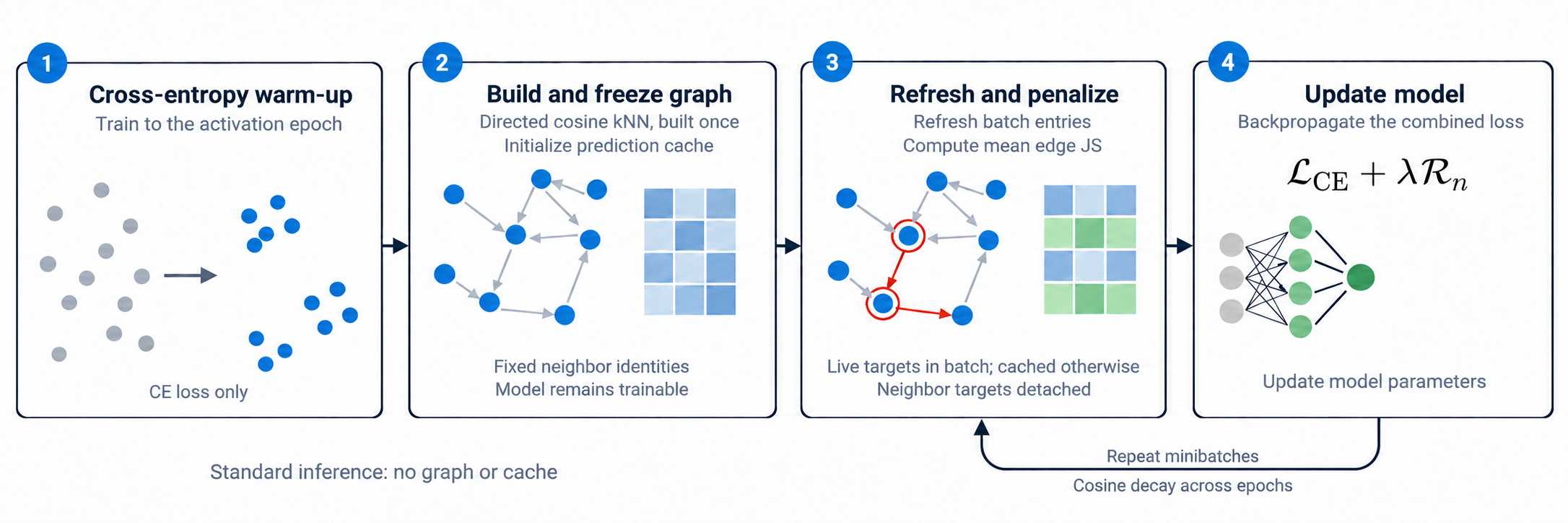}
\setlength{\abovecaptionskip}{0pt}
\setlength{\belowcaptionskip}{0pt}
\caption{Overview of \model training.}
\label{fig:nescal-overview}
\vspace{-3.5em}
\end{figure}

We instantiate graph smoothing as \model (\textbf{Ne}ighborhood \textbf{S}moothing for \textbf{Cal}ibration), a train-time regularizer that penalizes Jensen--Shannon (JS) divergence between neighboring predictive distributions, added to a task loss through a single weighting term at no inference-time cost.

The paper's contributions are quintuple. (i) We introduce \emph{graph smoothing} as a general principle for train-time calibration, in which a model is regularized to encourage similar predictive distributions for samples adjacent in a neighborhood graph over representation space (Section~\ref{sec:graph-smoothing}). (ii) We demonstrate the relationship between graph-based predictive smoothness and calibration, showing that the variation of the predictive distribution translates into a bound on the propagation of pointwise calibration error along paths in the graph (Proposition~\ref{cor:propagation}; Theorem~\ref{thm:pathwise-propagation}). (iii) We characterize when graph smoothing improves calibration and when it cannot, identifying the variation of true correctness probability across an edge as the quantity governing both (Propositions~\ref{prop:obstruction} and~\ref{prop:threshold}). (iv) We instantiate the principle as a train-time regularizer \modelNoSpace, penalizing the Jensen--Shannon divergence between the predictive distributions of neighbors in a $k$-nearest-neighbor graph~\ref{sec:nescal}. (v) We show empirically that \model improves negative log-likelihood (NLL) in all eleven main benchmarks and ranks first or second among the evaluated methods in seven of eight image and tabular settings. After temperature scaling, it attains the lowest NLL in seven of those eight settings (Section~\ref{sec:results}).

\section{Background} \label{sec:background}

Consider a generic $K$-class classification problem, with a model $f$ that receives inputs $x \in X$ and produces softmax probability vectors $p(x) := f(x) \in \Delta^{K-1}$, the $(K{-}1)$-dimensional probability simplex.  Write $p_i:=p(x_i)$ and let $p_{i,k}$ denote the $i$th sample's class-$k$ probability. Each input $x$ has an associated true label $y \in \{1,\ldots,K\}$, with $Y$ the corresponding label random variable. The model's predicted label and its associated confidence for $x$ are $\hat{y}(x) = \arg\max_k p_{k}(x)$ and $c(x) = \max_k p_{k}(x)$.

We write the true conditional class-probability vector (the Bayes posterior) $\eta(x) := \big(P(Y=1\mid X=x),\ldots,P(Y=K\mid X=x)\big)\in\Delta^{K-1}$. For the predicted class, the conditional probability of correctness is $a(x) := P(Y=\hat{y}(x)\mid X=x) = \eta_{\hat{y}(x)}(x)$, a single coordinate of $\eta(x)$. Thus, $c(x)$ is the model's predicted probability of being correct, while $a(x)$ is the corresponding true conditional probability of correctness. We define $\mathcal{E}_{\mathrm{cal}}(x) := |a(x) - c(x)|$ as the pointwise calibration error at $x$, writing $\mathcal{E}_{\mathrm{cal}_i} := \mathcal{E}_{\mathrm{cal}}(x_i)$ for samples. Perfect pointwise calibration is then achieved if $c(x) = a(x)$ for all $x \in \mathcal{X}$. The posterior $\eta(x)$ is a property of the data distribution and $a(x)$ also depends on the model's predicted label; neither is directly observed from a single labeled example, so empirical calibration metrics assess aggregated notions of calibration. The most widely used metric is Expected Calibration Error (ECE)~\cite{Naeini_2015_ECE}, which measures \emph{confidence calibration}~\citep{Guo_2107_calibration} by binning predictions according to confidence and comparing confidence means with empirical accuracy within each bin. ECE is a binning-based estimator, it is thus sensitive to the binning scheme and to finite-sample bias~\citep{Nixon_2019_measuring_calibration, roelos_2022_ECEBias, zhang_2020_mixnmatch_desiderata}. Moreover, since ECE aggregates across predictions, pointwise over- and under-confidence can cancel. A low ECE therefore does not imply that predictions are calibrated at a specific point. A discussion of the alternative metrics can be found in Appendix~\ref{app:metrics}.

\section{Graph Smoothing for Confidence Calibration} \label{sec:graph-smoothing}

\subsection{Problem Setup} \label{sec:gs:problem-def}

We adopt the setting of Section~\ref{sec:background}, with a dataset $\mathcal{D} = \{(x_i,y_i)\}_{i=1}^{n}$ of i.i.d.\ draws from $P(X,Y)$ with $X\subseteq\mathbb{R}^{d}$, and a probabilistic classifier $f_\theta : X \rightarrow \Delta^{K-1}$ with softmax output layer, trained to minimize $\mathbb{E}_{(X,Y)\sim P}[\mathcal{L}(f_\theta(X),Y)]$ for a task loss $\mathcal{L}$. Write $p_i := f_\theta(x_i)$, $a_i := a(x_i)$ and $c_i := c(x_i)$, so that $\mathcal{E}_{\mathrm{cal}_i} = |a_i - c_i|$. Our analysis targets pointwise calibration: the top-label guarantee is required at each point rather than on average, without necessitating the full predictive vector to match the posterior.

Let $h:\mathcal X\to\mathcal Z\subseteq\mathbb R^l$ produce representations $z_i=h(x_i)$. A neighborhood graph $\mathcal G=(V,E)$ has  nodes $V = \{z_i\}_{i=1}^n \subset \mathcal{Z}$ and connects nearby representations under a chosen distance. Its adjacency matrix $W$ assigns nonnegative weights $W_{ij}$ to edges $(i,j) \in E$. Let $\mathrm{D}(p_i \Vert p_j)$ be a nonnegative divergence between the predictive probability distributions at observations $i$ and $j$.

We define $R_n(p) := \sum_{i,j} W_{ij}\,\mathrm{D}(p_i \Vert  p_j)$ as the \emph{neighborhood consistency regularizer} and consider the penalized formulation
\begin{equation} \label{eq:opt-lagrangian}
  \min_{f_\theta}\;
  \mathcal{L}(\theta) = \mathcal{L}_{\mathrm{task}}(\theta) + \lambda\, R_n(p),
\end{equation}
where $\mathcal{L}_{\mathrm{task}}$ is the task loss, and $\lambda > 0$ controls the regularization strength.

\subsection{Formal Guarantees}\label{sec:gs:theory}

We relate predictive disagreement to confidence variation and pointwise calibration-error propagation on the training graph, at a fixed predictor. Held-out calibration is assessed empirically in Section~\ref{sec:results}. Proofs and the local quadratic interpretation of the regularizer appear in Appendix~\ref{app:gen-proofs}. We make the standing requirement that $\mathrm{D}(p\Vert q)\geq C_{\mathrm{D}}\|p-q\|_1^2$ for all $p,q\in\Delta^{K-1}$, with $C_{\mathrm{D}}>0$. This is a Pinsker-type condition that implies $|c_i-c_j|\leq\tfrac12\|p_i-p_j\|_1\leq\frac{1}{2\sqrt{C_{\mathrm{D}}}}\sqrt{\mathrm{D}(p_i\Vert p_j)}$, which bounds the propagation of pointwise calibration error:

\begin{proposition}[Calibration Propagation]
\label{cor:propagation}
Under the standing domination condition, for any edge $(i,j) \in E$,
\begin{equation}
  \mathcal{E}_{{\mathrm{cal}}_j} \;\leq\; \ \mathcal{E}_{{\mathrm{cal}}_i} + \frac{1}{2\sqrt{C_{\mathrm{D}}}}\sqrt{\mathrm{D}(p_i \Vert p_j)} + |a_i - a_j|
\end{equation}
\end{proposition}

The bound decomposes propagation into three terms: the calibration error already present at $i$, a smoothness term controlled directly by the regularizer through $\mathrm{D}(p_i \Vert p_j)$, and the local variation $|a_i - a_j|$ of the true correctness probability. Only the second is directly penalized. With a fixed graph, the correctness gap depends on $\theta$ only through the predicted labels. A calibrated prediction therefore constrains neighboring errors when connected points have similar correctness probabilities. The bound, however, does not detail how a calibrated prediction arises in the first place. This is left to the task loss. 

Theorem~\ref{thm:pathwise-propagation} extends Proposition~\ref{cor:propagation} from a single edge to paths through the graph: 

\begin{theorem}[Pathwise Calibration Propagation]
\label{thm:pathwise-propagation}
Under the standing domination condition, let $P=(v_0,v_1,\ldots,v_r)$ be a simple directed path in $\mathcal{G}$ from $v_0=s$ to $v_r=v$. Write $ D_\ell :=\mathrm{D}(p_{v_{\ell-1}} \Vert p_{v_\ell})$.
Then
\begin{equation}
\label{eq:pathwise-propagation}
    \mathcal{E}_{\mathrm{cal}_v} \leq \mathcal{E}_{\mathrm{cal}_s} +\sqrt{\frac{r}{4C_{\mathrm{D}}} \sum_{\ell=1}^{r}D_\ell} +|a_v-a_s|
\end{equation}
\end{theorem}

A well-calibrated prediction transfers to points connected by short paths along which both the predictive divergence and the variation of the true correctness probability remain small. The bound grows with path length and thus the  guarantee is most informative locally. Additional disagreement bounds appear in Appendix~\ref{app:gen-proofs} (Propositions~\ref{cor:confidence-smoothness} and~\ref{lem:markov-bound}, Corollary~\ref{cor:prob-confidence-smoothness}).

\subsection{When Smoothing Improves Calibration} \label{sec:gs:obstruction}

Section~\ref{sec:gs:theory} bounds the propagation of calibration error, given that it is small somewhere.  We next identify an obstruction to calibrating both endpoints of an edge and a condition under which idealized pairwise averaging reduces their summed expected squared calibration error.

\begin{proposition}[Smoothing Obstruction]
\label{prop:obstruction}
Under the standing domination condition, for every edge $(i,j)\in E$,
\begin{equation}
\label{eq:obstruction-D}
  \mathcal{E}_{\mathrm{cal}_i}+ \mathcal{E}_{\mathrm{cal}_j} \geq |a_i - a_j| - \frac{1}{2\sqrt{C_{\mathrm{D}}}}\sqrt{\mathrm{D}(p_i \Vert p_j)}
\end{equation}
\end{proposition}

For fixed predicted labels, reducing predictive disagreement raises the lower bound toward the correctness gap $|a_i-a_j|$. A rising floor, however, does not establish that the realized error rose, as the error may already have exceeded it. Proposition~\ref{prop:obstruction} therefore identifies the regime in which smoothing is provably obstructed from calibrating both endpoints. The next result characterizes when averaging reduces expected squared error.

To analyze the error removed by averaging, we fix an edge and randomize over training. Let $\mathcal T$ denote randomness from initialization, data order, and stochastic optimization. We analyze idealized pairwise averaging on a single edge, in which both endpoints adopt the mixture $m = \tfrac12(p_i+p_j)$. Both samples must predict the same fixed label almost surely under $\mathcal{T}$, so that $a_i$ and $a_j$ are fixed targets, and then averaging the predictive distributions is equivalent to averaging the confidences, $\max_k m_k = \tfrac12(c_i + c_j)$. The regularizer has access to neither $a$ nor the labels; it only draws predictions toward one another. The question is when doing so reduces the summed expected squared calibration error.

\begin{proposition}[Smoothing threshold]
\label{prop:threshold}
Let $(i,j)$ be an edge with $\hat{y}_i = \hat{y}_j = k^\ast$ almost surely under $\mathcal{T}$, and write $e_i := c_i - a_i$, $e_j := c_j - a_j$ for the pointwise errors. Replacing both confidences by their average strictly reduces the summed expected squared calibration error of the edge iff
\begin{equation}
\label{eq:threshold}
  (a_i - a_j)^2 \;<\; \mathbb{E}_{\mathcal{T}}\bigl[(e_i - e_j)^2\bigr] .
\end{equation}
\end{proposition}

A bias--variance decomposition and correlation-dependent specialization appear in Appendix~\ref{app:gs:threshold}. Only the difference $e_i-e_j$ enters~\eqref{eq:threshold}. Averaging can reduce error that differs between neighbors, whether random or systematic, but cannot touch the component they share, and systematic overconfidence is precisely the error deep classifiers are known to exhibit~\citep{Guo_2107_calibration}. This motivates testing whether gains from neighborhood smoothing remain useful after a global confidence adjustment. Section~\ref{sec:res-ts} evaluates this empirically by applying the same post-hoc calibrators to every train-time method.

\section{\modelNoSpace} \label{sec:nescal}

\subsection{The Regularizer}\label{sec:nescal:reg}\label{sec:nescal:js}
\model instantiates the graph-smoothing objective of Equation~\ref{eq:opt-lagrangian} with the Jensen--Shannon divergence and a $k$-nearest-neighbor graph over learned representations:

\begin{equation}
  \min_{f_\theta}\;
  \mathcal{L}(\theta) = \mathcal{L}_{\mathrm{CE}}(\theta) + \lambda\, \sum_{i,j} W_{ij}\,\mathrm{JS}(p_i\Vert p_j)
\end{equation}

where $\mathcal{L}_{\mathrm{CE}}$ is the cross-entropy (CE) loss and $\mathrm{JS}(p_i\Vert p_j)$ is the Jensen--Shannon divergence~\citep{Endres_2003_JSD, Fuglede_2004_JS_distance_metric}, $\mathrm{JS}(p_i\Vert p_j)=\tfrac12\KL(p_i\Vert m)+\tfrac12\KL(p_j\Vert m)$, with $m=\tfrac12(p_i+p_j)$ and $\KL(u\Vert v)=\sum_k u_k\log(u_k/v_k)$. We use base-$2$ logarithms in the JS divergence, so that $\mathrm{JS}(p_i\Vert p_j)\in[0,1]$. Cross-entropy and reported NLL use natural logarithms.

The adjacency matrix $W \in \{0,1\}^{n\times n}$ encodes a $k$-nearest-neighbor graph (kNN) in the representation space $\mathcal{Z}$, with $W_{ij} = 1 \iff j \in \mathrm{kNN}(i)$. For optimization, we average the penalty over the directed edge set $E=\{(i,j):W_{ij}=1\}$, using $\bar R_n(p)=R_n(p)/|E|$. 

JS divergence satisfies the total-variation domination condition with $C_{\mathrm{JS}}=1/(8\ln 2)$, allowing the general confidence and propagation bounds to apply. Its symmetry and bounded range make it a sufficient and practical (rather than uniquely optimal) choice for comparing neighboring predictive distributions, while comparison of the full vectors captures disagreement across all classes. The induced local quadratic geometry is described in Appendix~\ref{app:specific-proofs}.

\subsection{Design Choices} \label{sec:nescal:design}

\paragraph{Neighborhood Graph.}Neighbors are identified by cosine distance in $\mathcal{Z}$. The graph is constructed once at activation and held fixed, limiting the construction cost and keeping the neighborhoods unchanged during regularized training. We use $k=10$ by default. Section~\ref{sec:res-design} examines sensitivity to this choice.

\paragraph{Activation and Schedule.}The activation epoch is a hyperparameter, swept jointly with $\lambda$ on validation NLL. Proposition~\ref{prop:obstruction} motivates delaying graph construction until the representation matures and neighborhoods become semantically meaningful. If the across-run dispersion of neighboring errors shrinks as training converges, the tolerable correctness gap narrows (Proposition~\ref{prop:threshold}), motivating a reduction in regularization strength. \model accordingly activates the regularizer after an initial warm-up and decays $\lambda$ over the remainder of training on a cosine schedule. Further discussion appears in Appendix~\ref{app:nescal:activation}.

\paragraph{Task Loss.}\model uses cross-entropy as its task loss. Cross-entropy is a strictly proper scoring rule: its expected loss is uniquely minimized at the true conditional distribution. In contrast, the regularizer $R_n$ is minimized whenever neighboring predictions agree, regardless of their correctness. Train-time calibration methods are frequently combined with label smoothing~\cite{Szegedy_2016_labelsmoothing} or focal loss~\cite{Lin_2017_focalloss}, neither of which is a proper scoring rule since population minimizers generally differ from the true conditional distribution. Replacing cross-entropy with either would therefore remove the strictly proper scoring rule that anchors the supervised term to that distribution. The smoothing penalty, which does not itself identify the correct posterior, would then be susceptible to propagating a shared bias. We therefore retain a strictly proper task loss.

\subsection{Implementation}\label{sec:nescal:complexity}
Figure~\ref{fig:nescal-overview} presents an overview of the \model training process. Training uses minibatch updates with detached neighbor targets: current predictions for neighbors in the minibatch and cached predictions otherwise. Inference is unchanged. Algorithm~\ref{alg:nescal} and the full implementation and complexity analysis appear in Appendix~\ref{app:nescal:complexity}, while measured costs are reported in Appendix~\ref{app:cost}.

\section{Experimental Setup} \label{sec:exp-design}

We compare \model with train-time calibration methods across image, text, and tabular classification, evaluating each trained model both directly and after post-hoc recalibration. This section states the protocol; full implementation details are deferred to Appendix~\ref{app:train-details}.

\subsection{Datasets and Architectures}

\textbf{Image.} CIFAR-10 and CIFAR-100~\citep{krizhevsky_2009_cifar} with ResNet-50~\citep{he_2016_resnet} and WideResNet-28-10~\citep{zagoruyko_2016_wrn}, and Tiny-ImageNet~\citep{le_2015_tinyimagenet} with ResNet-50.

\textbf{Text.} 20 Newsgroups~\citep{lang_1995_20ng} (20NG) with the global-pooling CNN (GPCNN)~\citep{lin_2014_gpcnn}, and Amazon Reviews~\citep{hou_2024_amazonreviews} (binary and five-class) with a multilayer perceptron (MLP) head on fine-tuned and then frozen RoBERTa-base~\citep{liu_2019_roberta} representations. In the frozen-encoder setting, \model regularizes only the head with the encoder representation remaining fixed during training.

\textbf{Tabular.} Helena, Jannis~\citep{Guyon_2019_helena} and a subsample of Covertype~\citep{blackard_covertype_31}, with an MLP classifier.

\subsection{Compared Methods}

We compare with CE and six train-time calibration methods: focal loss (FL)~\citep{mukhoti_focal, Lin_2017_focalloss}, label smoothing (LS)~\citep{muller_label_smoothing, Szegedy_2016_labelsmoothing}, and margin-based label smoothing (MbLS)~\citep{Liu_2022_mbls}, which modify the training objective; and MMCE~\citep{Kumar_2018_MMCE}, MDCA~\citep{Hebbalaguppe_mdca}, and s-AvUC~\citep{karandikar_sAVuC_sbECE}, which introduce explicit calibration-related penalties. Every method is tuned under the protocol below. The main comparison uses \model with cosine decay; the constant schedule is examined separately.

To assess composition, we apply five post-hoc calibrators to every train-time method. Temperature scaling (TS)~\citep{Guo_2107_calibration} and Dirichlet calibration~\citep{kull_dirichlet_cwece} provide established output-based adjustments, ranging from a single temperature to a multiclass calibration map. LoCal Nets (LoCal)~\citep{barbera2026multiclass}, LoRe~\citep{luo_lce}, and DAC~\citep{Tomani_2023_DAC} use local or density information in representation space. Together, they test whether the gains from training persist under both global output recalibration and recalibration informed by representations.

\subsection{Model Selection and Evaluation}

All methods are trained for a fixed epoch budget, except the frozen-encoder heads, which use early stopping. Checkpoints and hyperparameters are selected by validation NLL, a strictly proper scoring rule for the full predictive distribution~\citep{Gneiting_2007_proper_scoring}. ECE measures agreement between confidence and empirical correctness after aggregation and can therefore be low when predictions are uninformative. This distinction matters when selecting regularization strength: strong smoothing can reduce ECE while degrading accuracy and NLL. Selection by a proper score remains sensitive to this loss of predictive information, although it can select a configuration with higher ECE. The shared criterion gives every method a common selection target, although this may differ from its original protocol. Appendix~\ref{app:selection} examines the resulting trade-offs. 

The primary evaluation metric is also NLL, for the reasons stated above. We additionally report the multiclass Brier score~\citep{brier_1950_brier}, accuracy and ECE, alongside secondary calibration metrics and the calibration--refinement decomposition~\citep{kull_2015_brier_decomposition} in Appendix~\ref{app:metrics}. We assess consistency across settings rather than significance within them (Appendix~\ref{app:tests}).

\section{Experimental Results}\label{sec:results}
We address three questions: whether \model improves calibration and predictive quality relative to CE and the competing train-time methods (Section~\ref{sec:res-quality}); whether these gains persist after post-hoc recalibration (Section~\ref{sec:res-ts}); and whether they depend on the neighborhood structure the theory relies on, and in which settings they arise (Sections~\ref{sec:res-design} and~\ref{sec:res-regime}).
Tables report test-set means, with standard deviations where shown. Unless a caption states otherwise, bold and italic identify the best and second-best distinct means, with ties at displayed precision sharing emphasis. \model denotes the cosine schedule on real-data benchmarks; synthetic results use the constant schedule. R50 and WRN denote ResNet-50 and WideResNet-28-10, respectively.

\subsection{Predictive quality and calibration}\label{sec:res-quality}

\model improves NLL over CE in all eleven main benchmark settings and Brier score in ten (Table~\ref{tab:res-main}; Appendix Table~\ref{tab:app-brier-acc}). These gains generally accompany improved accuracy, which increases in nine settings, including seven of the eight image and tabular settings .

Against the other train-time methods, \model ranks first or second (including ties) on NLL in seven of the eight image and tabular settings. Because NLL rewards both calibration and accuracy, we assess calibration improvement by asking whether any method is better on both at once. A method is Pareto-optimal if no other method is at least as accurate and has no higher ECE, with strict improvement in at least one. \model is Pareto-optimal in five of the eight image and tabular settings both before and after TS. Base-model ECE is where \model is least competitive; Section~\ref{sec:res-ts} shows that this ranking changes once confidence scale is corrected. On the text benchmarks, \modelNoSpace's improvements over CE are smaller and do not yield the lowest NLL against other methods. Section~\ref{sec:res-regime} characterizes this variation across settings.

\begin{table}[!htbp]
\centering
\caption{Test NLL before and after validation-fitted temperature scaling; lower is better. Values are means $\pm$ SD over seeds, with seed counts specified in Appendix~\ref{app:train-details}.}
\label{tab:res-main}
\begingroup
\scriptsize\setlength{\tabcolsep}{1.6pt}\renewcommand{\arraystretch}{1.03}
\resizebox{\linewidth}{!}{%
\begin{tabular}{@{}lcccccccc@{}}
\toprule
Setting & CE & FL & LS & MMCE & MDCA & s-AvUC & MbLS & NeSCal \\
\midrule
\multicolumn{9}{@{}l}{\emph{Base model}} \\
CIFAR-10 / R50
& \nescalResultCell{0.2065}{0.0091}
& \textbf{\nescalResultCell{0.1821}{0.0045}}
& \nescalResultCell{0.2285}{0.0064}
& \nescalResultCell{0.2031}{0.0043}
& \nescalResultCell{0.2089}{0.0074}
& \nescalResultCell{0.2065}{0.0052}
& \nescalResultCell{0.2227}{0.0052}
& \textit{\nescalResultCell{0.1843}{0.0058}} \\
CIFAR-10 / WRN
& \nescalResultCell{0.1972}{0.0077}
& \textbf{\nescalResultCell{0.1572}{0.0033}}
& \nescalResultCell{0.1838}{0.0034}
& \nescalResultCell{0.1791}{0.0083}
& \nescalResultCell{0.1951}{0.0071}
& \nescalResultCell{0.2090}{0.0102}
& \nescalResultCell{0.1791}{0.0025}
& \textit{\nescalResultCell{0.1752}{0.0028}} \\
CIFAR-100 / R50
& \nescalResultCell{0.8782}{0.0249}
& \textit{\nescalResultCell{0.8463}{0.0044}}
& \nescalResultCell{0.8966}{0.0186}
& \nescalResultCell{0.8935}{0.0393}
& \nescalResultCell{0.8889}{0.0139}
& \nescalResultCell{0.8763}{0.0170}
& \nescalResultCell{0.8614}{0.0174}
& \textbf{\nescalResultCell{0.8026}{0.0155}} \\
CIFAR-100 / WRN
& \nescalResultCell{0.8785}{0.0049}
& \textbf{\nescalResultCell{0.7681}{0.0058}}
& \nescalResultCell{0.8388}{0.0086}
& \nescalResultCell{0.8838}{0.0219}
& \nescalResultCell{0.8786}{0.0139}
& \nescalResultCell{0.8691}{0.0043}
& \nescalResultCell{0.8206}{0.0117}
& \textit{\nescalResultCell{0.7760}{0.0173}} \\
Tiny-ImageNet
& \nrc{1.4704}{0.0254} & \nrc{1.4960}{0.0101} & \nrc{1.4816}{0.0254} & \nrc{1.4797}{0.0344}
& \nrc{1.4406}{0.0225} & \nrc{1.4403}{0.0317} & \textit{\nrc{1.4287}{0.0251}} & \textbf{\nrc{1.4126}{0.0254}} \\
\addlinespace[2pt]
Helena
& \nescalResultCell{2.5651}{0.0034}
& \nescalResultCell{2.5909}{0.0032}
& \textit{\nescalResultCell{2.5504}{0.0075}}
& \nescalResultCell{2.5648}{0.0030}
& \nescalResultCell{2.5679}{0.0036}
& \nescalResultCell{2.5653}{0.0017}
& \textbf{\nescalResultCell{2.5500}{0.0030}}
& \nescalResultCell{2.5534}{0.0042} \\
Jannis
& \nescalResultCell{0.7125}{0.0001}
& \nescalResultCell{0.7955}{0.0055}
& \nescalResultCell{0.7168}{0.0038}
& \nescalResultCell{0.7127}{0.0029}
& \textit{\nescalResultCell{0.7121}{0.0019}}
& \nescalResultCell{0.7141}{0.0033}
& \nescalResultCell{0.7125}{0.0026}
& \textbf{\nescalResultCell{0.7066}{0.0014}} \\
Covertype
& \nescalResultCell{0.2516}{0.0033}
& \nescalResultCell{0.3647}{0.0052}
& \nescalResultCell{0.2771}{0.0035}
& \nescalResultCell{0.2505}{0.0016}
& \nescalResultCell{0.2527}{0.0021}
& \nescalResultCell{0.2507}{0.0043}
& \textbf{\nescalResultCell{0.2489}{0.0019}}
& \textit{\nescalResultCell{0.2499}{0.0015}} \\
\addlinespace[2pt]
20NG
& \nescalResultCell{0.8127}{0.0496}
& \nescalResultCell{0.7854}{0.0257}
& \textit{\nescalResultCell{0.6405}{0.0214}}
& \nescalResultCell{0.7648}{0.0251}
& \nescalResultCell{0.7799}{0.0188}
& \nescalResultCell{0.7789}{0.0195}
& \textbf{\nescalResultCell{0.6313}{0.0157}}
& \nescalResultCell{0.8023}{0.0369} \\
Amazon (5-class)
& \nescalResultCell{0.7809}{0.0029}
& \nescalResultCell{0.7948}{0.0006}
& \textit{\nescalResultCell{0.7779}{0.0027}}
& \textbf{\nescalResultCell{0.7725}{0.0028}}
& \nescalResultCell{0.7799}{0.0033}
& \nescalResultCell{0.7808}{0.0021}
& \nescalResultCell{0.7782}{0.0010}
& \nescalResultCell{0.7781}{0.0032} \\
Amazon (binary)
& \nescalResultCell{0.2254}{0.0012}
& \nescalResultCell{0.3023}{0.0091}
& \nescalResultCell{0.2234}{0.0014}
& \textbf{\nescalResultCell{0.2025}{0.0008}}
& \nescalResultCell{0.2255}{0.0053}
& \nescalResultCell{0.2241}{0.0020}
& \textit{\nescalResultCell{0.2233}{0.0015}}
& \nescalResultCell{0.2240}{0.0020} \\
\midrule
\multicolumn{9}{@{}l}{\emph{After temperature scaling}} \\
CIFAR-10 / R50
& \nescalResultCell{0.1863}{0.0084}
& \textit{\nescalResultCell{0.1813}{0.0044}}
& \nescalResultCell{0.2191}{0.0074}
& \nescalResultCell{0.2027}{0.0045}
& \nescalResultCell{0.1858}{0.0144}
& \nescalResultCell{0.1857}{0.0099}
& \nescalResultCell{0.2206}{0.0059}
& \textbf{\nescalResultCell{0.1698}{0.0046}} \\
CIFAR-10 / WRN
& \nescalResultCell{0.1575}{0.0083}
& \textit{\nescalResultCell{0.1568}{0.0033}}
& \nescalResultCell{0.1832}{0.0031}
& \nescalResultCell{0.1776}{0.0077}
& \nescalResultCell{0.1678}{0.0104}
& \nescalResultCell{0.1640}{0.0105}
& \nescalResultCell{0.1710}{0.0075}
& \textbf{\nescalResultCell{0.1566}{0.0052}} \\
CIFAR-100 / R50
& \nescalResultCell{0.8489}{0.0267}
& \textit{\nescalResultCell{0.8452}{0.0038}}
& \nescalResultCell{0.8951}{0.0188}
& \nescalResultCell{0.8631}{0.0402}
& \nescalResultCell{0.8577}{0.0137}
& \nescalResultCell{0.8478}{0.0137}
& \nescalResultCell{0.8558}{0.0168}
& \textbf{\nescalResultCell{0.7944}{0.0191}} \\
CIFAR-100 / WRN
& \nescalResultCell{0.8070}{0.0061}
& \textit{\nescalResultCell{0.7681}{0.0057}}
& \nescalResultCell{0.8374}{0.0076}
& \nescalResultCell{0.8309}{0.0151}
& \nescalResultCell{0.8141}{0.0155}
& \nescalResultCell{0.8212}{0.0099}
& \nescalResultCell{0.8173}{0.0103}
& \textbf{\nescalResultCell{0.7613}{0.0178}} \\
Tiny-ImageNet
& \nrc{1.4415}{0.0246} & \nrc{1.4950}{0.0099} & \nrc{1.4563}{0.0255} & \nrc{1.4614}{0.0319}
& \textit{\nrc{1.4096}{0.0228}} & \textbf{\nrc{1.4066}{0.0307}} & \nrc{1.4208}{0.0246} & \nrc{1.4098}{0.0255} \\
\addlinespace[2pt]
Helena
& \nescalResultCell{2.5618}{0.0015}
& \nescalResultCell{2.5907}{0.0035}
& \nescalResultCell{2.5503}{0.0072}
& \nescalResultCell{2.5615}{0.0036}
& \nescalResultCell{2.5653}{0.0025}
& \nescalResultCell{2.5624}{0.0027}
& \textit{\nescalResultCell{2.5498}{0.0033}}
& \textbf{\nescalResultCell{2.5497}{0.0049}} \\
Jannis
& \nescalResultCell{0.7115}{0.0008}
& \nescalResultCell{0.7582}{0.0097}
& \nescalResultCell{0.7160}{0.0039}
& \nescalResultCell{0.7111}{0.0025}
& \textit{\nescalResultCell{0.7104}{0.0011}}
& \nescalResultCell{0.7106}{0.0014}
& \nescalResultCell{0.7121}{0.0027}
& \textbf{\nescalResultCell{0.7042}{0.0011}} \\
Covertype
& \nescalResultCell{0.2478}{0.0028}
& \nescalResultCell{0.2930}{0.0026}
& \nescalResultCell{0.2589}{0.0037}
& \nescalResultCell{0.2463}{0.0011}
& \nescalResultCell{0.2490}{0.0015}
& \textit{\nescalResultCell{0.2453}{0.0033}}
& \nescalResultCell{0.2485}{0.0021}
& \textbf{\nescalResultCell{0.2449}{0.0012}} \\
\addlinespace[2pt]
20NG
& \nescalResultCell{0.7311}{0.0274}
& \nescalResultCell{0.7073}{0.0292}
& \textit{\nescalResultCell{0.6361}{0.0235}}
& \nescalResultCell{0.6971}{0.0200}
& \nescalResultCell{0.7339}{0.0374}
& \nescalResultCell{0.7209}{0.0281}
& \textbf{\nescalResultCell{0.6320}{0.0158}}
& \nescalResultCell{0.7290}{0.0231} \\
Amazon (5-class)
& \nescalResultCell{0.7676}{0.0044}
& \nescalResultCell{0.7694}{0.0012}
& \nescalResultCell{0.7779}{0.0020}
& \textbf{\nescalResultCell{0.7579}{0.0042}}
& \textit{\nescalResultCell{0.7593}{0.0037}}
& \nescalResultCell{0.7678}{0.0066}
& \nescalResultCell{0.7652}{0.0037}
& \nescalResultCell{0.7625}{0.0024} \\
Amazon (binary)
& \nescalResultCell{0.2207}{0.0061}
& \nescalResultCell{0.2113}{0.0071}
& \nescalResultCell{0.2205}{0.0017}
& \textbf{\nescalResultCell{0.2024}{0.0008}}
& \nescalResultCell{0.2235}{0.0038}
& \nescalResultCell{0.2237}{0.0017}
& \nescalResultCell{0.2214}{0.0029}
& \textit{\nescalResultCell{0.2102}{0.0019}} \\
\bottomrule
\end{tabular}}
\endgroup
\end{table}

\subsection{Composition Under Post-hoc Recalibration}\label{sec:res-ts}

An important assessment of a train-time calibration method is whether its gains survive the cheapest post-hoc correction. After the same validation-fitted temperature scaling is applied to every trained model, \model retains lower NLL than CE in all eleven settings and has the lowest NLL among the evaluated methods in seven of the eight image and tabular settings (Table~\ref{tab:res-main}). On CIFAR-10 and CIFAR-100 with ResNet-50, its NLL is respectively $6.3\%$ and $6.0\%$ below the strongest evaluated baseline after TS. TS compresses the spread of base-model ECE across methods on CIFAR from $2.2$--$6.3$ to $0.5$--$1.1$ percentage points, after which \model ranks first or second in five of the eight image and tabular settings (Table~\ref{tab:app-ece}). Much of the base-model ECE advantage of methods that lower confidence directly is therefore recoverable by a single temperature, whereas \model retains its NLL advantage after recalibration.

This pattern is consistent with the distinction motivated by Proposition~\ref{prop:threshold}: neighborhood smoothing can reduce variation in neighboring prediction errors, while temperature scaling adjusts confidence globally. On the two CIFAR settings with ResNet-50, \model reduces the reported proxy for neighboring-error dispersion and the across-seed spread of individual confidences (Appendix~\ref{app:theory-audit}). Failure-prediction AUROC and AURC also improve over cross-entropy in all five image settings (Appendix~\ref{app:discrimination}). These observations support a change in predictive quality beyond confidence scale, without requiring the two interventions to act on strictly independent error components.

The broader composition analysis supports the same conclusion: \model attains the lowest best-observed post-hoc NLL in six of eight tabular and image settings when each training method is evaluated with all five post-hoc calibrators (Appendix~\ref{app:posthoc}).

\subsection{Design choices, cost and robustness}\label{sec:res-design}

If \modelNoSpace's gains came from indiscriminate smoothing, a random graph of the same degree would reproduce them. It does not: kNN graphs yield lower NLL than random graphs in all three evaluated controls (Appendix~\ref{app:ablations}, Table~\ref{tab:graph-control}). This is consistent with the role of the correctness gap $|a_i-a_j|$ in Section~\ref{sec:graph-smoothing}: random edges connect points whose correctness probabilities differ, where enforcing agreement conflicts with calibration.

The validation-selected cosine configurations improve NLL, Brier score, and accuracy over constant regularization in all four CIFAR settings (Table~\ref{tab:app-schedule}). Yet constant regularization yields lower final edge JS on both evaluated ResNet-50 graphs (Table~\ref{tab:app-theory}): stronger final agreement does not necessarily imply better predictive quality. Sensitivity to neighborhood size is smaller than the NLL gain over CE in six of eight ablated settings.

No divergence uniformly dominates in the stable, scale-matched comparisons and activation rules do not transfer consistently across settings, so activation epoch and regularization strength remain validation-selected hyperparameters (Appendices~\ref{app:ablations} and~\ref{app:activation}). 

The regularizer is inexpensive per run: on CIFAR, the reported training-time increase is below $1\%$, with one-time graph construction taking $17$--$27$ seconds per run. Inference time is unchanged (Appendix~\ref{app:cost}). 

Finally, we test whether the gains survive shift and perturbation. Under input corruption, \model improves NLL and ECE over CE in four of five image settings, with higher accuracy in all five. On those same 4, it leads all other methods in accuracy. Under label noise, ECE improves in all four evaluated conditions, while NLL improves in only one. The gains therefore depend on the evaluation conditions and metric (Appendices~\ref{app:robustness} and~\ref{app:perturb}).

\subsection{When does \model help?}\label{sec:res-regime}

Proposition~\ref{prop:threshold} motivates two requirements for useful smoothing: neighboring errors must differ enough for averaging to help, and the connected points must have sufficiently similar correctness probabilities. Table~\ref{tab:regime} describes these conditions through mean edge JS and edge agreement ($|E|^{-1}\sum_{(i,j)\in E}\mathbf 1\{\hat y_i=\hat y_j\}$), measured at activation before regularized training. These diagnostics characterize the available disagreement and neighborhood coherence, although neither directly measures the true correctness gap. The small gains on Covertype, Jannis, and the two Amazon settings are consistent with limited disagreement to remove: their mean edge JS is only $0.003$--$0.013$, and their relative NLL improvements are $0.4$--$0.8\%$. Tiny-ImageNet presents a different limitation. Its high disagreement occurs on the least coherent graph, increasing the risk that smoothing suppresses genuine differences between neighbors as well as estimation error. Its relative NLL gain is $3.9\%$, compared with $8.6$--$11.7\%$ across the four CIFAR settings.

Helena and 20NG show why graph diagnostics alone are insufficient. Although their numbers overlap those of the CIFAR settings, their gains are much smaller. A plausible additional factor is how the model responds to the consistency constraint. The tabular MLPs learn representations, but through much shallower networks than the image ResNets. The 20NG model learns convolutional features over frozen word embeddings, while the Amazon models retain a frozen sentence encoder. These architectures offer different scopes for adapting features under the regularizer, including learning rate schedules. If neighboring predictions share errors rooted in the available features, making them agree will not correct those errors; the constraint would be satisfied by changing output probabilities. The observed gains are consistent with this limitation: they are $3.9$--$11.7\%$ in the image settings and $0.4$--$1.3\%$ in the tabular and text settings. Local consistency is therefore most useful when disagreement is both removable and supported by an informative, learnable representation.

\begin{table}[!htbp]
\centering
\caption{Mean edge JS and Edge Agreement at activation and improvement in test NLL over CE, sorted by edge JS. Positive $\Delta$NLL
indicates improvement; the relative column is as a fraction of CE's NLL.}
\label{tab:regime}
\small
\begin{tabular}{@{}lcccc@{}}
\toprule
Setting & Edge JS & Edge agreement & $\Delta$NLL & Relative \\
\midrule
Covertype        & 0.003 & 0.958 & 0.0017 & $0.7\%$ \\
Amazon (binary)  & 0.004 & 0.987 & 0.0014 & $0.6\%$ \\
Jannis           & 0.009 & 0.935 & 0.0059 & $0.8\%$ \\
Amazon (5-class) & 0.013 & 0.909 & 0.0028 & $0.4\%$ \\
CIFAR-10 / WRN   & 0.026 & 0.901 & 0.0220 & $11.2\%$ \\
20NG             & 0.027 & 0.748 & 0.0104 & $1.3\%$ \\
CIFAR-10 / R50   & 0.029 & 0.775 & 0.0222 & $10.8\%$ \\
Helena           & 0.078 & 0.678 & 0.0117 & $0.5\%$ \\
CIFAR-100 / R50  & 0.188 & 0.677 & 0.0756 & $8.6\%$ \\
CIFAR-100 / WRN  & 0.202 & 0.673 & 0.1025 & $11.7\%$ \\
Tiny-ImageNet     & 0.254 & 0.535 & 0.0578 & $3.9\%$ \\
\bottomrule
\end{tabular}
\end{table}

\section{Related Work}\label{sec:related}

\paragraph{Train-time calibration.}
Train-time approaches encourage calibration either explicitly or implicitly, through some modification to training. Explicit objectives include accuracy-uncertainty alignment~\citep{krishnan_avuc}, matching average class probabilities to empirical frequencies~\citep{Hebbalaguppe_mdca}, and penalizing confidence-accuracy discrepancies through soft binning~\citep{karandikar_sAVuC_sbECE} or a Huber loss~\citep{Patra_2023_huber}. Other approaches construct differentiable calibration objectives using kernel methods~\citep{Kumar_2018_MMCE,popordanoska_2022_dKDE}. Distribution matching~\citep{marx_2023_mmd} provides a flexible calibration objective whose conditioning kernels can explicitly target local calibration. Implicit approaches improve calibration without directly optimizing a calibration discrepancy, including FL~\citep{Lin_2017_focalloss,mukhoti_focal}, LS~\citep{Szegedy_2016_labelsmoothing,muller_label_smoothing}, and MbLS~\citep{Liu_2022_mbls}. SVLS~\citep{islam_2021_svls} constructs soft targets from neighboring ground-truth pixel labels and NACL~\citep{MURUGESAN_2025_NACL} penalizes logit deviations from a neighborhood-derived prior; both utilize local pixel structure specifically in image segmentation tasks. Mixup~\citep{zhang_2018_mixup} is also known to improve calibration~\citep{Thulasidasan_2019_mixup_cal}, but it is an augmentation protocol rather than an objective: it synthesizes inputs and interpolates labels between randomly paired samples. In comparison, \model uses representation-space neighborhoods to regularize agreement between full predictive distributions, with no label access required by the smoothing penalty. 

\paragraph{Local post-hoc calibration.}
Several post-hoc methods exploit locality to recalibrate pretrained models. LoRe~\citep{luo_lce} estimates correctness through kernel-weighted local accuracy, while KCal~\citep{lin_2023_kcal} constructs class probabilities using kernel density estimates in a learned metric space and DAC~\citep{Tomani_2023_DAC} uses density proxies from hidden representations to determine sample-dependent temperatures. LoCal~\citep{barbera2026multiclass} learns a representation projection and new logits, aligning predictions with kernel-estimated local class frequencies using Jensen--Shannon distance. LoCal is the post-hoc method most closely related to \model, but the objectives differ: LoCal matches predictions to local label-frequency estimates in a separate calibration stage, directly targeting multiclass local calibration, whereas \model encourages agreement between neighboring predictive distributions during classifier training, connecting predictive smoothness to confidence variation and the propagation of pointwise confidence calibration.

\paragraph{Graph regularization.}
Our approach builds on graph-based manifold regularization, which encourages functions to vary smoothly over the data geometry~\citep{Belkin_2006_manifold_learning}. Similarly, NCR~\citep{iscen_2022_learning} penalizes the KL divergence between each example's prediction and the similarity-weighted average of its neighbors' predictions, so that neighbors can override noisy labels. A mixture target can be met while individual neighbors disagree sharply; \model instead penalizes every edge, as its bounds require. NCR also draws neighbors from the current minibatch, whereas \model fixes a kNN graph over the full training set, and NCR targets accuracy under label noise rather than calibration.

\section{Conclusion}
We introduced graph smoothing for train-time calibration and instantiated it as \modelNoSpace, a JS neighborhood regularizer with no inference-time cost. Our analysis bounds pointwise calibration-error propagation and identifies both an obstruction and a condition for improvement under idealized averaging. Across eleven benchmarks, \model improves NLL over CE; after TS, it attains the lowest NLL among the evaluated train-time methods in seven of eight image and tabular settings. Three findings characterize the method's behavior: representation-based neighborhoods yield lower NLL than random neighborhoods; stronger final neighborhood agreement does not necessarily imply better predictive quality; and gains persist under a common temperature-scaling procedure. Together, these findings support neighborhood smoothing as a framework for train-time calibration and predictive improvement.

Future work should consider further development of the graph smoothing framework, and may focus on encouraging smoothing in the regimes where it helps, either through the penalty or through hyperparameters, and suppressing it on harmful edges. It would also be worthwhile to investigate alternative graph-based penalties or underlying graph structures, and explore graph smoothing on benchmarks where the graph is inherent rather than induced.

\section*{Reproducibility Statement}
We provide all materials necessary to reproduce and verify our results throughout the paper and appendices. Appendix~\ref{app:train-details} specifies data processing, splits, architectures, optimization, search spaces, selection, metrics, and hardware. Appendix~\ref{app:nescal:complexity} and Algorithm~\ref{alg:nescal} specify \modelNoSpace; Appendices~\ref{app:gen-proofs} and~\ref{app:specific-proofs} provide assumptions and proofs. Seed counts and split conventions, including exceptions, are documented in Appendix~\ref{app:train-details}. Run manifests identify selected configurations and reported aggregates. Reproduction code is available at \url{https://github.com/idanh8/nescal_paper_code}.

\bibliography{iclr2027_conference}

@inproceedings{Naeini_2015_ECE,
author = {Naeini, Mahdi Pakdaman and Cooper, Gregory F. and Hauskrecht, Milos},
title = {Obtaining well calibrated probabilities using bayesian binning},
year = {2015},
isbn = {0262511290},
publisher = {AAAI Press},
booktitle = {Proceedings of the Twenty-Ninth AAAI Conference on Artificial Intelligence},
pages = {2901–2907},
numpages = {7},
location = {Austin, Texas},
series = {AAAI'15}
}

@inproceedings{Guo_2107_calibration,
author = {Guo, Chuan and Pleiss, Geoff and Sun, Yu and Weinberger, Kilian Q.},
title = {On calibration of modern neural networks},
year = {2017},
publisher = {JMLR.org},
booktitle = {Proceedings of the 34th International Conference on Machine Learning - Volume 70},
pages = {1321–1330},
numpages = {10},
location = {Sydney, NSW, Australia},
series = {ICML'17}
}

@inproceedings{Blasiok_2023_measuring_calibration,
author = {Błasiok, Jarosław and Gopalan, Parikshit and Hu, Lunjia and Nakkiran, Preetum},
title = {A Unifying Theory of Distance from Calibration},
year = {2023},
isbn = {9781450399135},
publisher = {Association for Computing Machinery},
address = {New York, NY, USA},
url = {https://doi.org/10.1145/3564246.3585182},
doi = {10.1145/3564246.3585182},
booktitle = {Proceedings of the 55th Annual ACM Symposium on Theory of Computing},
pages = {1727–1740},
numpages = {14},
location = {Orlando, FL, USA},
series = {STOC 2023}
}

@article{Arrieta_Ibarra_2022_measuring_calibration,
author = {Arrieta-Ibarra, Imanol and Gujral, Paman and Tannen, Jonathan and Tygert, Mark and Xu, Cherie},
title = {Metrics of calibration for probabilistic predictions},
year = {2022},
issue_date = {January 2022},
publisher = {JMLR.org},
volume = {23},
number = {1},
issn = {1532-4435},
journal = {J. Mach. Learn. Res.},
month = jan,
articleno = {351},
numpages = {55}
}

@inproceedings{Widmann_2019_measuring_calibration,
 author = {Widmann, David and Lindsten, Fredrik and Zachariah, Dave},
 booktitle = {Advances in Neural Information Processing Systems},
 editor = {H. Wallach and H. Larochelle and A. Beygelzimer and F. d\textquotesingle Alch\'{e}-Buc and E. Fox and R. Garnett},
 pages = {},
 publisher = {Curran Associates, Inc.},
 title = {Calibration tests in multi-class classification: A unifying framework},
 url = {https://proceedings.neurips.cc/paper_files/paper/2019/file/1c336b8080f82bcc2cd2499b4c57261d-Paper.pdf},
 volume = {32},
 year = {2019}
}

@InProceedings{Nixon_2019_measuring_calibration,
author = {Nixon, Jeremy and Dusenberry, Michael W. and Zhang, Linchuan and Jerfel, Ghassen and Tran, Dustin},
title = {Measuring Calibration in Deep Learning},
booktitle = {Proceedings of the IEEE/CVF Conference on Computer Vision and Pattern Recognition (CVPR) Workshops},
month = {June},
year = {2019}
}

@InProceedings{Vaicenavicius_2019_measuring_calibration,
  title = 	 {Evaluating model calibration in classification},
  author =       {Vaicenavicius, Juozas and Widmann, David and Andersson, Carl and Lindsten, Fredrik and Roll, Jacob and Sch\"{o}n, Thomas},
  booktitle = 	 {Proceedings of the Twenty-Second International Conference on Artificial Intelligence and Statistics},
  pages = 	 {3459--3467},
  year = 	 {2019},
  editor = 	 {Chaudhuri, Kamalika and Sugiyama, Masashi},
  volume = 	 {89},
  series = 	 {Proceedings of Machine Learning Research},
  month = 	 {16--18 Apr},
  publisher =    {PMLR},
  url = 	 {https://proceedings.mlr.press/v89/vaicenavicius19a.html}
}

@InProceedings{Kumar_2018_MMCE,
  title =          {Trainable Calibration Measures for Neural Networks from Kernel Mean Embeddings},
  author =       {Kumar, Aviral and Sarawagi, Sunita and Jain, Ujjwal},
  booktitle =          {Proceedings of the 35th International Conference on Machine Learning},
  pages =          {2805--2814},
  year =          {2018},
  editor =          {Dy, Jennifer and Krause, Andreas},
  volume =          {80},
  series =          {Proceedings of Machine Learning Research},
  month =          {10--15 Jul},
  publisher =    {PMLR},
  url =          {https://proceedings.mlr.press/v80/kumar18a.html}
}

@InProceedings{zhang_2020_mixnmatch_desiderata,
  title = 	 {Mix-n-Match : Ensemble and Compositional Methods for Uncertainty Calibration in Deep Learning},
  author =       {Zhang, Jize and Kailkhura, Bhavya and Han, T. Yong-Jin},
  booktitle = 	 {Proceedings of the 37th International Conference on Machine Learning},
  pages = 	 {11117--11128},
  year = 	 {2020},
  editor = 	 {III, Hal Daumé and Singh, Aarti},
  volume = 	 {119},
  series = 	 {Proceedings of Machine Learning Research},
  month = 	 {13--18 Jul},
  publisher =    {PMLR},
  url = 	 {https://proceedings.mlr.press/v119/zhang20k.html}
}

@ARTICLE{Endres_2003_JSD,
  author={Endres, D.M. and Schindelin, J.E.},
  journal={IEEE Transactions on Information Theory}, 
  title={A new metric for probability distributions}, 
  year={2003},
  volume={49},
  number={7},
  pages={1858-1860},
  doi={10.1109/TIT.2003.813506}}

@INPROCEEDINGS{Fuglede_2004_JS_distance_metric,
  author={Fuglede, B. and Topsoe, F.},
  booktitle={International Symposium onInformation Theory, 2004. ISIT 2004. Proceedings.}, 
  title={Jensen-Shannon divergence and Hilbert space embedding}, 
  year={2004},
  volume={},
  number={},
  pages={31-},
  doi={10.1109/ISIT.2004.1365067}}

@article{Belkin_2006_manifold_learning,
  author  = {Mikhail Belkin and Partha Niyogi and Vikas Sindhwani},
  title   = {Manifold  Regularization: A Geometric Framework for Learning from Labeled and Unlabeled Examples},
  journal = {Journal of Machine Learning Research},
  year    = {2006},
  volume  = {7},
  number  = {85},
  pages   = {2399--2434},
  url     = {http://jmlr.org/papers/v7/belkin06a.html}
}

@inproceedings{kull_dirichlet_cwece,
 author = {Kull, Meelis and Perello Nieto, Miquel and K\"{a}ngsepp, Markus and Silva Filho, Telmo and Song, Hao and Flach, Peter},
 booktitle = {Advances in Neural Information Processing Systems},
 editor = {H. Wallach and H. Larochelle and A. Beygelzimer and F. d\textquotesingle Alch\'{e}-Buc and E. Fox and R. Garnett},
 pages = {},
 publisher = {Curran Associates, Inc.},
 title = {Beyond temperature scaling: Obtaining well-calibrated multi-class probabilities with Dirichlet calibration},
 url = {https://proceedings.neurips.cc/paper_files/paper/2019/file/8ca01ea920679a0fe3728441494041b9-Paper.pdf},
 volume = {32},
 year = {2019}
}

@InProceedings{luo_lce,
  title = 	 {Local calibration: metrics and recalibration},
  author =       {Luo, Rachel and Bhatnagar, Aadyot and Bai, Yu and Zhao, Shengjia and Wang, Huan and Xiong, Caiming and Savarese, Silvio and Ermon, Stefano and Schmerling, Edward and Pavone, Marco},
  booktitle = 	 {Proceedings of the Thirty-Eighth Conference on Uncertainty in Artificial Intelligence},
  pages = 	 {1286--1295},
  year = 	 {2022},
  editor = 	 {Cussens, James and Zhang, Kun},
  volume = 	 {180},
  series = 	 {Proceedings of Machine Learning Research},
  month = 	 {01--05 Aug},
  publisher =    {PMLR},
  url = 	 {https://proceedings.mlr.press/v180/luo22a.html}
}

@article{zadrozny_posthoc_isotonic_regression,
author = {Zadrozny, Bianca and Elkan, Charles},
year = {2002},
month = {08},
pages = {},
title = {Transforming Classifier Scores into Accurate Multiclass Probability Estimates},
journal = {Proceedings of the ACM SIGKDD International Conference on Knowledge Discovery and Data Mining},
doi = {10.1145/775047.775151}
}

@article{platt__posthoc_platt_scaling,
author = {Platt, John},
year = {2000},
month = {06},
pages = {},
title = {Probabilistic Outputs for Support Vector Machines and Comparisons to Regularized Likelihood Methods},
volume = {10},
journal = {Adv. Large Margin Classif.}
}

@inproceedings{karandikar_sAVuC_sbECE,
 author = {Karandikar, Archit and Cain, Nicholas and Tran, Dustin and Lakshminarayanan, Balaji and Shlens, Jonathon and Mozer, Michael C and Roelofs, Becca},
 booktitle = {Advances in Neural Information Processing Systems},
 editor = {M. Ranzato and A. Beygelzimer and Y. Dauphin and P.S. Liang and J. Wortman Vaughan},
 pages = {29768--29779},
 publisher = {Curran Associates, Inc.},
 title = {Soft Calibration Objectives for Neural Networks},
 url = {https://proceedings.neurips.cc/paper_files/paper/2021/file/f8905bd3df64ace64a68e154ba72f24c-Paper.pdf},
 volume = {34},
 year = {2021}
}

@inproceedings{krishnan_avuc,
 author = {Krishnan, Ranganath and Tickoo, Omesh},
 booktitle = {Advances in Neural Information Processing Systems},
 editor = {H. Larochelle and M. Ranzato and R. Hadsell and M.F. Balcan and H. Lin},
 pages = {18237--18248},
 publisher = {Curran Associates, Inc.},
 title = {Improving model calibration with accuracy versus uncertainty optimization},
 url = {https://proceedings.neurips.cc/paper_files/paper/2020/file/d3d9446802a44259755d38e6d163e820-Paper.pdf},
 volume = {33},
 year = {2020}
}

@inproceedings{Hebbalaguppe_mdca,
  author={Ramya Hebbalaguppe and Jatin Prakash and Neelabh Madan and Chetan Arora},
  title={A Stitch in Time Saves Nine: A Train-Time Regularizing Loss for Improved Neural Network Calibration},
  booktitle={CVPR},
  year={2022},
  pages={16060--16069},
  url={https://doi.org/10.1109/CVPR52688.2022.01561}
}

@inproceedings{mukhoti_focal,
 author = {Mukhoti, Jishnu and Kulharia, Viveka and Sanyal, Amartya and Golodetz, Stuart and Torr, Philip and Dokania, Puneet},
 booktitle = {Advances in Neural Information Processing Systems},
 editor = {H. Larochelle and M. Ranzato and R. Hadsell and M.F. Balcan and H. Lin},
 pages = {15288--15299},
 publisher = {Curran Associates, Inc.},
 title = {Calibrating Deep Neural Networks using Focal Loss},
 url = {https://proceedings.neurips.cc/paper_files/paper/2020/file/aeb7b30ef1d024a76f21a1d40e30c302-Paper.pdf},
 volume = {33},
 year = {2020}
}

@inproceedings{muller_label_smoothing,
 author = {M\"{u}ller, Rafael and Kornblith, Simon and Hinton, Geoffrey E},
 booktitle = {Advances in Neural Information Processing Systems},
 editor = {H. Wallach and H. Larochelle and A. Beygelzimer and F. d\textquotesingle Alch\'{e}-Buc and E. Fox and R. Garnett},
 pages = {},
 publisher = {Curran Associates, Inc.},
 title = {When does label smoothing help?},
 url = {https://proceedings.neurips.cc/paper_files/paper/2019/file/f1748d6b0fd9d439f71450117eba2725-Paper.pdf},
 volume = {32},
 year = {2019}
}

@InProceedings{roelos_2022_ECEBias,
  title = 	 { Mitigating Bias in Calibration Error Estimation },
  author =       {Roelofs, Rebecca and Cain, Nicholas and Shlens, Jonathon and Mozer, Michael C.},
  booktitle = 	 {Proceedings of The 25th International Conference on Artificial Intelligence and Statistics},
  pages = 	 {4036--4054},
  year = 	 {2022},
  editor = 	 {Camps-Valls, Gustau and Ruiz, Francisco J. R. and Valera, Isabel},
  volume = 	 {151},
  series = 	 {Proceedings of Machine Learning Research},
  month = 	 {28--30 Mar},
  publisher =    {PMLR},
  url = 	 {https://proceedings.mlr.press/v151/roelofs22a.html}
}

@InProceedings{Szegedy_2016_labelsmoothing,
author = {Szegedy, Christian and Vanhoucke, Vincent and Ioffe, Sergey and Shlens, Jon and Wojna, Zbigniew},
title = {Rethinking the Inception Architecture for Computer Vision},
booktitle = {Proceedings of the IEEE Conference on Computer Vision and Pattern Recognition (CVPR)},
month = {June},
year = {2016}
}

@InProceedings{Lin_2017_focalloss,
author = {Lin, Tsung-Yi and Goyal, Priya and Girshick, Ross and He, Kaiming and Dollar, Piotr},
title = {Focal Loss for Dense Object Detection},
booktitle = {Proceedings of the IEEE International Conference on Computer Vision (ICCV)},
month = {Oct},
year = {2017}
}

@book{amari_200_fisher,
  title={Methods of Information Geometry},
  author={Amari, S. and Nagaoka, H.},
  isbn={9780821805312},
  lccn={00059362},
  series={Fields Institute Communications},
  url={https://books.google.co.za/books?id=qGzCwAEACAAJ},
  year={2000},
  publisher={American Mathematical Society}
}

@inproceedings{popordanoska_2022_dKDE,
 author = {Popordanoska, Teodora and Sayer, Raphael and Blaschko, Matthew},
 booktitle = {Advances in Neural Information Processing Systems},
 editor = {S. Koyejo and S. Mohamed and A. Agarwal and D. Belgrave and K. Cho and A. Oh},
 pages = {7933--7946},
 publisher = {Curran Associates, Inc.},
 title = {A Consistent and Differentiable Lp Canonical Calibration Error Estimator},
 url = {https://proceedings.neurips.cc/paper_files/paper/2022/file/33d6e648ee4fb24acec3a4bbcd4f001e-Paper-Conference.pdf},
 volume = {35},
 year = {2022}
}

@InProceedings{Patra_2023_huber,
    author    = {Patra, Rishabh and Hebbalaguppe, Ramya and Dash, Tirtharaj and Shroff, Gautam and Vig, Lovekesh},
    title     = {Calibrating Deep Neural Networks Using Explicit Regularisation and Dynamic Data Pruning},
    booktitle = {Proceedings of the IEEE/CVF Winter Conference on Applications of Computer Vision (WACV)},
    month     = {January},
    year      = {2023},
    pages     = {1541-1549}
}

@inproceedings{marx_2023_mmd,
 author = {Marx, Charlie and Zalouk, Sofian and Ermon, Stefano},
 booktitle = {Advances in Neural Information Processing Systems},
 editor = {A. Oh and T. Naumann and A. Globerson and K. Saenko and M. Hardt and S. Levine},
 pages = {25910--25928},
 publisher = {Curran Associates, Inc.},
 title = {Calibration by Distribution Matching: Trainable Kernel Calibration Metrics},
 url = {https://proceedings.neurips.cc/paper_files/paper/2023/file/52493d82db00e73abb2858a5a5f28717-Paper-Conference.pdf},
 volume = {36},
 year = {2023}
}

@InProceedings{Liu_2022_mbls,
    author    = {Liu, Bingyuan and Ben Ayed, Ismail and Galdran, Adrian and Dolz, Jose},
    title     = {The Devil Is in the Margin: Margin-Based Label Smoothing for Network Calibration},
    booktitle = {Proceedings of the IEEE/CVF Conference on Computer Vision and Pattern Recognition (CVPR)},
    month     = {June},
    year      = {2022},
    pages     = {80-88}
}

@article{MURUGESAN_2025_NACL,
title = {Neighbor-aware calibration of segmentation networks with penalty-based constraints},
journal = {Medical Image Analysis},
volume = {101},
pages = {103501},
year = {2025},
issn = {1361-8415},
doi = {https://doi.org/10.1016/j.media.2025.103501},
url = {https://www.sciencedirect.com/science/article/pii/S1361841525000490},
author = {Balamurali Murugesan and Sukesh {Adiga Vasudeva} and Bingyuan Liu and Herve Lombaert and Ismail {Ben Ayed} and Jose Dolz}
}

@inproceedings{Tomani_2023_DAC,
author = {Tomani, Christian and Waseda, Futa and Shen, Yuesong and Cremers, Daniel},
title = {Beyond in-domain scenarios: robust density-aware calibration},
year = {2023},
publisher = {JMLR.org},
booktitle = {Proceedings of the 40th International Conference on Machine Learning},
articleno = {1429},
numpages = {25},
location = {Honolulu, Hawaii, USA},
series = {ICML'23}
}

@inproceedings{
lin_2023_kcal,
title={Taking a Step Back with {KC}al: Multi-Class Kernel-Based Calibration for Deep Neural Networks},
author={Zhen Lin and Shubhendu Trivedi and Jimeng Sun},
booktitle={The Eleventh International Conference on Learning Representations },
year={2023},
url={https://openreview.net/forum?id=p_jIy5QFB7}
}

@inproceedings{
barbera2026multiclass,
title={Multiclass Local Calibration with the Jensen-Shannon Distance},
author={Cesare Barbera and Lorenzo Perini and Giovanni De Toni and Andrea Passerini and Andrea Pugnana},
booktitle={The 29th International Conference on Artificial Intelligence and Statistics},
year={2026},
url={https://openreview.net/forum?id=atUXOpyCox}
}

@inproceedings{islam_2021_svls,
author = {Islam, Mobarakol and Glocker, Ben},
title = {Spatially Varying Label Smoothing: Capturing Uncertainty from Expert Annotations},
year = {2021},
isbn = {978-3-030-78190-3},
publisher = {Springer-Verlag},
address = {Berlin, Heidelberg},
url = {https://doi.org/10.1007/978-3-030-78191-0_52},
doi = {10.1007/978-3-030-78191-0_52},
booktitle = {Information Processing in Medical Imaging: 27th International Conference, IPMI 2021, Virtual Event, June 28–June 30, 2021, Proceedings},
pages = {677–688},
numpages = {12}
}

@techreport{krizhevsky_2009_cifar,
  title={Learning multiple layers of features from tiny images},
  author={Krizhevsky, Alex and Hinton, Geoffrey},
  year={2009},
  institution={University of Toronto},
  address={Toronto, Ontario},
  url={https://www.cs.toronto.edu/~kriz/learning-features-2009-TR.pdf}
}

@InProceedings{he_2016_resnet,
author = {He, Kaiming and Zhang, Xiangyu and Ren, Shaoqing and Sun, Jian},
title = {Deep Residual Learning for Image Recognition},
booktitle = {Proceedings of the IEEE Conference on Computer Vision and Pattern Recognition (CVPR)},
month = {June},
year = {2016}
}

@inproceedings{zagoruyko_2016_wrn,
        	title={Wide Residual Networks},
        	author={Sergey Zagoruyko and Nikos Komodakis},
        	year={2016},
        	month={September},
        	pages={87.1-87.12},
        	articleno={87},
        	numpages={12},
        	booktitle={Proceedings of the British Machine Vision Conference (BMVC)},
        	publisher={BMVA Press},
        	editor={Richard C. Wilson, Edwin R. Hancock and William A. P. Smith},
        	doi={10.5244/C.30.87},
        	isbn={1-901725-59-6},
        	url={https://dx.doi.org/10.5244/C.30.87}
        }

@article{le_2015_tinyimagenet,
title={Tiny imagenet visual recognition challenge},
author={Le, Y. and Yang, X.},
journal={CS 231N},
volume={7},
number={7},
pages={3},
year={2015}
}

@inproceedings{lang_1995_20ng,
  title={NewsWeeder: Learning to filter netnews},
  author={Lang, Ken},
  booktitle={Proceedings of the 12th International Conference on Machine Learning (ICML)},
  pages={331--339},
  year={1995}, 
  url = {http://qwone.com/~jason/20Newsgroups/},
  note = {Retrived from source, which documents that the cited paper does not explicitly acknowledge collecting this dataset}
}

@inproceedings{pennington_2014_glove,
    title = "{G}lo{V}e: Global Vectors for Word Representation",
    author = "Pennington, Jeffrey  and
      Socher, Richard  and
      Manning, Christopher",
    editor = "Moschitti, Alessandro  and
      Pang, Bo  and
      Daelemans, Walter",
    booktitle = "Proceedings of the 2014 Conference on Empirical Methods in Natural Language Processing ({EMNLP})",
    month = oct,
    year = "2014",
    address = "Doha, Qatar",
    publisher = "Association for Computational Linguistics",
    url = "https://aclanthology.org/D14-1162/",
    doi = "10.3115/v1/D14-1162",
    pages = "1532--1543"
}

@misc{liu_2019_roberta,
      title={RoBERTa: A Robustly Optimized BERT Pretraining Approach}, 
      author={Yinhan Liu and Myle Ott and Naman Goyal and Jingfei Du and Mandar Joshi and Danqi Chen and Omer Levy and Mike Lewis and Luke Zettlemoyer and Veselin Stoyanov},
      year={2019},
      eprint={1907.11692},
      archivePrefix={arXiv},
      primaryClass={cs.CL},
      url={https://arxiv.org/abs/1907.11692}, 
}

@article{hou_2024_amazonreviews,
  title={Bridging Language and Items for Retrieval and Recommendation},
  author={Hou, Yupeng and Li, Jiacheng and He, Zhankui and Yan, An and Chen, Xiusi and McAuley, Julian},
  journal={arXiv preprint arXiv:2403.03952},
  year={2024}
}

@inproceedings{gorishniy_2021_tabular,
 author = {Gorishniy, Yury and Rubachev, Ivan and Khrulkov, Valentin and Babenko, Artem},
 booktitle = {Advances in Neural Information Processing Systems},
 editor = {M. Ranzato and A. Beygelzimer and Y. Dauphin and P.S. Liang and J. Wortman Vaughan},
 pages = {18932--18943},
 publisher = {Curran Associates, Inc.},
 title = {Revisiting Deep Learning Models for Tabular Data},
 url = {https://proceedings.neurips.cc/paper_files/paper/2021/file/9d86d83f925f2149e9edb0ac3b49229c-Paper.pdf},
 volume = {34},
 year = {2021}
}

@misc{blackard_covertype_31,
  author       = {Blackard, Jock},
  title        = {{Covertype}},
  year         = {1998},
  howpublished = {UCI Machine Learning Repository},
  note         = {{DOI}: https://doi.org/10.24432/C50K5N}
}

@incollection{Guyon_2019_helena,
author="Guyon, Isabelle
and Sun-Hosoya, Lisheng
and Boull{\'e}, Marc
and Escalante, Hugo Jair
and Escalera, Sergio
and Liu, Zhengying
and Jajetic, Damir
and Ray, Bisakha
and Saeed, Mehreen
and Sebag, Mich{\`e}le
and Statnikov, Alexander
and Tu, Wei-Wei
and Viegas, Evelyne",
editor="Hutter, Frank
and Kotthoff, Lars
and Vanschoren, Joaquin",
title="Analysis of the AutoML Challenge Series 2015--2018",
bookTitle="Automated Machine Learning: Methods, Systems, Challenges",
year="2019",
publisher="Springer International Publishing",
address="Cham",
pages="177--219",
isbn="978-3-030-05318-5",
doi="10.1007/978-3-030-05318-5_10",
url="https://doi.org/10.1007/978-3-030-05318-5_10"
}

@InProceedings{kull_2015_brier_decomposition,
author="Kull, Meelis
and Flach, Peter",
editor="Appice, Annalisa
and Rodrigues, Pedro Pereira
and Santos Costa, V{\'i}tor
and Soares, Carlos
and Gama, Jo{\~a}o
and Jorge, Al{\'i}pio",
title="Novel Decompositions of Proper Scoring Rules for Classification: Score Adjustment as Precursor to Calibration",
booktitle="Machine Learning and Knowledge Discovery in Databases",
year="2015",
publisher="Springer International Publishing",
address="Cham",
pages="68--85",
isbn="978-3-319-23528-8"
}

@article{Gneiting_2007_proper_scoring,
author = {Tilmann Gneiting and Adrian E Raftery},
title = {Strictly Proper Scoring Rules, Prediction, and Estimation},
journal = {Journal of the American Statistical Association},
volume = {102},
number = {477},
pages = {359--378},
year = {2007},
publisher = {Taylor \& Francis},
doi = {10.1198/016214506000001437},
URL = {https://doi.org/10.1198/016214506000001437},
eprint = {https://doi.org/10.1198/016214506000001437}
}

@article {brier_1950_brier,
      author = "Glenn W.  Brier",
      title = "VERIFICATION OF FORECASTS EXPRESSED IN TERMS OF PROBABILITY",
      journal = "Monthly Weather Review",
      year = "1950",
      publisher = "American Meteorological Society",
      address = "Boston MA, USA",
      volume = "78",
      number = "1",
      doi = "10.1175/1520-0493(1950)078<0001:VOFEIT>2.0.CO;2",
      pages=      "1 - 3",
      url = "https://journals.ametsoc.org/view/journals/mwre/78/1/1520-0493_1950_078_0001_vofeit_2_0_co_2.xml"
}

@inproceedings{
hendrycks_2019_cifarc,
title={Benchmarking Neural Network Robustness to Common Corruptions and Perturbations},
author={Dan Hendrycks and Thomas Dietterich},
booktitle={International Conference on Learning Representations},
year={2019},
url={https://openreview.net/forum?id=HJz6tiCqYm},
}

@article{netzer_2011_svhn,
author = {Netzer, Yuval and Wang, Tao and Coates, Adam and Bissacco, Alessandro and Wu, Bo and Ng, Andrew},
year = {2011},
month = {01},
pages = {},
title = {Reading Digits in Natural Images with Unsupervised Feature Learning},
journal = {NIPS}
}

@inproceedings{hendrycks_2017_OOD,
title={A Baseline for Detecting Misclassified and Out-of-Distribution Examples in Neural Networks},
author={Dan Hendrycks and Kevin Gimpel},
booktitle={International Conference on Learning Representations},
year={2017},
url={https://openreview.net/forum?id=Hkg4TI9xl}
}

@inproceedings{geifman_2019_AURC,
title={Bias-Reduced Uncertainty Estimation for Deep Neural Classifiers},
author={Yonatan Geifman and Guy Uziel and Ran El-Yaniv},
booktitle={International Conference on Learning Representations},
year={2019},
url={https://openreview.net/forum?id=SJfb5jCqKm},
}

@inproceedings{zhang_2018_mixup,
title={mixup: Beyond Empirical Risk Minimization},
author={Hongyi Zhang and Moustapha Cisse and Yann N. Dauphin and David Lopez-Paz},
booktitle={International Conference on Learning Representations},
year={2018},
url={https://openreview.net/forum?id=r1Ddp1-Rb},
}

@inproceedings{Thulasidasan_2019_mixup_cal,
 author = {Thulasidasan, Sunil and Chennupati, Gopinath and Bilmes, Jeff A and Bhattacharya, Tanmoy and Michalak, Sarah},
 booktitle = {Advances in Neural Information Processing Systems},
 editor = {H. Wallach and H. Larochelle and A. Beygelzimer and F. d\textquotesingle Alch\'{e}-Buc and E. Fox and R. Garnett},
 pages = {},
 publisher = {Curran Associates, Inc.},
 title = {On Mixup Training: Improved Calibration and Predictive Uncertainty for Deep Neural Networks},
 url = {https://proceedings.neurips.cc/paper_files/paper/2019/file/36ad8b5f42db492827016448975cc22d-Paper.pdf},
 volume = {32},
 year = {2019}
}

@article{lin_2014_gpcnn,
  title={Network in network},
  author={Lin, Min and Chen, Qiang and Yan, Shuicheng},
  journal={arXiv preprint arXiv:1312.4400},
  year={2013}
}

@inproceedings{iscen_2022_learning,
author = { Iscen, Ahmet and Valmadre, Jack and Arnab, Anurag and Schmid, Cordelia },
booktitle = { 2022 IEEE/CVF Conference on Computer Vision and Pattern Recognition (CVPR) },
title = {{ Learning with Neighbor Consistency for Noisy Labels }},
year = {2022},
volume = {},
ISSN = {},
pages = {4662-4671},
doi = {10.1109/CVPR52688.2022.00463},
url = {https://doi.ieeecomputersociety.org/10.1109/CVPR52688.2022.00463},
publisher = {IEEE Computer Society},
address = {Los Alamitos, CA, USA},
month =Jun}
\bibliographystyle{iclr2027_conference}

\newpage
\appendix
\section{Further Theoretical Results and Proofs} \label{app:gen-proofs}

\subsection{Graph Laplacian Approximation}

\begin{definition}
    Given an unweighted, symmetric adjacency matrix $W \in \{0,1\}^{n \times n}$, the graph Laplacian is $L := \operatorname{diag}(d_1,\ldots,d_n) - W$, where $d_i := \sum_j W_{ij}$. For any function $u : V \to \mathbb{R}$ on the nodes, the Laplacian quadratic form
    \begin{equation}\label{eq:laplacian-quad}
    u^\top L u = \frac{1}{2}\sum_{i,j} W_{ij}\left(u_i - u_j\right)^2
\end{equation}
measures the smoothness of $u$ over the graph, growing as $u$ varies more sharply between neighboring nodes.
\end{definition}

\begin{proposition}[Graph Laplacian Approximation]
\label{lem:laplacian}
Let $\mathrm{D}:\Delta^{K-1}\times\Delta^{K-1}\to\mathbb{R}_{\ge 0}$ satisfy $\mathrm{D}(p \Vert p)=0$ and, for some $\varepsilon>0$, let $\mathrm{D}(p\Vert\cdot)$ be three times continuously differentiable on a neighborhood of $\Delta_\varepsilon := \{u\in\Delta^{K-1}: \min_k u_k \ge \varepsilon\}$ in $\mathbb{R}^K$, with third derivatives jointly continuous in $(p,q)$ on $\Delta_\varepsilon\times\Delta_\varepsilon$; write $M_3(\varepsilon) := \sup_{p,q\in\Delta_\varepsilon}\sup_{\|u\|_2=1}\bigl|\nabla_q^3\mathrm{D}(p\Vert q)[u,u,u]\bigr|$, which is finite by compactness. Suppose the predictive distributions obey the probability floor $\min_{i,k} p_{i,k} \ge \varepsilon$. Define the induced local form: $G_{\mathrm{D}}(p) := \tfrac{1}{2} \nabla_q^2 \mathrm{D}(p \Vert q)\big|_{q=p}$.  Then for any edge $(i,j) \in E$,
\begin{equation}
  \mathrm{D}(p_i \Vert p_j)  = (p_i-p_j)^{\mathsf{T}} G_{\mathrm{D}}(p_i)(p_i-p_j) + \rho_{ij}
\end{equation}
where $|\rho_{ij}| \le \tfrac{1}{6}M_3(\varepsilon)\,\|p_i-p_j\|_2^{3}$. Consequently,
\begin{equation}
  R_n(p)=\sum_{i,j} W_{ij}\,(p_i-p_j)^{\mathsf{T}}G_{\mathrm{D}}(p_i)\,(p_i-p_j) + \rho, \qquad |\rho| \le \tfrac{1}{6}M_3(\varepsilon)\sum_{i,j} W_{ij}\,\|p_i-p_j\|_2^{3}.
\end{equation}
\end{proposition}

Expanding $\mathrm{D}$ on each edge about the source prediction shows that $R_n(p)$ is, up to an explicitly bounded remainder, a weighted graph-Laplacian quadratic form, giving it a geometric interpretation as a smoothness penalty over some representation manifold $\mathcal{M}\subseteq\mathcal{Z}$. The result holds for any sufficiently regular divergence; the specific weighting is determined by the divergence's local curvature. The expansion is anchored at the source prediction $p_i$, weighting edge $(i,j)$ by $G_{\mathrm{D}}(p_i)$. Since $p_i=f_\theta(x_i)$, minimizing $R_n$ acts as a smoothness penalty on $f_\theta$. Its magnitude depends on both the predictive distributions and the edge weights.

Proposition~\ref{lem:laplacian} expresses $R_n$ as a weighted graph-Laplacian-like smoothness penalty and determines the weighting $G_{\mathrm{D}}(p_i)$, but no result below depends on it; the guarantees follow instead from Proposition~\ref{cor:confidence-smoothness}. The quadratic form is an accurate approximation only where $|p_{i,k}-p_{j,k}|\ll p_{i,k}$; it is an interpretive device, not a bound we optimize.

\paragraph{Proof.}

Fix an edge $(i,j)$ and write $p:=p_i$, $q:=p_j$, and $\delta:=q-p$. Since $p,q\in\Delta^{K-1}$ are probability vectors, they satisfy $ \sum_{k=1}^Kp_k = \sum_{k=1}^Kq_k = 1$, so $\mathbf{1}^\top\delta=\sum_k\delta_k=0$. The expansion below is Taylor's theorem with Lagrange remainder applied to $t\mapsto\mathrm{D}(p\Vert p+t\delta)$; it is exact for any $\delta$.

We take the multivariate Taylor expansion of $D(p \Vert q)$ about $q=p$ in its second argument:
\begin{equation}\label{eq:gen-taylor}
  \mathrm{D}(p \Vert q) = \mathrm{D}(p \Vert p) + \nabla_q\mathrm{D}(p \Vert q)\big|_{q=p}^{\top}\delta + \tfrac{1}{2}\delta^\top\nabla_q^2\mathrm{D}(p \Vert q)\big|_{q=p}\,\delta+ \rho_{ij}
\end{equation}

\emph{Zeroth order.} By hypothesis $\mathrm{D}(p \Vert p)= 0$.

\emph{First order.} For fixed $p$, $\mathrm{D}(p \Vert q)$ is nonnegative in its second argument and equals $0$ at $q=p$, so $q=p$ minimizes $\mathrm{D}(p \Vert \cdot)$ over the simplex. Since $p$ lies in the interior, the only binding constraint at this minimum is that probabilities sum to one, $\mathbf{1}^\top q = 1$. Stationarity then forces the gradient to be constant across coordinates, $\nabla_q\mathrm{D}(p \Vert q)\big|_{q=p} = \mu\mathbf{1}$ for some $\mu\in\mathbb{R}$. Because $p_i$ and $p_j$ are both probability vectors, their difference satisfies $\mathbf{1}^\top\delta = \sum_k \delta_k = 0$, and hence the first-order term vanishes:
\[
  \nabla_q\mathrm{D}(p \Vert q)\big|_{q=p}^{\top}\delta = \mu\mathbf{1}^\top\delta = 0.
\]

\emph{Second order.} By the definition of $G_{\mathrm{D}}$,
\begin{equation}
  \frac{1}{2}\delta^\top\nabla_q^2\mathrm{D}(p \Vert q)\big|_{q=p}\delta = \delta^\top G_{\mathrm{D}}(p)\delta = (p_i-p_j)^\top G_{\mathrm{D}}(p_i)(p_i-p_j),
\end{equation}
with the last equality holding because the quadratic form is even in $\delta$. Since $q=p$ minimizes $\mathrm{D}(p \Vert \cdot)$ over the simplex, the second-order necessary condition gives $\delta^\top G_{\mathrm{D}}(p)\delta \ge 0$ for every admissible difference $\delta$ (those with $\sum_k\delta_k=0$); that is, $G_{\mathrm{D}}(p)$ is positive semidefinite on the tangent space of the simplex, though not necessarily positive definite as a matrix.

\emph{Remainder.} By Taylor's theorem with Lagrange remainder, $\rho_{ij} = \tfrac16\nabla_q^3\mathrm{D}(p\Vert q)\big|_{q=\xi}[\delta,\delta,\delta]$ for some $\xi$ on the segment between $p_i$ and $p_j$. Both endpoints lie in $\Delta_\varepsilon$, which is convex, so $\xi\in\Delta_\varepsilon$ and $|\rho_{ij}|\le\tfrac16 M_3(\varepsilon)\|\delta\|_2^3$. For log-based divergences the third derivatives grow as a coordinate approaches zero, which is why the floor is required. Summing over edges with $W_{ij}\ge0$ gives the bound on $\rho$. \hfill $\square$

\subsection{Predictive Smoothness}

\begin{definition}[Quadratic total-variation domination]
\label{def:tv-domination}
For $p,q\in\Delta^{K-1}$, let $\operatorname{TV}(p,q):=\tfrac{1}{2}\|p-q\|_1$ denote the total variation distance. A divergence $\mathrm{D}$ on $\Delta^{K-1}$ \emph{quadratically dominates total variation} with constant $C_{\mathrm{D}}>0$ if
\begin{equation}
  \mathrm{D}(p \Vert q) \;\geq\; 4\,C_{\mathrm{D}}\operatorname{TV}(p,q)^2 \qquad \forall\, p,q\in\Delta^{K-1},
\end{equation}
equivalently $\mathrm{D}(p \Vert q) \geq C_{\mathrm{D}}\|p-q\|_1^2$.
\end{definition}

This is a Pinsker-type condition requiring $\mathrm{D}$ to grow at least quadratically in total variation, which rules out divergences that vanish faster than $\|p-q\|_1^2$ near the diagonal.

\begin{lemma}[Predictive-distribution smoothness]
\label{lem:predictive-smoothness}
Let $\mathrm{D}$ dominate squared total variation with constant $C_{\mathrm{D}}$ according to Definition~\ref{def:tv-domination}. Then, for every edge $(i,j)\in E$,
\begin{equation}
\|p_i-p_j\|_1 \leq \frac{1}{\sqrt{C_{\mathrm{D}}}}\sqrt{\mathrm{D}(p_i \Vert p_j)}
\end{equation}
\end{lemma}

\paragraph{Proof.}
By Definition~\ref{def:tv-domination}, $\mathrm{D}(p_i \Vert p_j) \geq C_{\mathrm{D}}\|p_i-p_j\|_1^2$. Since both sides are non-negative, taking square roots and rearranging gives
\begin{equation}
    \|p_i-p_j\|_1 \leq \frac{1}{\sqrt{C_{\mathrm{D}}}} \sqrt{\mathrm{D}(p_i \Vert p_j)}
\end{equation} \hfill $\square$

\subsection{Confidence Smoothness}

\begin{proposition}[Confidence Smoothness]
\label{cor:confidence-smoothness}
Under the conditions of Lemma~\ref{lem:predictive-smoothness}, for every edge $(i,j) \in E$,
\begin{equation}
  |c_i - c_j| \leq \tfrac12\|p_i-p_j\|_1 \leq \frac{1}{2\sqrt{C_{\mathrm{D}}}}\sqrt{\mathrm{D}(p_i \Vert p_j)} .
\end{equation}
\end{proposition}

\paragraph{Proof.}
The maximum operator is $1$-Lipschitz with respect to the supremum norm. Let $k_i\in\arg\max_k p_{i,k}$. Then
\begin{equation}
c_i-c_j = p_{i,k_i}-\max_k p_{j,k} \leq p_{i,k_i}-p_{j,k_i} \leq \| p_i- p_j \|_\infty.
\end{equation}
Exchanging the roles of $i$ and $j$ yields $c_j-c_i\leq\|p_j-p_i\|_\infty$, and hence
\begin{equation}
|c_i-c_j| \leq \|p_i-p_j\|_\infty
\end{equation}
Since $p_i,p_j\in\Delta^{K-1}$, we have
\begin{equation}
    \sum_{k=1}^{K}(p_{i,k}-p_{j,k}) = \sum_{k=1}^{K}p_{i,k} - \sum_{k=1}^{K}p_{j,k} = 0.
\end{equation}

Let $k^\star \in \arg\max_k |p_{i,k}-p_{j,k}|$. Then $\sum_{k\neq k^\star}(p_{i,k}-p_{j,k}) = - (p_{i,k^\star}-p_{j,k^\star})$ and we have:
\begin{align*}
\sum_{k\neq k^\star}|p_{i,k}-p_{j,k}| &\geq \left|\sum_{k\neq k^\star}(p_{i,k}-p_{j,k}) \right| = |p_{i,k^\star}-p_{j,k^\star}|\\ \Longrightarrow
\|p_i-p_j\|_1 &= |p_{i,k^\star}-p_{j,k^\star}| +
\sum_{k\neq k^\star}|p_{i,k}-p_{j,k}|\\ &\geq 2|p_{i,k^\star}-p_{j,k^\star}| = 2\|p_i-p_j\|_\infty.
\end{align*}
and thus
\begin{equation}
\|p_i -p_j\|_\infty \leq \frac{1}{2}\|p_i - p_j\|_1
\end{equation}

Combining these inequalities with Lemma~\ref{lem:predictive-smoothness} gives
\begin{equation}
   |c_i-c_j| \leq \|p_i - p_j\|_\infty \leq \frac{1}{2}\|p_i - p_j\|_1 \leq \frac{1}{2\sqrt{C_{\mathrm{D}}}}
\sqrt{\mathrm{D}(p_i \Vert p_j)},
\end{equation} \hfill $\square$

\subsection{Local Disagreement Bound}

\begin{proposition}[Local Disagreement Bound]
\label{lem:markov-bound}
Let $\mathcal{L}(\theta) = \mathcal{L}_{\mathrm{task}}(\theta) + \lambda R_n(p)$ with $\mathcal{L}_{\mathrm{task}} \geq 0$, and sample an edge $(i,j)$ with probability $W_{ij}/|E|$, where $|E| = \sum_{i,j} W_{ij}$. Then for any $t>0$,
\begin{equation}
  \Pr\left(\mathrm{D}(p_i \Vert p_j) \geq t\right) \leq \frac{R_n(p)}{|E|\,t} \leq \frac{\mathcal{L}(\theta)}{\lambda\,|E|\,t}
\end{equation}
\end{proposition}

\paragraph{Proof.}
Since $\mathcal{L}_{\mathrm{task}}(\theta) \geq 0$, rearranging $\mathcal{L}(\theta) = \mathcal{L}_{\mathrm{task}}(\theta) + \lambda R_n(p)$ gives
\begin{equation}
  R_n(p) = \frac{\mathcal{L}(\theta) - \mathcal{L}_{\mathrm{task}}(\theta)}{\lambda}
  \;\leq\; \frac{\mathcal{L}(\theta)}{\lambda}.
\end{equation}
Let $X := \mathrm{D}(p_i \Vert p_j)$ for an edge sampled with probability $W_{ij}/|E|$. Since $\mathrm{D} \geq 0$, $X$ is a nonnegative random variable with
\begin{equation}
  \mathbb{E}[X] = \frac{1}{|E|}\sum_{i,j} W_{ij}\,\mathrm{D}(p_i \Vert p_j)
  = \frac{R_n(p)}{|E|} \;\leq\; \frac{\mathcal{L}(\theta)}{\lambda\,|E|}.
\end{equation}
Markov's inequality then gives, for any $t>0$,
\begin{equation}
    \Pr(X \geq t) \leq \frac{\mathbb{E}[X]}{t} = \frac{R_n(p)}{|E|t} \leq \frac{\mathcal{L}(\theta)}{\lambda|E|t}
\end{equation}

\hfill$\square$

\subsection{Probabilistic Confidence Smoothness Bound}

\begin{corollary} [Probabilistic Confidence Smoothness]
\label{cor:prob-confidence-smoothness}
Under the conditions of Proposition~\ref{cor:confidence-smoothness} and Proposition~\ref{lem:markov-bound}, for any $t>0$, with probability at least $1 - \frac{\mathcal{L}(\theta)}{\lambda|E|t}$ over a sampled edge $(i,j)$,
\begin{equation}
  |c_i - c_j| \;\leq\; \frac{1}{2}\sqrt{\frac{t}{C_{\mathrm{D}}}}
\end{equation}
\end{corollary}

\paragraph{Proof.}
By Proposition~\ref{lem:markov-bound}, the event $\{\mathrm{D}(p_i \Vert p_j) < t\}$ holds with probability at least $1 - \mathcal{L}(\theta)/(\lambda|E|t)$. On this event, Proposition~\ref{cor:confidence-smoothness} gives
\begin{equation}
    |c_i - c_j| \leq \frac{1}{2\sqrt{C_{\mathrm{D}}}}\sqrt{\mathrm{D}(p_i \Vert p_j)} < \frac{1}{2}\sqrt{t/C_{\mathrm{D}}}
\end{equation}
\hfill$\square$

\subsection{Proof of Proposition~\ref{cor:propagation}}
By the triangle inequality,
\begin{equation}
  |a_j - c_j| \;\leq\; |a_j - a_i| + |a_i - c_i| + |c_i - c_j| .
\end{equation}
The second term is $\mathcal{E}_{\mathrm{cal}_i}$ and the third is bounded by Proposition~\ref{cor:confidence-smoothness}. \hfill$\square$

\subsection{Proof of Theorem~\ref{thm:pathwise-propagation}}

By the triangle inequality applied at the endpoints of $P$,
\begin{equation}
  \mathcal{E}_{\mathrm{cal}_v} = |a_v - c_v| \leq |a_v - a_s| + |a_s - c_s| + |c_s - c_v| = \mathcal{E}_{\mathrm{cal}_s} + |c_s - c_v| + |a_v - a_s|
\end{equation}
Telescoping the confidence term along $P$ and applying Proposition~\ref{cor:confidence-smoothness} to each edge,
\begin{equation}
  |c_s - c_v| \;\leq\; \sum_{\ell=1}^{r}\bigl|c_{v_{\ell-1}} - c_{v_\ell}\bigr| \;\leq\; \frac{1}{2\sqrt{C_{\mathrm{D}}}}\sum_{\ell=1}^{r}\sqrt{D_\ell}.
\end{equation}
By Cauchy--Schwarz, $\sum_{\ell=1}^{r}\sqrt{D_\ell} \leq \sqrt{r\sum_{\ell=1}^{r}D_\ell}$, so $|c_s-c_v| \leq \sqrt{\tfrac{r}{4C_{\mathrm{D}}}\sum_{\ell}D_\ell}$. Substituting yields the stated bound. \hfill$\square$

\subsection{Proof of Proposition~\ref{prop:obstruction}}
By the triangle inequality,
\begin{equation}
  |a_i - a_j| \;\leq\; |a_i - c_i| + |c_i - c_j| + |c_j - a_j| = \mathcal{E}_{\mathrm{cal}_i} + |c_i - c_j| + \mathcal{E}_{\mathrm{cal}_j}
\end{equation}
Rearranging gives
\begin{equation}
  \mathcal{E}_{\mathrm{cal}_i} + \mathcal{E}_{\mathrm{cal}_j} \geq |a_i - a_j| - |c_i - c_j| .
\end{equation}
Substituting $|c_i - c_j| \leq \frac{1}{2\sqrt{C_{\mathrm{D}}}}\sqrt{\mathrm{D}(p_i \Vert p_j)}$ from Proposition~\ref{cor:confidence-smoothness} yields~\eqref{eq:obstruction-D}. \hfill$\square$

Perfect calibration at both endpoints forces $|c_i-c_j|=|a_i-a_j|$ and hence, by Proposition~\ref{cor:confidence-smoothness}, $\mathrm{D}(p_i\Vert p_j)\ge 4C_{\mathrm{D}}(a_i-a_j)^2$: no predictor with divergence below this level can be perfectly calibrated at both endpoints unless $a_i=a_j$.

\subsection{Proposition~\ref{prop:threshold}: Bias--Variance Decomposition and Proof}\label{app:gs:threshold}

Under the fixed-class assumption of Proposition~\ref{prop:threshold}, $a_i$ and $a_j$ are fixed under $\mathcal{T}$. Decompose
\begin{equation}
  c_i = a_i + b_i + \varepsilon_i, \qquad b_i := \mathbb{E}_{\mathcal{T}}[c_i] - a_i, \qquad \mathbb{E}_{\mathcal{T}}[\varepsilon_i] = 0,
\end{equation}
and analogously at $j$, so that the expected squared pointwise calibration error decomposes as $\mathbb{E}_{\mathcal{T}}[(c_i-a_i)^2] = b_i^2 + \mathrm{Var}(\varepsilon_i)$ into a systematic and a random part. Hence the right-hand side of~\eqref{eq:threshold} is
\begin{equation}
  \mathbb{E}_{\mathcal{T}}\bigl[(e_i - e_j)^2\bigr] = (b_i - b_j)^2 + \mathrm{Var}(\varepsilon_i - \varepsilon_j) .
\end{equation}
In particular, if $b_i=b_j$, $\mathrm{Var}(\varepsilon_i)=\mathrm{Var}(\varepsilon_j)=\sigma^2$ and $\mathrm{Corr}(\varepsilon_i,\varepsilon_j)=\rho$, the condition is $|a_i-a_j|<\sigma\sqrt{2(1-\rho)}$.

\paragraph{Proof.}
Write $\Delta := a_i - a_j$ and $\bar{c} := \tfrac12(c_i + c_j)$, with $c_i = p_{i,k^\ast}$ and $c_j = p_{j,k^\ast}$. For the mixture $m = \tfrac12(p_i+p_j)$ we have $m_{k^\ast} = \bar{c}$, while for every coordinate $k$, $m_k = \tfrac12(p_{i,k}+p_{j,k}) \le \bar{c}$. Hence $\max_k m_k = \bar{c}$, the predicted class is preserved, and $a_i, a_j$ are unchanged. Before averaging, the summed expected squared calibration error is $\mathbb{E}[e_i^2]+\mathbb{E}[e_j^2]$. After, both endpoints receive $\bar{c}$, with errors $\bar{c}-a_i = \tfrac12(e_i+e_j) - \tfrac12\Delta$ and $\bar{c}-a_j = \tfrac12(e_i+e_j)+\tfrac12\Delta$, whose squares sum to $\tfrac12(e_i+e_j)^2 + \tfrac12\Delta^2$. Taking expectations and subtracting, the change is
\begin{equation}
  \tfrac12\Delta^2 + \tfrac12\mathbb{E}[(e_i+e_j)^2] - \mathbb{E}[e_i^2] - \mathbb{E}[e_j^2]
  = \tfrac12\Delta^2 - \tfrac12\mathbb{E}[(e_i-e_j)^2],
\end{equation}
strictly negative if and only if $\Delta^2 < \mathbb{E}[(e_i-e_j)^2]$. Since $\mathbb{E}[\varepsilon_i]=\mathbb{E}[\varepsilon_j]=0$, $\mathbb{E}[(e_i-e_j)^2] = (b_i-b_j)^2 + \mathrm{Var}(\varepsilon_i-\varepsilon_j)$, which equals $2\sigma^2(1-\rho)$ under the symmetric assumptions. \hfill$\square$

\paragraph{Interpretation.}

For fixed predicted labels, Proposition~\ref{prop:obstruction} limits how small the summed edge calibration error can be at a given predictive divergence. Under Proposition~\ref{prop:threshold}'s assumptions, pairwise averaging removes $\tfrac12\mathbb E[(e_i-e_j)^2]$ from summed expected squared error while adding $(a_i-a_j)^2/2$ due to the correctness gap. Improvement occurs exactly when the former exceeds the latter.

Under the equal-bias, equal-variance specialization, the tolerable correctness gap scales with $\sqrt{1-\rho}$ and vanishes at $\rho=1$, so predictions that err together cannot correct one another. Since neighboring predictions from a single network are plausibly positively correlated, $\sqrt{2}\sigma$ is a ceiling rather than an estimate of the operative threshold, and whatever benefit smoothing confers is a reduction of within-neighborhood error dispersion, not a correction of the neighborhood's mean error.

The threshold depends on the training stage through $\mathbb E_{\mathcal T}[(e_i-e_j)^2]$, which includes systematic differences between neighboring errors as well as random variation. If this quantity decreases, the tolerable correctness gap narrows, motivating weaker smoothing later in training.

\section{\modelNoSpace-Specific Guarantees} \label{app:specific-proofs}

Here we instantiate the results of Section~\ref{sec:gs:theory}. For base-$2$ JS divergence, the domination constant is $C_{\mathrm{JS}}=1/(8\ln 2)$. Substituting this constant into the general bounds yields the following guarantees.

\begin{corollary}[Confidence smoothness and propagation under \model]
\label{cor:js-confidence}
For every edge $(i,j) \in E$,
\begin{equation}
  |c_i - c_j| \leq \sqrt{2\ln(2)\mathrm{JS}(p_i \Vert p_j)}
\end{equation}
and consequently
\begin{equation}
  \mathcal{E}_{\mathrm{cal}_j} \leq \mathcal{E}_{\mathrm{cal}_i} + \sqrt{2\ln(2)\,\mathrm{JS}(p_i \Vert p_j)} + |a_i - a_j| .
\end{equation}
Along any simple directed path $P = (v_0,\ldots,v_r)$ in $\mathcal{G}$,
\begin{equation}
    \mathcal{E}_{\mathrm{cal}_{v_r}} \leq \mathcal{E}_{\mathrm{cal}_{v_0}} + \sqrt{2\ln(2)\,r \sum_{\ell=1}^{r}\mathrm{JS}(p_{v_{\ell-1}}\Vert p_{v_\ell})} + \bigl|a_{v_r}-a_{v_0}\bigr| .
\end{equation}
\end{corollary}

\paragraph{Proof.} Write $m = (p+q)/2$, so that $\mathrm{JS}_{\ln}(p\Vert q) = \tfrac12 \KL(p\Vert m) + \tfrac12 \KL(q\Vert m)$. Since $p - m = \tfrac12(p-q)$ and $q - m = -\tfrac12(p-q)$, we have $\|p-m\|_1 = \|q-m\|_1 = \tfrac12\|p-q\|_1$. Pinsker's inequality $\KL(u\Vert v) \geq \tfrac12\|u-v\|_1^2$ (in nats) then gives
\begin{equation}
  \KL(p\Vert m),\; \KL(q\Vert m) \geq \tfrac12\cdot\tfrac14\|p-q\|_1^2 = \tfrac18\|p-q\|_1^2 
\end{equation}
thus $\mathrm{JS}_{\ln}(p\Vert q) \geq \tfrac18\|p-q\|_1^2$. Converting to base $2$ via $\mathrm{JS} = \mathrm{JS}_{\ln}/\ln 2$ yields $\mathrm{JS}(p\Vert q) \geq \frac{1}{8\ln 2}\|p-q\|_1^2$, i.e. $C_{\mathrm{JS}} = 1/(8\ln 2)$. Substituting into Proposition~\ref{cor:confidence-smoothness} gives $\frac{1}{2\sqrt{C_{\mathrm{JS}}}} = \tfrac12\sqrt{8\ln 2} = \sqrt{2\ln 2}$, which is the stated confidence bound.

The remaining claims follow by the same substitution. In Proposition~\ref{cor:propagation}, $\tfrac{1}{2}C_{\mathrm{D}}^{-1/2} \sqrt{\mathrm{D}(p_i \Vert p_j)}$ becomes $\sqrt{2\ln(2)\,\mathrm{JS}(p_i\Vert p_j)}$; in Theorem~\ref{thm:pathwise-propagation}, $\tfrac{r}{4C_{\mathrm{D}}}$ becomes
$2\ln(2)r$ inside the first radical; and in Proposition~\ref{prop:obstruction} the same replacement gives the \modelNoSpace-specific obstruction. \hfill$\square$

Proposition~\ref{lem:markov-bound} bounds large disagreement at a fixed predictor: for an edge sampled with probability $W_{ij}/|E|$ and $t>0$, $\Pr(\mathrm{JS}(p_i\Vert p_j)\geq t)\leq\mathcal L(\theta)/(\lambda|E|t)$. Instantiating Proposition~\ref{prop:obstruction} gives
\begin{equation}
  \mathcal{E}_{\mathrm{cal}_i} + \mathcal{E}_{\mathrm{cal}_j} \geq |a_i - a_j| - \sqrt{2\ln(2)\,\mathrm{JS}(p_i \Vert p_j)}
\end{equation}

Taken together, these delimit what \model can and cannot achieve. The penalty $\mathrm{JS}(p_i\Vert p_j)$ appears in every bound as the quantity the
regularizer directly controls, and $\lambda$ governs how strongly it is suppressed. The residual $|a_i - a_j|$ appears in each with the opposite sign; it is not penalized, and is determined by which points the frozen graph connects and by the predicted labels. \model is therefore expected to improve calibration to the extent that its graph connects points which genuinely share a true probability of correctness - a condition the design choices of Section~\ref{sec:nescal:design} are intended to promote and which Section~\ref{sec:results} examines empirically.

\subsection{JS-Specific Laplacian Approximation Lemma}

\begin{corollary}[JS-induced weighting]
\label{lem:js-laplacian}
For $\mathrm{D}=\mathrm{JS}$ (base $2$), under the probability floor $\min_{i,k}p_{i,k}\ge\varepsilon$ of Proposition~\ref{lem:laplacian}, the induced local form is diagonal,
\begin{equation}
  G_{\mathrm{JS}}(p_i) = H(p_i) = \mathrm{diag}\left(\frac{1}{8\ln(2)\,p_{i,1}}, \ldots, \frac{1}{8\ln(2)\,p_{i,K}}\right)
\end{equation}
which is positive definite since $p_{i,k}>0$ for softmax outputs. Consequently,
\begin{equation}
    R_n(p) = \sum_{i,j} W_{ij}\,(p_i-p_j)^{\mathsf{T}} H(p_i)(p_i-p_j) + \rho, \qquad |\rho| \le \frac{1}{12\ln 2\,\varepsilon^{2}}\sum_{i,j}W_{ij}\|p_i-p_j\|_2^{3}.
\end{equation}
\end{corollary}

\paragraph{Proof.}
The $\mathrm{JS}$ divergence satisfies $\mathrm{JS}(p\Vert p)=0$, and its formula is smooth on the positive orthant with third derivatives jointly continuous in $(p,q)$, so under the probability floor of Proposition~\ref{lem:laplacian} its conditions hold. By Proposition~\ref{lem:laplacian}, the zeroth- and first-order terms of the expansion of $\mathrm{JS}(p_i \Vert p_j)$ about $p_i=p_j$ vanish, and the weighting is determined by $G_{\mathrm{JS}}(p)=\tfrac12\nabla_q^2\mathrm{JS}(p \Vert q)|_{q=p}$. It therefore suffices to extract the second-order coefficient. Write $\delta_k=q_k-p_k$ and $m_k=p_k+\tfrac{\delta_k}{2}$, and expand $\mathrm{JS}$ in natural-log units,
\begin{align*}
      \mathrm{JS_{\ln}}(p \Vert q) &= \sum_k \frac{1}{2}\Bigl[ p_k \ln p_k + q_k \ln q_k - (p_k + q_k)\ln m_k \Bigr] \\ &= \sum_k \frac{1}{2}\Bigl[p_k \ln p_k  +  (p_k + \delta_k) \ln (p_k + \delta_k) - (2p_k + \delta_k)\ln( p_k + \frac{\delta_k}{2})\Bigr]  
\end{align*}
Using the second-order Taylor expansions
\begin{align*}
  \ln(p_k+\delta_k) &= \ln p_k + \tfrac{\delta_k}{p_k} - \tfrac{\delta_k^2}{2p_k^2} + O(\delta^3),\\
  \ln\left(p_k+\tfrac{\delta_k}{2}\right) &= \ln p_k + \tfrac{\delta_k}{2p_k} - \tfrac{\delta_k^2}{8p_k^2} + O(\delta^3),
\end{align*}
and retaining only the second-order terms (the lower orders cancel by Proposition~\ref{lem:laplacian}) gives
\begin{equation}
  \sum_k \tfrac{1}{2}\left[ \tfrac{\delta_k^2}{p_k} - \tfrac{\delta_k^2}{2p_k} + \tfrac{\delta_k^2}{4p_k} - \tfrac{\delta_k^2}{2p_k} \right] = \sum_k \frac{\delta_k^2}{8 p_k}.
\end{equation}

Hence $G_{\mathrm{JS,\ln}}(p)=\mathrm{diag}\left(1/(8p_k)\right)$. For the remainder, $\mathrm{JS}_{\ln}(p\Vert q)=\sum_k\varphi_k(q_k)$ is separable, with $\varphi_k'''(q)=\tfrac12\bigl(-q^{-2}+\tfrac14 m^{-2}\bigr)$, $m=(p_k+q)/2$, so $|\varphi_k'''|\le 1/(2\varepsilon^2)$ on the floor; since $\sum_k|u_k|^3\le\|u\|_2^3$, $M_3(\varepsilon)\le 1/(2\varepsilon^2)$ in natural-log units, and Proposition~\ref{lem:laplacian} gives $|\rho_{ij}|\le\|p_i-p_j\|_2^3/(12\varepsilon^2)$, which becomes $\|p_i-p_j\|_2^3/(12\ln 2\,\varepsilon^2)$ in base $2$. Converting to base $2$ via the exact identity $\mathrm{JS}=\mathrm{JS}_{\ln}/\ln 2$ (each $\KL$ term rescales identically under change of base) gives $H(p)=\mathrm{diag}\left(1/(8\ln(2)\,p_k)\right)$, and substituting $\delta_k=p_{i,k}-p_{j,k}$ yields the matrix form. \hfill$\square$

Up to the constant $1/(8\ln 2)$, $H(p_i)$ is the Fisher information metric of the categorical distribution~\citep{amari_200_fisher}, which every $f$-divergence induces up to $\tfrac12 f''(1)$. Penalizing $\mathrm{JS}$ therefore measures disagreement in the intrinsic geometry of the simplex rather than the Euclidean geometry of the probability vector. The weighting is a property of the geometry rather than a performance guarantee, and it holds only where the expansion is accurate, $|p_{i,k}-p_{j,k}|\ll p_{i,k}$.

\section{\model Implementation and Computational Cost}\label{app:nescal:complexity}

\subsection{Objective Normalization}\label{app:nescal:normalization}

The objective in Section~\ref{sec:nescal:reg} uses the summed penalty $R_n(p)$; optimization uses the directed-edge mean $\bar R_n(p)=R_n(p)/|E|$. This rescales the coefficient: $\lambda_{\mathrm{mean}}=|E|\lambda_{\mathrm{sum}}$. Algorithm~\ref{alg:nescal} and all reported regularization strengths use the edge-mean convention.

\subsection{Training Procedure and Cost}
\begin{algorithm}[t]
\caption{\model}
\label{alg:nescal}
\begin{algorithmic}[1]

\Require Training set $\mathcal{D}=\{(x_i,y_i)\}_{i=1}^n$; model $f_\theta$ with representation map $h$; number of neighbors $k$; regularization strength $\lambda_0$; activation epoch $T_a$; total epochs $T$

\State Initialize model parameters $\theta$

\For{$t = 1$ \textbf{to} $T$}

    \If{$t = T_a$}
        \State Compute $z_i \gets h(x_i)$ for all $i$
            \Comment{no gradient}
        \State $W \gets \textsc{kNNGraph}(\{z_i\}_{i=1}^n,k,\mathrm{cosine})$
            \Comment{fixed thereafter}
\State $\widehat{p}_i \gets \operatorname{stopgrad}(f_\theta(x_i))$ for all $i$
            \Comment{initialize cache}
    \EndIf

    \If{$t \ge T_a$}
        \State $\displaystyle
        \lambda_t \gets
        \frac{\lambda_0}{2}
        \left[
        1+\cos\left(
        \pi\frac{t-T_a}{T-T_a}
        \right)
        \right]$
    \Else
        \State $\lambda_t \gets 0$
    \EndIf

    \ForAll{minibatches $B \subseteq \mathcal{D}$}

        \State $p_i \gets f_\theta(x_i)$ for all $i \in B$

        \State $\displaystyle
        \mathcal{L}_{\mathrm{CE}}
        \gets
        \frac{1}{|B|}
        \sum_{i\in B}
        \mathrm{CE}(p_i,y_i)$

        \If{$t \ge T_a$}
            \State $\widehat{p}_i \gets \operatorname{stopgrad}(p_i)$
                for all $i\in B$
                \Comment{current targets for neighbors in this batch}
            \State $\displaystyle
            E_B \gets
            \{(i,j): i\in B,\; W_{ij}=1\}$

            \State $\displaystyle
            R \gets
            \frac{1}{|E_B|}
            \sum_{(i,j)\in E_B}
            \mathrm{JS}
            \!\left(
            p_i \,\Vert\, \widehat{p}_j
            \right)$
            \Comment{$\widehat{p}_j$ detached}
        \Else
            \State $R \gets 0$
        \EndIf

        \State $\displaystyle
        \theta \gets
        \theta
        -
        \alpha\nabla_\theta
        \left(
        \mathcal{L}_{\mathrm{CE}}+\lambda_t R
        \right)$

    \EndFor
\EndFor

\end{algorithmic}
\end{algorithm}

Algorithm~\ref{alg:nescal} summarizes the training procedure. Training proceeds by standard cross-entropy minimization until the activation epoch $T_a$, at which point a single inference pass supplies the embeddings from which the $k$-nearest-neighbor graph is constructed, and a second seeds a cache of predictive distributions for every training sample. From $T_a$ onward each minibatch incurs the neighborhood-consistency penalty in addition to the task loss, with $\lambda$ annealed over the remaining epochs and the cache refreshed with the predictions computed in that step. The graph itself is never rebuilt.

Computing live predictions for every neighbor would require additional network evaluations. Neighbor targets are therefore detached current-step predictions when the neighbor belongs to the minibatch, and cached predictions otherwise. Each cached prediction is refreshed when its sample is visited, once per epoch. The predictive bounds are evaluated using current predictions, as described in Appendix~\ref{app:theory-audit}.

The graph construction is a one-off cost. The $k$-nearest-neighbor search is exact, computed as a chunked cosine similarity, and is therefore $O(n^2 l)$ in the number of samples $n$ and the embedding dimension $l$. Because the graph is built once and held fixed (Section~\ref{sec:nescal:design}), this cost is incurred a single time and amortized over the remaining $T - T_a$ epochs.

The recurring cost is the penalty itself. It requires one divergence evaluation per edge, each $O(K)$ in the number of classes, giving $O(nkK)$ per epoch against the $O(nC)$ cost of the forward and backward passes, where $C$ is the per-sample cost of the network. Since $C \gg kK$ for most of the architectures considered in Section~\ref{sec:exp-design}, the relative overhead is small. Memory overhead is $O(nk)$ for the edge list, held on host memory, and $O(nK)$ for the cached predictions. Measured graph-construction times, peak memory, and training wall-clock relative to the baselines are reported in Appendix~\ref{app:cost}.

\subsection{Directed Neighborhoods}

Because the $k$-nearest-neighbor relation is not symmetric ($j$ may be among the $k$ nearest neighbors of $i$ without the opposite being true), $\mathcal{G}$ is a directed graph, and $W$ is not in general symmetric. Nothing in Section~\ref{sec:graph-smoothing} requires symmetry. Larger neighborhoods can reduce averaging variance while admitting more heterogeneous points.

\subsection{Activation and Schedule Rationale}\label{app:nescal:activation}

Activation epoch and $\lambda$ are selected jointly by validation NLL; the theory motivates the schedule without determining it. Under Proposition~\ref{prop:threshold}'s assumptions, a decrease in $\mathbb E[(e_i-e_j)^2]$ narrows the correctness gap that averaging can tolerate, motivating decay. Proposition~\ref{prop:obstruction} also motivates delaying graph construction until neighborhoods are informative: smoothing across a large correctness gap can preclude calibration at both endpoints. Delaying may stabilize predicted labels, which fixes the corresponding correctness probabilities. Activation balances these considerations.

\model activates after a warm-up and then decays $\lambda$ on a cosine schedule. Appendix~\ref{app:ablations} compares constant and cosine schedules; Appendix~\ref{app:activation} evaluates diagnostic activation rules and their transfer across settings.

\section{Training and Implementation Details} \label{app:train-details}

Unless stated otherwise, every method in a given setting shares the architecture, optimizer, schedule, epoch budget, batch size, data splits and seeds; only the training objective differs.

\subsection{Datasets, Splits and Preprocessing}

\begin{table}[htpb]
\centering
\caption{Datasets. Splits are train/val/test}
\label{tab:app-datasets}
\small
\begin{tabular}{llrrl}
\toprule
Dataset & Modality & Classes & Size & Split \\
\midrule
CIFAR-10 / CIFAR-100 & image & 10 / 100 & 60{,}000 & 45k / 5k / 10k \\
Tiny-ImageNet & image & 200 & 110{,}000 & 90k / 10k / 10k \\
20 Newsgroups & text & 20 & 18{,}846 & 13{,}570 / 1{,}507 / 3{,}769 \\
Amazon (binary) (frozen RoBERTa) & text & 2 & 11{,}349 & 7{,}264 / 1{,}816 /  2{,}269 \\
Amazon (5-class) (frozen RoBERTa) & text & 5 & 22{,}815 & 14{,}602 / 3{,}650 / 4{,}563 \\
Helena & tabular & 100 & 65{,}196 & 41{,}724 / 10{,}432 / 13{,}040 \\
Jannis & tabular & 4 & 83{,}733 & 53{,}589 / 13{,}397 / 16{,}747 \\
Covertype (subsampled) & tabular & 7 & 100{,}000 & 64{,}000 / 16{,}000 / 20{,}000 \\
\bottomrule
\end{tabular}
\end{table}

\textbf{Image.} Standard augmentation (random crop with 4-pixel padding, random horizontal flip) and per-channel normalization; evaluation uses normalization only. The validation split is a seeded $10\%$ partition of the training set, fixed across seeds.

\textbf{Tabular.} Following~\citet{gorishniy_2021_tabular}, features are quantile-transformed to a normal output distribution, except for Helena, which is standardized. The transform is fitted on the training partition only. The test and validation splits are fixed across seeds.

\textbf{Text.} The 20 Newsgroups pipeline follows the MMCE reference implementation: vocabulary of $20{,}000$, sequence length $1{,}000$, GloVe-100d initialization~\citep{pennington_2014_glove}. Frozen-encoder embeddings are mean-pooled over tokens with the attention mask and extracted once, so the head sees a fixed vector dataset. The test split is fixed across seeds in every setting; the validation split is redrawn per seed on the 20NG setting, and for the frozen-encoder Amazon settings it is held fixed so that the same encoder embeddings are reused.

\subsection{Architectures and Optimization}

\begin{table}[htpb]
\centering
\caption{Training configuration per setting.}
\label{tab:app-optim}
\small
\begin{tabular}{llllll}
\toprule
Setting & Optimizer & LR & LR schedule & Epochs & Batch \\
\midrule
CIFAR-10/100, ResNet-50 \& WRN-28-10 & SGD ($0.9$, wd $5\!\times\!10^{-4}$) & $0.1$ & $\times 0.1$ at $50\%,75\%$ & 200 & 128 \\
Tiny-ImageNet, ResNet-50 & SGD ($0.9$, wd $5\!\times\!10^{-4}$) & $0.1$ & $\times 0.1$ at $40\%,60\%$ & 100 & 64 \\
20 Newsgroups, GPCNN & Adam & $10^{-3}$ & constant & 50 & 128 \\
Frozen encoder + MLP head & SGD ($0.9$, wd $10^{-4}$) & $10^{-3}$ & constant & 50 (early stopping) & 256 \\
Tabular, MLP $3\times256$ (dropout $0.1$) & AdamW (wd $10^{-5}$) & $10^{-3}$ & constant & 100 & 512 \\
\bottomrule
\end{tabular}
\end{table}

Image runs use mixed-precision training; all others use full precision. Tabular and text settings use a constant learning rate by design, which removes any interaction between the $\lambda$ schedule and LR decay - a useful control for the schedule comparison.

\subsection{\model Configuration and Selection}

Unless ablated, $k=10$, cosine distance, JS divergence, a directed unweighted kNN graph over penultimate features, frozen at the activation epoch. The penalty uses the directed-edge mean $\bar R_n=R_n/|E|$, removing its direct scaling with the number of edges. Cached neighbor predictions are refreshed when their samples are visited (Appendix~\ref{app:nescal:complexity}). All reported regularization strengths use the edge-mean convention (Appendix~\ref{app:nescal:normalization}).

Activation epoch and $\lambda$ are selected jointly. A single cross-entropy warm-up run per seed is checkpointed at every candidate onset; each grid point resumes from the corresponding checkpoint, so all branches share a bit-identical prefix and differ only in the intervention under study. Resumption restores optimizer state and the scheduler's absolute epoch. This checkpoint-resumption setup enables a wider grid search space at lower cost. Within each schedule, activation epoch and $\lambda$ are selected by mean validation NLL over seeds $18$ and $42$. Evaluation of the selected configuration is on seeds $18, 42, 613$ and $2026$. The same selection-evaluation protocol applies to all methods, with the following exceptions: for 20NG, CE and \model additionally include evaluation on seeds $7$ and $1234$; s-AvUC for Tiny-ImageNet uses three evaluation seeds, excluding $2026$. Table~\ref{tab:app-nescal-grid} gives the search grids. Divergent configurations are excluded from selection; Helena omits $\lambda=0.5$.

\begin{table}[htpb]
\centering
\caption{\modelNoSpace\ search grids. Onsets are epochs of the total budget.}
\label{tab:app-nescal-grid}
\small
\begin{tabular}{lll}
\toprule
Setting & Candidate onsets & $\lambda$ grid (cosine / constant) \\
\midrule
CIFAR-10, ResNet-50 & $\{50,65,80,95,110,125\}$ & $\{1,3,5,10,20\}$ / $\{1,3,5\}$ \\
CIFAR-10, WRN-28-10 & $\{50,65,80,95,110,125\}$ & $\{5,10,20,30,50\}$ / $\{3,5,10\}$ \\
CIFAR-100, ResNet-50 & $\{50,65,80,95,100,110,125\}$ & $\{5,10,20,30,50\}$ / $\{3,5,8,10\}$ \\
CIFAR-100, WRN-28-10 & $\{50,65,80,95,100,110,125\}$ & $\{5,10,20,30,40\}$ / $\{3,5,8,10\}$ \\
Tiny-ImageNet & $\{20,30,38,40,50,58\}$ & $\{5,10,20,30,40\}$ / $\{5,10,20\}$ \\
20NG & $\{3,5,8,12,20\}$ & $\{0.5,1,2,5\}$ / $\{0.5,1,2,5\}$ \\
Tabular & $\{5,10,20,30,50\}$ & $\{0.5,1,2,5,10\}$ / $\{1,2,5\}$ \\
Frozen encoder & $\{1,3,5,10,20,30\}$ & $\{0.01,0.05,0.1,0.3,0.5,0.7,1.0,2.0\}$ / constant identical \\
\bottomrule
\end{tabular}
\end{table}

\subsection{Competitor Search Spaces}
Each competitor is tuned over setting-dependent candidate ranges informed by the search ranges or parameter-sensitivity analyses reported in its original paper. Configurations are selected by seed-averaged validation NLL, using the same selection criterion as \modelNoSpace. For compactness, the values below report the widest range of candidate values considered for each method across settings; individual settings may use subsets of these values, dependent on author recommendations under parallel settings. We use each method's own notation. FL is searched over fixed $\gamma\in\{2,3\}$ and the adaptive, sample-dependent variant with $\gamma_{\mathrm{high}}\in\{2,3\}$, selecting once across both families. LS uses $\alpha \in \{0.05, 0.1\}$; MMCE $\lambda \in \{1,3,7,10,15,20\}$ at the Laplacian kernel width of~\citet{Kumar_2018_MMCE}, owing to their reported insensitivity to $\lambda$ in that range; MDCA $\beta \in \{1,5,10,15,20,25\}$; MbLS $\lambda \in \{0,0.05,0.1,0.2,0.3\}$ with margin $m \in \{4,6,8,10,12\}$. s-AvUC is searched sequentially in the order $\kappa \in \{0.1,0.5,0.9\}$, temperature $T \in \{0.5,1,2\}$, penalty weight $\beta \in \{0.1,1,3\}$, matching the authors' own recommendation, as a full grid is prohibitive. Note that $\lambda$ denotes the penalty weight for MMCE, MbLS and \modelNoSpace, but the $\ell_2$ coefficient in the notation of~\citet{karandikar_sAVuC_sbECE}.

\subsection{Post-hoc Calibrators}

Post-hoc calibrators are fitted on the validation split used for base-model selection and evaluated on the test split, following the protocol of~\cite{Guo_2107_calibration}. TS uses one-dimensional numerical minimization of validation NLL. Dirichlet calibration uses the full matrix parameterization with off-diagonal and intercept regularization (ODIR), with regularization strength selected on validation. LoCal uses its local calibration objective and early stopping. Feature preprocessing and other departures from the reference methods are described below.

\subsection{Metrics} \label{app:metrics}

The limitations of ECE have motivated debiased and binning-free alternatives~\citep{Arrieta_Ibarra_2022_measuring_calibration, Blasiok_2023_measuring_calibration}. Other notions retain different information: classwise calibration requires $\mathbb{P}(Y=k\mid p_k(X))=p_k(X)$ for each class, while canonical calibration requires $\mathbb{P}(Y=k\mid p(X))=p_k(X)$~\citep{kull_dirichlet_cwece, Vaicenavicius_2019_measuring_calibration, Widmann_2019_measuring_calibration}. Local calibration assesses reliability within neighborhoods in representation space~\citep{luo_lce}.

We report accuracy, ECE~\citep{Naeini_2015_ECE}, an equal-mass-binned ECE variant~\citep{Nixon_2019_measuring_calibration}, classwise ECE~\citep{kull_dirichlet_cwece}, the local calibration error (LCE), using the classwise extension of~\citet{barbera2026multiclass} to the confidence-based measure of~\citet{luo_lce}, and the binned calibration--refinement decomposition of the top-label Brier score~\citep{brier_1950_brier, kull_2015_brier_decomposition}.

ECE and MCE use $15$ equal-width bins of top-label confidence, the last bin closed; the equal-mass variant uses $15$ equal-count bins. Classwise ECE applies the same binning to each class probability and averages over classes. LCE uses a Gaussian kernel ($\gamma=10$) in a $50$-dimensional PCA projection of the penultimate features, fitted on the validation split, excluding bins with at most $20$ samples. The calibration--refinement decomposition bins top-label confidence into the same $15$ bins; because it ignores within-bin variation, its two terms need not sum exactly to the top-label Brier score.

\subsection{Additional Experiments and Ablations}

We ablate the components of Section~\ref{sec:nescal:design}: neighborhood size, random versus nearest-neighbor graphs, graphs built on PCA- and UMAP-reduced representations, and alternative divergences at matched initial penalty magnitude. We also compare regularization schedules and diagnostic activation rules. Further studies assess input corruption, out-of-distribution detection, class imbalance, and label noise. Results appear in Appendix~\ref{app:results}.

\subsection{Theoretical Diagnostics}

The quantities of Sections~\ref{sec:gs:theory} and Appendix~\ref{app:specific-proofs} are computed on the training nodes of the frozen graph, at the selected weights, from a full inference pass with current predictions rather than from the cached distributions used during training. We report whether bounds hold on every edge, the median per-edge tightness of Corollary~\ref{cor:js-confidence} and the fraction of edges satisfying the small-disagreement condition under which the Laplacian expansion of Proposition~\ref{lem:laplacian} is valid. We do not report an aggregate ratio of the Laplacian quadratic form to $R_n$: it is dominated by a small number of edges carrying near-zero class probabilities, where the third-derivative remainder diverges, and is therefore uninformative about the graph as a whole. This is consistent with the lemma being an interpretive device rather than a bound we optimize.

The regime diagnostics of Table~\ref{tab:regime} are logged by the training runs at the activation epoch and averaged over seeds, on the same frozen graph. 

\subsection{Robustness and Perturbation Studies}\label{app:implementation:robust}

Corruption robustness is evaluated on CIFAR-10-C, CIFAR-100-C, and Tiny-ImageNet-C using the 15 standard corruption types (14 on Tiny-ImageNet-C) at all five severity levels~\citep{hendrycks_2019_cifarc}; out-of-distribution (OOD) separation uses SVHN~\citep{netzer_2011_svhn} as the OOD dataset for models trained on CIFAR-10, CIFAR-100, and Tiny-ImageNet. SVHN images are resized where necessary to match the model input resolution and normalized using the corresponding in-distribution statistics. Detection is based on maximum softmax probability~\citep{hendrycks_2017_OOD} and evaluated using AUROC. We additionally train under long-tailed class imbalance (imbalance factors $10$, $50$, $100$) and symmetric label noise ($20\%$, $40\%$), reusing the clean-selected configuration.

\subsection{Cost Measurement}

We report wall-clock training time excluding evaluation, one-off graph construction time, and peak memory, measured on the same hardware for all methods within a setting. The reported \model wall-clock includes the warm-up prefix, so the comparison against baselines is like-for-like. All settings ran on either a single NVIDIA RTX 5000 Ada Gen. or NVIDIA RTX A6000 GPU, with the GPU model kept consistent for every run in that setting.

\subsection{Deviations from Reference Implementations}

We note the following deviations from the original implementations and protocols.

\emph{Protocol.} All methods share a validation split carved from the training set ($45$k/$5$k on CIFAR), are evaluated at the epoch of best validation NLL rather than at the final epoch, and are tuned on validation NLL; several of the compared methods tune on a calibration metric in their own papers. CIFAR budgets (Table~\ref{tab:app-optim}) are shorter than the $350$-epoch schedule of~\citet{mukhoti_focal}.

\emph{Architectures.} ResNet-50 uses the standard small-input stem ($3{\times}3$ convolution, no max-pooling) on both CIFAR and Tiny-ImageNet. The 20 Newsgroups model follows the canonical global-pooling CNN of~\citet{Kumar_2018_MMCE}, with the GloVe embedding matrix held frozen rather than fine-tuned, and a $50$-epoch budget. Tabular models use a fixed $3{\times}256$ MLP with a fixed epoch budget, rather than the per-dataset tuned architectures and early stopping of~\citet{gorishniy_2021_tabular}; accuracies are correspondingly $0.5$--$1.5$ points below their tuned reference values. Covertype is subsampled to $100$k rows.

\emph{Baselines.} MDCA is evaluated as cross-entropy plus MDCA, without the focal or label-smoothing compositions reported by its authors. Two further departures are shared across methods rather than specific to one. All CIFAR runs use batch size $128$, following~\citet{mukhoti_focal} and~\citet{Kumar_2018_MMCE}, rather than the $512$ of~\citet{karandikar_sAVuC_sbECE}. MDCA is trained end-to-end on the standard architecture rather than on the frozen-backbone head with additional convolutional and fully connected layers used by~\citet{Hebbalaguppe_mdca}. MbLS on 20 Newsgroups uses the $50$-epoch constant-rate schedule shared by all methods in that setting rather than the $100$-epoch decayed schedule of~\citet{Liu_2022_mbls}. Calibration metrics use $15$ bins throughout, following three of the four reference papers; \citet{Kumar_2018_MMCE} use $20$.

\emph{Post-hoc.} LoRe is applied in the same standardized PCA feature space and Gaussian kernel as our locality metric rather than at its published Laplacian bandwidth, which was tuned on different features. LoCal is run at its authors' published configuration in every setting, without per-setting tuning.

\section{Additional Results}\label{app:results}
 
\subsection{Full per-setting metrics}\label{app:full-metrics}
Tables~\ref{tab:app-brier-acc}--\ref{tab:app-local} report the metrics accompanying the NLL comparison in the main text. \model improves Brier score over CE in ten of eleven settings and accuracy in nine.

\model attains the lowest or second-lowest LCE in five of eight image and tabular settings, counting ties at displayed precision. Improvements are not uniform across settings or calibration measures. Classwise ECE values are small in absolute terms on the many-class datasets and should be interpreted alongside the other metrics.

\begin{table}[!htbp]\centering
\caption{Brier score (top, lower is better) and accuracy in percent (bottom, higher is better). Conventions as in Table~\ref{tab:res-main}.}
\label{tab:app-brier-acc}
\begingroup\scriptsize\setlength{\tabcolsep}{1.6pt}\renewcommand{\arraystretch}{1.03}
\resizebox{\linewidth}{!}{%
\begin{tabular}{@{}lcccccccc@{}}
\toprule
Setting & CE & FL & LS & MMCE & MDCA & s-AvUC & MbLS & NeSCal \\
\midrule
\multicolumn{9}{@{}l}{\emph{Brier score}} \\
CIFAR-10 / R50 & \nrc{0.0953}{0.0045} & \textbf{\nrc{0.0866}{0.0029}} & \textit{\nrc{0.0872}{0.0035}} & \nrc{0.0988}{0.0021} & \nrc{0.0943}{0.0087} & \nrc{0.0949}{0.0054} & \nrc{0.0948}{0.0057} & \textit{\nrc{0.0872}{0.0015}} \\
CIFAR-10 / WRN & \nrc{0.0782}{0.0051} & \textbf{\nrc{0.0733}{0.0016}} & \textit{\nrc{0.0750}{0.0019}} & \nrc{0.0854}{0.0030} & \nrc{0.0852}{0.0066} & \nrc{0.0831}{0.0052} & \nrc{0.0807}{0.0052} & \nrc{0.0785}{0.0018} \\
CIFAR-100 / R50 & \nrc{0.3447}{0.0081} & \textit{\nrc{0.3296}{0.0019}} & \nrc{0.3430}{0.0083} & \nrc{0.3481}{0.0128} & \nrc{0.3482}{0.0041} & \nrc{0.3443}{0.0057} & \nrc{0.3372}{0.0045} & \textbf{\nrc{0.3178}{0.0050}} \\
CIFAR-100 / WRN & \nrc{0.3321}{0.0033} & \textbf{\nrc{0.3024}{0.0022}} & \nrc{0.3162}{0.0015} & \nrc{0.3396}{0.0066} & \nrc{0.3353}{0.0058} & \nrc{0.3337}{0.0025} & \nrc{0.3161}{0.0060} & \textit{\nrc{0.3057}{0.0061}} \\
Tiny-ImageNet & \nrc{0.4778}{0.0057} & \nrc{0.4991}{0.0030} & \nrc{0.4757}{0.0063} & \nrc{0.4819}{0.0079} & \nrc{0.4696}{0.0071} & \textit{\nrc{0.4691}{0.0096}} & \textbf{\nrc{0.4651}{0.0073}} & \nrc{0.4693}{0.0064} \\
\addlinespace[2pt]
Helena & \nrc{0.7597}{0.0004} & \nrc{0.7707}{0.0011} & \textit{\nrc{0.7563}{0.0029}} & \nrc{0.7590}{0.0010} & \nrc{0.7589}{0.0013} & \nrc{0.7593}{0.0007} & \textbf{\nrc{0.7559}{0.0004}} & \nrc{0.7573}{0.0016} \\
Jannis & \nrc{0.4012}{0.0005} & \nrc{0.4520}{0.0033} & \textit{\nrc{0.3992}{0.0023}} & \nrc{0.4012}{0.0015} & \nrc{0.3998}{0.0008} & \nrc{0.4013}{0.0010} & \nrc{0.4012}{0.0014} & \textbf{\nrc{0.3973}{0.0006}} \\
Covertype & \nrc{0.1401}{0.0021} & \nrc{0.2085}{0.0030} & \nrc{0.1401}{0.0019} & \nrc{0.1396}{0.0006} & \nrc{0.1417}{0.0011} & \textbf{\nrc{0.1387}{0.0017}} & \textit{\nrc{0.1388}{0.0014}} & \nrc{0.1389}{0.0015} \\
\addlinespace[2pt]
20NG& \nrc{0.3247}{0.0116} & \nrc{0.2944}{0.0095} & \textbf{\nrc{0.2456}{0.0084}} & \nrc{0.2842}{0.0160} & \nrc{0.3263}{0.0126} & \nrc{0.3128}{0.0134} & \textit{\nrc{0.2491}{0.0113}} & \nrc{0.3200}{0.0061} \\
Amazon (5-class) & \nrc{0.4303}{0.0013} & \nrc{0.4418}{0.0007} & \textbf{\nrc{0.4241}{0.0009}} & \textit{\nrc{0.4264}{0.0013}} & \nrc{0.4303}{0.0017} & \nrc{0.4296}{0.0018} & \nrc{0.4291}{0.0009} & \nrc{0.4288}{0.0017} \\
Amazon (binary) & \nrc{0.1272}{0.0010} & \nrc{0.1597}{0.0059} & \textit{\nrc{0.1239}{0.0009}} & \textbf{\nrc{0.1211}{0.0005}} & \nrc{0.1275}{0.0033} & \nrc{0.1275}{0.0012} & \nrc{0.1261}{0.0014} & \nrc{0.1262}{0.0007} \\
\midrule
\multicolumn{9}{@{}l}{\emph{Accuracy (\%)}} \\
CIFAR-10 / R50 & \nrc{93.70}{0.47} & \nrc{94.24}{0.23} & \textbf{\nrc{94.56}{0.31}} & \nrc{93.22}{0.22} & \nrc{93.77}{0.78} & \nrc{93.76}{0.50} & \nrc{93.94}{0.64} & \textit{\nrc{94.28}{0.15}} \\
CIFAR-10 / WRN & \nrc{95.11}{0.51} & \textit{\nrc{95.19}{0.18}} & \textbf{\nrc{95.30}{0.14}} & \nrc{94.26}{0.12} & \nrc{94.49}{0.59} & \nrc{94.86}{0.45} & \nrc{94.76}{0.59} & \nrc{94.94}{0.11} \\
CIFAR-100 / R50 & \nrc{75.29}{0.82} & \textit{\nrc{76.41}{0.22}} & \nrc{75.18}{0.82} & \nrc{75.04}{1.07} & \nrc{75.02}{0.33} & \nrc{75.36}{0.37} & \nrc{75.66}{0.41} & \textbf{\nrc{77.19}{0.52}} \\
CIFAR-100 / WRN & \nrc{76.84}{0.26} & \textbf{\nrc{78.47}{0.19}} & \nrc{77.46}{0.28} & \nrc{75.94}{0.40} & \nrc{76.60}{0.45} & \nrc{76.32}{0.31} & \nrc{77.40}{0.52} & \textit{\nrc{78.10}{0.65}} \\
Tiny-ImageNet & \nrc{64.29}{0.52} & \nrc{62.37}{0.37} & \nrc{64.75}{0.53} & \nrc{63.83}{0.74} & \textit{\nrc{64.92}{0.61}} & \nrc{64.89}{0.67} & \textbf{\nrc{65.28}{0.56}} & \nrc{64.51}{0.34} \\
\addlinespace[2pt]
Helena & \nrc{37.55}{0.14} & \nrc{37.30}{0.15} & \textit{\nrc{37.83}{0.40}} & \nrc{37.59}{0.16} & \nrc{37.57}{0.21} & \nrc{37.46}{0.12} & \textbf{\nrc{37.91}{0.07}} & \nrc{37.76}{0.18} \\
Jannis & \nrc{70.46}{0.11} & \nrc{69.62}{0.46} & \textit{\nrc{70.84}{0.25}} & \nrc{70.49}{0.23} & \nrc{70.67}{0.07} & \nrc{70.54}{0.14} & \nrc{70.47}{0.20} & \textbf{\nrc{70.94}{0.10}} \\
Covertype & \nrc{90.43}{0.20} & \nrc{89.24}{0.25} & \nrc{90.56}{0.09} & \nrc{90.58}{0.05} & \nrc{90.37}{0.09} & \textit{\nrc{90.59}{0.13}} & \nrc{90.55}{0.15} & \textbf{\nrc{90.62}{0.17}} \\
\addlinespace[2pt]
20NG& \nrc{76.94}{1.55} & \nrc{79.53}{0.69} & \textbf{\nrc{83.24}{0.53}} & \nrc{80.13}{1.48} & \nrc{76.46}{1.62} & \nrc{77.71}{1.40} & \textit{\nrc{82.67}{1.05}} & \nrc{77.20}{0.81} \\
Amazon (5-class) & \nrc{68.58}{0.06} & \nrc{68.47}{0.33} & \textit{\nrc{68.67}{0.28}} & \nrc{68.60}{0.08} & \nrc{68.55}{0.18} & \nrc{68.53}{0.17} & \textbf{\nrc{68.70}{0.10}} & \nrc{68.46}{0.15} \\
Amazon (binary) & \nrc{91.74}{0.10} & \nrc{91.64}{0.11} & \textit{\nrc{91.79}{0.07}} & \textbf{\nrc{91.82}{0.10}} & \nrc{91.70}{0.10} & \nrc{91.70}{0.15} & \nrc{91.77}{0.13} & \textbf{\nrc{91.82}{0.05}} \\
\bottomrule
\end{tabular}}
\endgroup
\end{table}

\begin{table}[!htbp]\centering
\caption{ECE in percent for the base model (top) and after temperature scaling (bottom); lower is better.}
\label{tab:app-ece}
\begingroup\scriptsize\setlength{\tabcolsep}{1.6pt}\renewcommand{\arraystretch}{1.03}
\resizebox{\linewidth}{!}{%
\begin{tabular}{@{}lcccccccc@{}}
\toprule
Setting & CE & FL & LS & MMCE & MDCA & s-AvUC & MbLS & NeSCal \\
\midrule
\multicolumn{9}{@{}l}{\emph{Base model}} \\
CIFAR-10 / R50 & \nrc{2.46}{0.49} & \textit{\nrc{1.14}{0.04}} & \nrc{3.98}{0.21} & \textbf{\nrc{0.83}{0.10}} & \nrc{2.56}{0.40} & \nrc{2.55}{0.53} & \nrc{2.08}{0.49} & \nrc{2.29}{0.11} \\
CIFAR-10 / WRN & \nrc{2.75}{0.30} & \textbf{\nrc{0.78}{0.07}} & \nrc{1.64}{0.31} & \textit{\nrc{1.13}{0.25}} & \nrc{2.66}{0.14} & \nrc{3.00}{0.20} & \nrc{2.10}{0.51} & \nrc{2.39}{0.18} \\
CIFAR-100 / R50 & \nrc{5.90}{0.58} & \textit{\nrc{2.73}{0.50}} & \textbf{\nrc{1.39}{0.18}} & \nrc{6.02}{0.49} & \nrc{6.08}{0.24} & \nrc{5.83}{0.94} & \nrc{3.84}{0.51} & \nrc{3.95}{0.67} \\
CIFAR-100 / WRN & \nrc{8.19}{0.25} & \textbf{\nrc{1.87}{0.12}} & \textit{\nrc{3.41}{0.57}} & \nrc{7.45}{0.38} & \nrc{7.91}{0.18} & \nrc{7.03}{0.61} & \nrc{3.62}{0.29} & \nrc{4.60}{0.62} \\
Tiny-ImageNet & \nrc{5.75}{0.29} & \nrc{3.12}{0.23} & \nrc{4.63}{0.29} & \nrc{4.56}{0.35} & \nrc{6.00}{0.25} & \nrc{6.27}{0.47} & \textit{\nrc{2.69}{0.63}} & \textbf{\nrc{2.02}{0.36}} \\
\addlinespace[2pt]
Helena & \nrc{1.36}{0.24} & \nrc{5.66}{0.42} & \nrc{1.32}{0.25} & \nrc{1.53}{0.33} & \textbf{\nrc{1.19}{0.08}} & \nrc{1.36}{0.14} & \textit{\nrc{1.23}{0.17}} & \nrc{1.25}{0.07} \\
Jannis & \nrc{1.51}{0.49} & \nrc{14.36}{0.93} & \textbf{\nrc{1.37}{0.26}} & \nrc{1.96}{0.32} & \nrc{1.77}{0.39} & \nrc{2.55}{0.56} & \textit{\nrc{1.46}{0.38}} & \nrc{2.09}{0.39} \\
Covertype & \nrc{1.14}{0.15} & \nrc{14.25}{0.58} & \nrc{3.23}{0.21} & \textit{\nrc{1.11}{0.29}} & \nrc{1.15}{0.25} & \nrc{1.39}{0.16} & \textbf{\nrc{0.48}{0.10}} & \nrc{1.23}{0.27} \\
\addlinespace[2pt]
20NG& \nrc{5.72}{2.40} & \nrc{4.60}{1.65} & \textit{\nrc{2.32}{0.83}} & \nrc{5.30}{0.65} & \nrc{4.29}{2.32} & \nrc{5.18}{1.20} & \textbf{\nrc{2.11}{0.73}} & \nrc{4.80}{1.18} \\
Amazon (5-class) & \nrc{8.14}{0.62} & \nrc{10.63}{0.14} & \textbf{\nrc{5.50}{0.88}} & \textit{\nrc{7.49}{0.67}} & \nrc{8.82}{0.27} & \nrc{7.80}{1.33} & \nrc{8.01}{0.59} & \nrc{8.37}{0.75} \\
Amazon (binary) & \nrc{3.07}{0.63} & \nrc{13.09}{0.93} & \nrc{4.17}{0.18} & \textbf{\nrc{0.92}{0.18}} & \nrc{2.93}{0.72} & \textit{\nrc{2.58}{0.62}} & \nrc{2.60}{0.42} & \nrc{3.75}{0.04} \\
\midrule
\multicolumn{9}{@{}l}{\emph{After temperature scaling}} \\
CIFAR-10 / R50 & \textit{\nrc{0.64}{0.12}} & \nrc{0.91}{0.15} & \nrc{1.41}{0.23} & \nrc{0.81}{0.16} & \nrc{0.67}{0.16} & \nrc{0.66}{0.18} & \nrc{1.52}{0.09} & \textbf{\nrc{0.60}{0.14}} \\
CIFAR-10 / WRN & \textbf{\nrc{0.70}{0.15}} & \nrc{0.80}{0.07} & \nrc{1.19}{0.13} & \nrc{0.82}{0.20} & \nrc{0.73}{0.11} & \nrc{0.94}{0.14} & \nrc{1.20}{0.17} & \textit{\nrc{0.71}{0.20}} \\
CIFAR-100 / R50 & \textbf{\nrc{1.61}{0.25}} & \nrc{2.08}{0.16} & \nrc{2.52}{0.22} & \nrc{1.81}{0.31} & \nrc{1.68}{0.28} & \nrc{1.88}{0.25} & \nrc{1.94}{0.28} & \textit{\nrc{1.63}{0.24}} \\
CIFAR-100 / WRN & \nrc{1.86}{0.23} & \nrc{1.93}{0.20} & \nrc{2.65}{0.34} & \nrc{1.92}{0.25} & \nrc{1.94}{0.12} & \textit{\nrc{1.85}{0.33}} & \nrc{2.42}{0.26} & \textbf{\nrc{1.54}{0.23}} \\
Tiny-ImageNet & \nrc{1.98}{0.25} & \nrc{2.26}{0.17} & \nrc{2.46}{0.37} & \nrc{1.80}{0.32} & \nrc{1.59}{0.12} & \textbf{\nrc{1.39}{0.38}} & \textit{\nrc{1.45}{0.29}} & \nrc{1.74}{0.23} \\
\addlinespace[2pt]
Helena & \nrc{2.19}{0.10} & \nrc{5.84}{0.28} & \textbf{\nrc{1.33}{0.28}} & \nrc{2.18}{0.24} & \nrc{2.24}{0.42} & \nrc{2.15}{0.21} & \textit{\nrc{1.59}{0.31}} & \nrc{2.35}{0.32} \\
Jannis & \textit{\nrc{0.93}{0.15}} & \nrc{5.58}{0.97} & \nrc{2.12}{0.38} & \nrc{1.01}{0.07} & \nrc{1.08}{0.19} & \nrc{1.08}{0.10} & \nrc{1.21}{0.30} & \textbf{\nrc{0.74}{0.22}} \\
Covertype & \nrc{1.27}{0.21} & \nrc{2.90}{0.44} & \nrc{1.19}{0.23} & \nrc{1.10}{0.15} & \textit{\nrc{0.99}{0.23}} & \nrc{1.09}{0.08} & \textbf{\nrc{0.58}{0.14}} & \nrc{1.02}{0.10} \\
\addlinespace[2pt]
20NG& \textbf{\nrc{2.08}{0.53}} & \nrc{4.53}{1.30} & \nrc{2.67}{0.84} & \nrc{3.81}{1.28} & \textit{\nrc{2.39}{0.60}} & \nrc{2.66}{0.35} & \nrc{2.56}{0.47} & \nrc{3.05}{0.35} \\
Amazon (5-class) & \nrc{4.84}{0.15} & \nrc{5.20}{0.60} & \nrc{5.56}{0.30} & \textbf{\nrc{4.11}{0.20}} & \textit{\nrc{4.49}{0.30}} & \nrc{4.77}{0.22} & \nrc{4.87}{0.17} & \nrc{4.84}{0.08} \\
Amazon (binary) & \nrc{2.68}{0.78} & \textit{\nrc{1.85}{0.24}} & \nrc{3.31}{0.18} & \textbf{\nrc{0.95}{0.19}} & \nrc{2.26}{0.40} & \nrc{2.69}{0.45} & \nrc{2.67}{0.42} & \nrc{1.95}{0.22} \\
\bottomrule
\end{tabular}}
\endgroup
\end{table}

\begin{table}[!htbp]\centering
\caption{Classwise ECE, LCE, and equal-mass ECE, all in percent; lower is better.}
\label{tab:app-local}
\begingroup\scriptsize\setlength{\tabcolsep}{1.6pt}\renewcommand{\arraystretch}{1.03}
\resizebox{\linewidth}{!}{%
\begin{tabular}{@{}lcccccccc@{}}
\toprule
Setting & CE & FL & LS & MMCE & MDCA & s-AvUC & MbLS & NeSCal \\
\midrule
\multicolumn{9}{@{}l}{\emph{Classwise ECE}} \\
CIFAR-10 / R50 & \nrc{0.58}{0.08} & \textit{\nrc{0.43}{0.02}} & \nrc{0.83}{0.02} & \textbf{\nrc{0.40}{0.05}} & \nrc{0.58}{0.08} & \nrc{0.58}{0.07} & \nrc{0.57}{0.04} & \nrc{0.54}{0.02} \\
CIFAR-10 / WRN & \nrc{0.61}{0.06} & \textit{\nrc{0.45}{0.04}} & \nrc{0.53}{0.04} & \textbf{\nrc{0.42}{0.02}} & \nrc{0.61}{0.03} & \nrc{0.68}{0.04} & \nrc{0.50}{0.07} & \nrc{0.59}{0.03} \\
CIFAR-100 / R50 & \nrc{0.21}{0.00} & \textbf{\nrc{0.18}{0.00}} & \textit{\nrc{0.19}{0.00}} & \nrc{0.21}{0.01} & \nrc{0.21}{0.01} & \nrc{0.21}{0.01} & \textit{\nrc{0.19}{0.01}} & \textbf{\nrc{0.18}{0.01}} \\
CIFAR-100 / WRN & \nrc{0.24}{0.00} & \textbf{\nrc{0.18}{0.00}} & \nrc{0.20}{0.01} & \nrc{0.23}{0.01} & \nrc{0.24}{0.01} & \nrc{0.22}{0.01} & \nrc{0.20}{0.00} & \textit{\nrc{0.19}{0.00}} \\
Tiny-ImageNet & \textit{\nrc{0.14}{0.00}} & \textit{ \nrc{0.14}{0.00}} & \textit{\nrc{0.14}{0.00}} & \textit{\nrc{0.14}{0.00}} & \textit{\nrc{0.14}{0.00}} & \nrc{0.15}{0.00} & \textit{\nrc{0.14}{0.00}} & \textbf{\nrc{0.13}{0.00}} \\
\addlinespace[2pt]
Helena & \textit{\nrc{0.16}{0.01}} & \nrc{0.20}{0.01} & \textit{\nrc{0.16}{0.00}} & \textit{\nrc{0.16}{0.01}} & \textit{\nrc{0.16}{0.01}} & \textit{\nrc{0.16}{0.01}} & \textit{\nrc{0.16}{0.00}} & \textbf{\nrc{0.15}{0.00}} \\
Jannis & \nrc{1.34}{0.18} & \nrc{6.71}{0.23} & \nrc{1.74}{0.27} & \nrc{1.39}{0.28} & \nrc{1.36}{0.23} & \nrc{1.69}{0.21} & \textbf{\nrc{1.23}{0.23}} & \textit{\nrc{1.25}{0.11}} \\
Covertype & \nrc{0.53}{0.05} & \nrc{4.05}{0.18} & \nrc{1.16}{0.06} & \textit{\nrc{0.49}{0.09}} & \nrc{0.50}{0.05} & \nrc{0.54}{0.03} & \textbf{\nrc{0.40}{0.04}} & \nrc{0.54}{0.05} \\
\addlinespace[2pt]
20NG& \nrc{1.01}{0.15} & \nrc{0.98}{0.07} & \textit{\nrc{0.80}{0.06}} & \nrc{0.85}{0.06} & \nrc{0.90}{0.12} & \nrc{0.91}{0.09} & \textbf{\nrc{0.73}{0.03}} & \nrc{0.95}{0.05} \\
Amazon (5-class) & \nrc{3.59}{0.17} & \nrc{4.45}{0.10} & \textbf{\nrc{3.04}{0.24}} & \textit{\nrc{3.41}{0.23}} & \nrc{3.77}{0.05} & \nrc{3.44}{0.39} & \nrc{3.54}{0.16} & \nrc{3.57}{0.25} \\
Amazon (binary) & \nrc{3.33}{0.62} & \nrc{13.22}{0.81} & \nrc{4.34}{0.22} & \textbf{\nrc{2.01}{0.22}} & \nrc{3.70}{0.96} & \textit{\nrc{2.78}{0.57}} & \nrc{3.14}{0.52} & \nrc{3.79}{0.07} \\
\midrule
\multicolumn{9}{@{}l}{\emph{LCE}} \\
CIFAR-10 / R50 & \nrc{0.70}{0.13} & \nrc{0.90}{0.05} & \nrc{1.07}{0.05} & \nrc{0.93}{0.06} & \textit{\nrc{0.69}{0.17}} & \nrc{0.71}{0.17} & \nrc{0.95}{0.29} & \textbf{\nrc{0.61}{0.04}} \\
CIFAR-10 / WRN & \nrc{0.45}{0.04} & \nrc{0.46}{0.03} & \nrc{0.45}{0.04} & \nrc{0.45}{0.03} & \nrc{0.49}{0.04} & \nrc{0.50}{0.03} & \textbf{\nrc{0.42}{0.03}} & \textit{\nrc{0.43}{0.04}} \\
CIFAR-100 / R50 & \textbf{\nrc{0.24}{0.01}} & \nrc{0.28}{0.00} & \nrc{0.32}{0.01} & \textit{\nrc{0.25}{0.01}} & \textbf{\nrc{0.24}{0.01}} & \textbf{\nrc{0.24}{0.01}} & \nrc{0.28}{0.01} & \nrc{0.26}{0.01} \\
CIFAR-100 / WRN & \nrc{0.13}{0.00} & \textit{\nrc{0.10}{0.00}} & \nrc{0.11}{0.00} & \nrc{0.13}{0.01} & \nrc{0.13}{0.01} & \nrc{0.12}{0.00} & \textit{\nrc{0.10}{0.00}} & \textbf{\nrc{0.09}{0.01}} \\
Tiny-ImageNet & \textit{\nrc{0.18}{0.00}} & \nrc{0.25}{0.00} & \nrc{0.24}{0.00} & \nrc{0.19}{0.00} & \textit{\nrc{0.18}{0.00}} & \textbf{\nrc{0.17}{0.00}} & \nrc{0.23}{0.00} & \nrc{0.20}{0.00} \\
\addlinespace[2pt]
Helena & \textit{\nrc{0.14}{0.01}} & \nrc{0.19}{0.01} & \nrc{0.15}{0.01} & \nrc{0.15}{0.01} & \textit{\nrc{0.14}{0.01}} & \nrc{0.15}{0.01} & \textit{\nrc{0.14}{0.00}} & \textbf{\nrc{0.13}{0.00}} \\
Jannis & \textit{\nrc{1.42}{0.14}} & \nrc{6.81}{0.22} & \nrc{1.89}{0.27} & \nrc{1.47}{0.19} & \textit{\nrc{1.42}{0.23}} & \nrc{1.77}{0.18} & \textbf{\nrc{1.34}{0.15}} & \textbf{\nrc{1.34}{0.10}} \\
Covertype & \nrc{0.52}{0.05} & \nrc{4.07}{0.16} & \nrc{1.22}{0.06} & \textit{\nrc{0.49}{0.08}} & \nrc{0.50}{0.07} & \nrc{0.53}{0.03} & \textbf{\nrc{0.42}{0.04}} & \nrc{0.53}{0.05} \\
\addlinespace[2pt]
20NG& \nrc{0.71}{0.13} & \nrc{0.67}{0.05} & \textit{\nrc{0.54}{0.06}} & \nrc{0.55}{0.05} & \nrc{0.66}{0.08} & \nrc{0.61}{0.09} & \textbf{\nrc{0.45}{0.03}} & \nrc{0.68}{0.04} \\
Amazon (5-class) & \nrc{4.10}{0.15} & \nrc{4.97}{0.09} & \textbf{\nrc{3.71}{0.16}} & \textit{\nrc{3.84}{0.19}} & \nrc{4.22}{0.11} & \nrc{4.03}{0.27} & \nrc{4.07}{0.10} & \nrc{4.07}{0.22} \\
Amazon (binary) & \nrc{3.63}{0.47} & \nrc{13.47}{0.79} & \nrc{4.73}{0.15} & \textbf{\nrc{2.77}{0.14}} & \nrc{4.52}{0.91} & \nrc{3.56}{0.50} & \nrc{3.73}{0.62} & \textit{\nrc{3.22}{0.14}} \\
\midrule
\multicolumn{9}{@{}l}{\emph{Equal-mass ECE}} \\
CIFAR-10 / R50 & \nrc{2.45}{0.50} & \textit{\nrc{0.93}{0.12}} & \nrc{3.93}{0.29} & \textbf{\nrc{0.77}{0.17}} & \nrc{2.53}{0.39} & \nrc{2.53}{0.53} & \nrc{2.62}{0.31} & \nrc{2.23}{0.13} \\
CIFAR-10 / WRN & \nrc{2.73}{0.31} &\textbf{ \nrc{0.71}{0.14}} & \nrc{2.17}{0.15} & \textit{\nrc{0.94}{0.16}} & \nrc{2.62}{0.11} & \nrc{2.99}{0.20} & \nrc{2.14}{0.49} & \nrc{2.33}{0.15} \\
CIFAR-100 / R50 & \nrc{5.85}{0.59} & \textit{\nrc{2.70}{0.48}} & \textbf{\nrc{1.41}{0.18}} & \nrc{6.00}{0.48} & \nrc{6.07}{0.23} & \nrc{5.81}{0.95} & \nrc{3.82}{0.49} & \nrc{3.87}{0.71} \\
CIFAR-100 / WRN & \nrc{8.18}{0.25} & \textbf{\nrc{1.81}{0.22}} & \textit{\nrc{3.27}{0.51}} & \nrc{7.42}{0.40} & \nrc{7.87}{0.21} & \nrc{7.02}{0.61} & \nrc{3.55}{0.30} & \nrc{4.52}{0.70} \\
Tiny-ImageNet & \nrc{5.75}{0.29} & \nrc{3.19}{0.27} & \nrc{4.73}{0.21} & \nrc{4.55}{0.33} & \nrc{5.97}{0.29} & \nrc{6.27}{0.47} & \textit{\nrc{2.69}{0.59}} & \textbf{\nrc{1.92}{0.37}} \\
\addlinespace[2pt]
Helena & \nrc{1.47}{0.38} & \nrc{5.74}{0.47} & \nrc{1.39}{0.28} & \nrc{1.49}{0.28} & \nrc{1.37}{0.18} & \nrc{1.46}{0.07} & \textit{\nrc{1.35}{0.17}} & \textbf{\nrc{1.33}{0.10}} \\
Jannis & \textit{\nrc{1.64}{0.32}} & \nrc{14.35}{0.92} & \textbf{\nrc{1.45}{0.26}} & \nrc{1.98}{0.23} & \nrc{1.84}{0.55} & \nrc{2.64}{0.46} & \nrc{1.65}{0.22} & \nrc{2.03}{0.38} \\
Covertype & \nrc{1.13}{0.15} & \nrc{14.24}{0.58} & \nrc{3.21}{0.22} & \textit{\nrc{1.04}{0.31}} & \nrc{1.12}{0.17} & \nrc{1.33}{0.21} & \textbf{\nrc{0.55}{0.03}} & \nrc{1.21}{0.28} \\
\addlinespace[2pt]
20NG & \nrc{5.62}{2.52} & \nrc{4.77}{1.44} &\textit{ \nrc{2.45}{0.69}} & \nrc{5.14}{0.65} & \nrc{4.19}{2.37} & \nrc{5.13}{1.22} & \textbf{\nrc{2.22}{0.52}} & \nrc{4.76}{1.15} \\
Amazon (5-class) & \nrc{8.14}{0.57} & \nrc{10.64}{0.13} & \textbf{ \nrc{5.82}{0.81}} & \textit{\nrc{7.52}{0.65}} & \nrc{8.89}{0.23} & \nrc{7.84}{1.32} & \nrc{7.98}{0.57} & \nrc{8.43}{0.68} \\
Amazon (binary) & \nrc{3.89}{0.39} & \nrc{13.15}{1.08} & \nrc{4.00}{0.29} & \textbf{\nrc{2.26}{0.16}} & \nrc{4.38}{0.86} & \nrc{3.73}{0.20} & \nrc{3.54}{0.38} & \textit{\nrc{3.31}{0.09}} \\
\bottomrule
\end{tabular}}
\endgroup
\end{table}

\subsection{Across-setting comparisons}\label{app:tests}
We summarize pairwise wins across setting means and report exploratory Wilcoxon signed-rank comparisons in Table~\ref{tab:res-tests}, which counts, for each baseline, the settings in which \model attains the lower mean, and applies an exact two-sided Wilcoxon signed-rank test to the per-setting mean differences. $p$-values are uncorrected; with eight settings the smallest attainable is $0.008$. Several settings share datasets, thus the uncorrected $p$-values are interpreted as exploratory summaries. For NLL after temperature scaling \model wins in all eight image and tabular settings against FL, LS, MMCE and MbLS, and in seven of eight against MDCA and s-AvUC.

\begin{table}[!htbp]
\centering
\caption{\model against each baseline across settings. ``All'' covers the eleven completed settings, ``I+T'' the eight image and tabular settings. }
\label{tab:res-tests}
\begingroup\scriptsize\setlength{\tabcolsep}{3pt}
\begin{tabular}{@{}lcccccccccc@{}}
\toprule
 & \multicolumn{2}{c}{NLL (all)} & \multicolumn{2}{c}{NLL (I+T)} & \multicolumn{2}{c}{NLL+TS (all)} &
   \multicolumn{2}{c}{NLL+TS (I+T)} & \multicolumn{2}{c}{Brier (all)} \\
\cmidrule(lr){2-3}\cmidrule(lr){4-5}\cmidrule(lr){6-7}\cmidrule(lr){8-9}\cmidrule(lr){10-11}
vs.\ & wins & $p$ & wins & $p$ & wins & $p$ & wins & $p$ & wins & $p$ \\
\midrule
CE     & 11/11 & 0.001 & 8/8 & 0.008 & 11/11 & 0.001 & 8/8 & 0.008 & 10/11 & 0.002 \\
FL     &  7/11 & 0.067 & 5/8 & 0.109 & 10/11 & 0.014 & 8/8 & 0.008 &  7/11 & 0.083 \\
LS     &  7/11 & 0.175 & 7/8 & 0.016 & 10/11 & 0.042 & 8/8 & 0.008 &  5/11 & 0.878 \\
MMCE   &  8/11 & 0.206 & 8/8 & 0.008 &  8/11 & 0.083 & 8/8 & 0.008 &  8/11 & 0.240 \\
MDCA   & 10/11 & 0.019 & 8/8 & 0.008 &  9/11 & 0.005 & 7/8 & 0.016 & 11/11 & 0.001 \\
s-AvUC & 10/11 & 0.019 & 8/8 & 0.008 &  9/11 & 0.024 & 7/8 & 0.023 &  8/11 & 0.054 \\
MbLS   &  7/11 & 0.278 & 6/8 & 0.039 & 10/11 & 0.054 & 8/8 & 0.008 &  6/11 & 0.520 \\
\bottomrule
\end{tabular}
\endgroup
\end{table}

\subsection{Calibration--refinement decomposition}\label{app:refinement}
Table~\ref{tab:refinement-all} reports the binned refinement component of the base-model top-label Brier decomposition. This component measures within-bin variation in correctness and is lower for \model than for CE in ten of eleven real-data settings. It provides a complementary diagnostic under the specified binning scheme; recalibration can change bin membership and therefore the estimate.
\providecommand{\nesref}[2]{\nrc{#1}{#2}}

\begin{table}[!htbp]
\centering
\caption{Reported binned refinement component of the base-model top-label Brier decomposition; lower is better.}
\label{tab:refinement-all}
\begingroup
\scriptsize
\setlength{\tabcolsep}{1.6pt}
\renewcommand{\arraystretch}{1.03}
\resizebox{\linewidth}{!}{%
\begin{tabular}{@{}lcccccccc@{}}
\toprule
Setting & CE & FL & LS & MMCE & MDCA & s-AvUC & MbLS & NeSCal \\
\midrule
CIFAR-10 / R50
& \nesref{0.0424}{0.0020}
& \nesref{0.0470}{0.0046}
& \textbf{\nesref{0.0381}{0.0012}}
& \nesref{0.0439}{0.0012}
& \nesref{0.0420}{0.0036}
& \nesref{0.0424}{0.0026}
& \nesref{0.0424}{0.0027}
& \textit{\nesref{0.0388}{0.0006}} \\
CIFAR-10 / WRN
& \textit{\nesref{0.0346}{0.0024}}
& \nesref{0.0377}{0.0038}
& \textbf{\nesref{0.0326}{0.0008}}
& \nesref{0.0382}{0.0011}
& \nesref{0.0377}{0.0030}
& \nesref{0.0365}{0.0023}
& \nesref{0.0359}{0.0023}
& \nesref{0.0352}{0.0010} \\
CIFAR-100 / R50
& \nesref{0.1242}{0.0014}
& \textit{\nesref{0.1229}{0.0015}}
& \nesref{0.1249}{0.0007}
& \nesref{0.1239}{0.0032}
& \nesref{0.1248}{0.0013}
& \nesref{0.1237}{0.0019}
& \nesref{0.1231}{0.0016}
& \textbf{\nesref{0.1160}{0.0019}} \\
CIFAR-100 / WRN
& \nesref{0.1185}{0.0018}
& \nesref{0.1151}{0.0010}
& \textbf{\nesref{0.1069}{0.0005}}
& \nesref{0.1203}{0.0016}
& \nesref{0.1197}{0.0018}
& \nesref{0.1189}{0.0009}
& \nesref{0.1161}{0.0026}
& \textit{\nesref{0.1118}{0.0014}} \\
Tiny-ImageNet
& \nesref{0.1463}{0.0014}
& \nesref{0.1558}{0.0011}
& \nesref{0.1497}{0.0016}
& \nesref{0.1484}{0.0022}
& \textit{\nesref{0.1448}{0.0016}}
& \textbf{\nesref{0.1434}{0.0029}}
& \nesref{0.1468}{0.0019}
& \nesref{0.1453}{0.0030} \\
Helena
& \nesref{0.1777}{0.0008}
& \nesref{0.1789}{0.0013}
& \nesref{0.1772}{0.0006}
& \nesref{0.1770}{0.0010}
& \textit{\nesref{0.1767}{0.0009}}
& \textbf{\nesref{0.1762}{0.0013}}
& \nesref{0.1772}{0.0006}
& \nesref{0.1773}{0.0009} \\
Jannis
& \nesref{0.1783}{0.0005}
& \nesref{0.1860}{0.0007}
& \nesref{0.1784}{0.0001}
& \nesref{0.1783}{0.0004}
& \textit{\nesref{0.1773}{0.0003}}
& \nesref{0.1781}{0.0003}
& \nesref{0.1783}{0.0004}
& \textbf{\nesref{0.1770}{0.0002}} \\
Covertype
& \nesref{0.0690}{0.0011}
& \nesref{0.0787}{0.0019}
& \textbf{\nesref{0.0673}{0.0010}}
& \nesref{0.0690}{0.0003}
& \nesref{0.0698}{0.0005}
& \textit{\nesref{0.0683}{0.0007}}
& \nesref{0.0685}{0.0007}
& \nesref{0.0685}{0.0007} \\
20NG
& \nesref{0.1227}{0.0039}
& \nesref{0.1397}{0.0045}
& \textbf{\nesref{0.0930}{0.0025}}
& \nesref{0.1058}{0.0048}
& \nesref{0.1231}{0.0043}
& \nesref{0.1178}{0.0055}
& \textit{\nesref{0.0953}{0.0042}}
& \nesref{0.1220}{0.0024} \\
Amazon (5-class)
& \nesref{0.1808}{0.0008}
& \nesref{0.1796}{0.0008}
& \nesref{0.1798}{0.0010}
& \textit{\nesref{0.1793}{0.0008}}
& \textbf{\nesref{0.1790}{0.0004}}
& \nesref{0.1805}{0.0007}
& \nesref{0.1801}{0.0002}
& \textit{\nesref{0.1793}{0.0003}} \\
Amazon (binary)
& \nesref{0.0618}{0.0007}
& \nesref{0.0608}{0.0006}
& \textbf{\nesref{0.0603}{0.0012}}
& \textit{\nesref{0.0605}{0.0003}}
& \nesref{0.0621}{0.0013}
& \nesref{0.0617}{0.0011}
& \nesref{0.0615}{0.0011}
& \nesref{0.0608}{0.0005} \\
\bottomrule
\end{tabular}}
\endgroup
\end{table}
 
\subsection{Post-hoc composition}\label{app:posthoc}
Table~\ref{tab:best-posthoc} reports each train-time method's best observed NLL and ECE across the five calibrators, selected separately and retrospectively on test means. \model attains the lowest NLL in six of eleven real-data settings, all of them image or tabular.

Composition does not improve every metric: on Helena, TS slightly lowers \modelNoSpace's NLL while increasing ECE from $1.25\%$ to $2.35\%$.

\providecommand{\nespc}[3]{\mbox{\nrc{#1}{#2}\,\textsuperscript{#3}}}

\begin{table}[t]
\centering
\caption{Best observed post-hoc results for each train-time method. NLL and ECE are minimized separately over TS (T), LoCal (L), Dirichlet (D), DAC (A), and LoRe (R); ECE is in percent. Best method selection is retrospective on test means. Calibrator ties at reported precision use the order T, L, D, A, R. }
\label{tab:best-posthoc}
\begingroup
\scriptsize
\setlength{\tabcolsep}{1.6pt}
\renewcommand{\arraystretch}{1.03}
\resizebox{\linewidth}{!}{%
\begin{tabular}{@{}lcccccccc@{}}
\toprule
Setting & CE & FL & LS & MMCE & MDCA & s-AvUC & MbLS & NeSCal \\
\midrule
\multicolumn{9}{c}{NLL $\downarrow$}\\
\midrule
CIFAR-10 / R50
& \nespc{0.1863}{0.0084}{T}
& \textit{\nespc{0.1790}{0.0052}{L}}
& \nespc{0.2108}{0.0046}{D}
& \nespc{0.1993}{0.0061}{L}
& \nespc{0.1858}{0.0144}{T}
& \nespc{0.1855}{0.0121}{L}
& \nespc{0.2156}{0.0064}{A}
& \textbf{\nespc{0.1698}{0.0046}{T}} \\
CIFAR-10 / WRN
& \textit{\nespc{0.1495}{0.0129}{L}}
& \nespc{0.1508}{0.0043}{L}
& \nespc{0.1637}{0.0028}{L}
& \nespc{0.1726}{0.0063}{L}
& \nespc{0.1618}{0.0115}{L}
& \nespc{0.1540}{0.0117}{L}
& \nespc{0.1599}{0.0095}{L}
& \textbf{\nespc{0.1444}{0.0041}{L}} \\
CIFAR-100 / R50
& \nespc{0.8489}{0.0267}{T}
& \textit{\nespc{0.8452}{0.0038}{T}}
& \nespc{0.8944}{0.0190}{A}
& \nespc{0.8631}{0.0402}{T}
& \nespc{0.8577}{0.0137}{T}
& \nespc{0.8478}{0.0137}{T}
& \nespc{0.8558}{0.0168}{T}
& \textbf{\nespc{0.7944}{0.0191}{T}} \\
CIFAR-100 / WRN
& \nespc{0.8070}{0.0061}{T}
& \textit{\nespc{0.7681}{0.0057}{T}}
& \nespc{0.8334}{0.0049}{D}
& \nespc{0.8309}{0.0151}{T}
& \nespc{0.8141}{0.0155}{T}
& \nespc{0.8212}{0.0099}{T}
& \nespc{0.8173}{0.0103}{T}
& \textbf{\nespc{0.7613}{0.0178}{T}} \\
Tiny-ImageNet
& \nespc{1.4415}{0.0246}{T}
& \nespc{1.4950}{0.0099}{T}
& \nespc{1.4563}{0.0255}{T}
& \nespc{1.4614}{0.0319}{T}
& \textit{\nespc{1.4096}{0.0228}{T}}
& \textbf{\nespc{1.4066}{0.0307}{T}}
& \nespc{1.4208}{0.0246}{T}
& \nespc{1.4098}{0.0255}{T} \\
Helena
& \nespc{2.5618}{0.0015}{T}
& \nespc{2.5760}{0.0058}{D}
& \textbf{\nespc{2.5461}{0.0099}{D}}
& \nespc{2.5615}{0.0036}{T}
& \nespc{2.5653}{0.0025}{T}
& \nespc{2.5624}{0.0027}{T}
& \textit{\nespc{2.5463}{0.0039}{D}}
& \nespc{2.5497}{0.0049}{T} \\
Jannis
& \nespc{0.7085}{0.0014}{D}
& \nespc{0.7267}{0.0040}{D}
& \textit{\nespc{0.7068}{0.0019}{D}}
& \nespc{0.7090}{0.0025}{D}
& \nespc{0.7085}{0.0015}{D}
& \nespc{0.7084}{0.0011}{D}
& \nespc{0.7094}{0.0021}{D}
& \textbf{\nespc{0.7038}{0.0010}{D}} \\
Covertype
& \nespc{0.2461}{0.0023}{D}
& \nespc{0.2728}{0.0023}{R}
& \nespc{0.2483}{0.0028}{D}
& \nespc{0.2455}{0.0011}{A}
& \nespc{0.2485}{0.0013}{A}
& \textit{\nespc{0.2444}{0.0033}{A}}
& \nespc{0.2454}{0.0016}{D}
& \textbf{\nespc{0.2443}{0.0014}{A}} \\
20NG
& \nespc{0.7305}{0.0267}{A}
& \nespc{0.7008}{0.0318}{L}
& \textit{\nespc{0.6301}{0.0161}{D}}
& \nespc{0.6969}{0.0201}{A}
& \nespc{0.7337}{0.0377}{A}
& \nespc{0.7209}{0.0281}{T}
& \textbf{\nespc{0.6075}{0.0171}{L}}
& \nespc{0.7285}{0.0219}{A} \\
Amazon (5-class)
& \nespc{0.7534}{0.0024}{L}
& \nespc{0.7534}{0.0025}{R}
& \textbf{\nespc{0.7486}{0.0027}{L}}
& \textit{\nespc{0.7499}{0.0016}{L}}
& \nespc{0.7513}{0.0021}{L}
& \nespc{0.7526}{0.0015}{L}
& \nespc{0.7513}{0.0007}{L}
& \nespc{0.7521}{0.0016}{L} \\
Amazon (binary)
& \nespc{0.2111}{0.0021}{R}
& \nespc{0.2097}{0.0052}{A}
& \textit{\nespc{0.2045}{0.0012}{L}}
& \textbf{\nespc{0.2012}{0.0010}{D}}
& \nespc{0.2107}{0.0047}{R}
& \nespc{0.2098}{0.0032}{R}
& \nespc{0.2086}{0.0041}{R}
& \nespc{0.2052}{0.0014}{R} \\
\midrule
Synthetic-noise
& \nespc{1.8432}{0.0224}{T}
& \nespc{1.8465}{0.0243}{A}
& \nespc{1.8475}{0.0222}{T}
& \nespc{1.8434}{0.0229}{T}
& \nespc{1.8448}{0.0215}{T}
& \nespc{1.8450}{0.0228}{T}
& \textit{\nespc{1.8431}{0.0223}{T}}
& \textbf{\nespc{1.8294}{0.0241}{T}} \\
Synthetic-geometry
& \nespc{0.9275}{0.0205}{D}
& \nespc{0.9262}{0.0221}{D}
& \textbf{\nespc{0.9243}{0.0219}{D}}
& \nespc{0.9271}{0.0193}{D}
& \nespc{0.9417}{0.0162}{D}
& \nespc{0.9286}{0.0201}{D}
& \textit{\nespc{0.9250}{0.0187}{D}}
& \nespc{0.9291}{0.0183}{D} \\
\midrule
\multicolumn{9}{c}{ECE (\%) $\downarrow$}\\
\midrule
CIFAR-10 / R50
& \textit{\nespc{0.64}{0.12}{T}}
& \nespc{0.85}{0.16}{A}
& \nespc{1.23}{0.15}{A}
& \nespc{0.81}{0.16}{T}
& \nespc{0.67}{0.16}{T}
& \nespc{0.66}{0.18}{T}
& \nespc{1.22}{0.17}{A}
& \textbf{\nespc{0.59}{0.12}{A}} \\
CIFAR-10 / WRN
& \nespc{0.67}{0.15}{A}
& \nespc{0.76}{0.17}{L}
& \textbf{\nespc{0.56}{0.05}{R}}
& \nespc{0.63}{0.07}{R}
& \nespc{0.63}{0.15}{L}
& \nespc{0.88}{0.11}{A}
& \nespc{0.65}{0.15}{R}
& \textit{\nespc{0.59}{0.04}{L}} \\
CIFAR-100 / R50
& \nespc{1.61}{0.25}{T}
& \nespc{1.63}{0.25}{L}
& \textit{\nespc{1.42}{0.36}{L}}
& \nespc{1.80}{0.32}{A}
& \nespc{1.68}{0.28}{T}
& \nespc{1.88}{0.25}{T}
& \nespc{1.60}{0.27}{L}
& \textbf{\nespc{1.11}{0.39}{L}} \\
CIFAR-100 / WRN
& \nespc{1.86}{0.23}{T}
& \nespc{1.70}{0.19}{R}
& \nespc{1.88}{0.25}{R}
& \nespc{1.92}{0.25}{T}
& \nespc{1.94}{0.12}{T}
& \nespc{1.85}{0.33}{T}
& \textit{\nespc{1.61}{0.36}{R}}
& \textbf{\nespc{1.54}{0.23}{T}} \\
Tiny-ImageNet
& \nespc{1.98}{0.25}{T}
& \nespc{2.20}{0.20}{A}
& \nespc{2.46}{0.37}{T}
& \nespc{1.80}{0.32}{T}
& \nespc{1.58}{0.11}{A}
& \textbf{\nespc{1.38}{0.33}{A}}
& \textit{\nespc{1.45}{0.29}{T}}
& \nespc{1.74}{0.23}{T} \\
Helena
& \nespc{1.39}{0.24}{D}
& \nespc{1.48}{0.17}{R}
& \textbf{\nespc{1.33}{0.28}{T}}
& \nespc{1.38}{0.14}{D}
& \nespc{1.42}{0.10}{D}
& \nespc{1.43}{0.30}{D}
& \textit{\nespc{1.34}{0.30}{D}}
& \nespc{1.36}{0.14}{D} \\
Jannis
& \textit{\nespc{0.87}{0.19}{D}}
& \nespc{1.18}{0.21}{D}
& \nespc{1.07}{0.36}{D}
& \nespc{1.01}{0.07}{T}
& \nespc{1.08}{0.19}{T}
& \nespc{0.92}{0.17}{R}
& \nespc{1.03}{0.29}{D}
& \textbf{\nespc{0.74}{0.22}{T}} \\
Covertype
& \nespc{0.61}{0.16}{R}
& \nespc{0.50}{0.10}{R}
& \textit{\nespc{0.42}{0.17}{R}}
& \nespc{0.53}{0.18}{R}
& \nespc{0.47}{0.08}{R}
& \nespc{0.59}{0.13}{R}
& \textbf{\nespc{0.39}{0.05}{D}}
& \nespc{0.67}{0.17}{R} \\
20NG
& \textbf{\nespc{1.44}{0.25}{D}}
& \nespc{2.06}{0.80}{D}
& \nespc{2.12}{0.45}{R}
& \nespc{2.14}{0.50}{R}
& \textit{\nespc{1.68}{0.31}{D}}
& \nespc{2.34}{0.16}{D}
& \nespc{1.96}{0.37}{D}
& \nespc{2.04}{0.28}{D} \\
Amazon (5-class)
& \nespc{2.99}{0.12}{L}
& \nespc{2.80}{0.34}{R}
& \nespc{3.00}{0.24}{L}
& \textbf{\nespc{2.42}{0.24}{L}}
& \textit{\nespc{2.63}{0.23}{L}}
& \nespc{3.29}{0.39}{L}
& \nespc{3.16}{0.26}{L}
& \nespc{2.73}{0.17}{L} \\
Amazon (binary)
& \nespc{1.38}{0.20}{R}
& \nespc{1.27}{0.36}{R}
& \nespc{1.39}{0.10}{R}
& \textbf{\nespc{0.93}{0.18}{D}}
& \nespc{1.49}{0.21}{R}
& \nespc{1.87}{0.33}{R}
& \nespc{1.50}{0.42}{R}
& \textit{\nespc{1.22}{0.28}{R}} \\
\midrule
Synthetic-noise
& \textbf{\nespc{1.96}{0.88}{R}}
& \nespc{2.84}{0.63}{R}
& \nespc{2.60}{0.68}{R}
& \nespc{2.59}{1.00}{R}
& \nespc{3.04}{1.55}{R}
& \nespc{2.96}{0.91}{R}
& \nespc{2.67}{1.01}{R}
& \textit{\nespc{2.12}{0.96}{R}} \\
Synthetic-geometry
& \nespc{1.97}{0.57}{R}
& \nespc{1.71}{0.14}{R}
& \nespc{1.93}{0.59}{R}
& \textbf{\nespc{1.47}{0.31}{R}}
& \nespc{2.81}{1.36}{R}
& \nespc{2.43}{0.29}{R}
& \textit{\nespc{1.57}{0.51}{R}}
& \nespc{1.98}{0.08}{R} \\
\bottomrule
\end{tabular}}
\endgroup
\end{table}

\subsection{Failure prediction and selective classification}\label{app:discrimination}
We assess how well confidence separates correct from incorrect predictions using AUROC~\citep{hendrycks_2017_OOD} and area under the risk--coverage curve (AURC)~\citep{geifman_2019_AURC}. Temperature scaling preserves predicted labels but can change confidence rankings across multiclass examples, so we report both base-model and recalibrated results.

\model attains the best or second-best AUROC (higher is better) and AURC (lower is better) on all CIFAR  settings, winning outright in three, both on the base model and after TS. Relative to CE, base-model AURC decreases by approximately $15.5\%$ on CIFAR-10/ResNet-50 and $13.5\%$ on CIFAR-100/ResNet-50. (Table~\ref{tab:app-discrimination}).

\begin{table}[!htbp]\centering
\caption{Failure-prediction AUROC (higher is better) and AURC ($\times 10^{3}$, lower is better), before and after temperature scaling. Means over seeds; SDs omitted.}
\label{tab:app-discrimination}
\begingroup
\scriptsize
\setlength{\tabcolsep}{1.6pt}
\renewcommand{\arraystretch}{1.03}
\begin{tabular}{@{}llcccccccc@{}}
\toprule
Setting & Metric & CE & FL & LS & MMCE & MDCA & s-AvUC & MbLS & \model \\
\midrule
CIFAR-10 / R50 & AUROC & 0.9357 & 0.9278 & 0.8537 & \textit{0.9359} & 0.9351 & 0.9352 & 0.8995 & \textbf{0.9392} \\
 & AUROC+TS & 0.9352 & 0.9270 & 0.8563 & \textit{0.9361} & 0.9349 & 0.9349 & 0.9001 & \textbf{0.9395} \\
 & AURC & 6.60 & 8.57 & 19.08 & 7.25 & \textit{6.53} & \textit{6.53} & 11.21 & \textbf{5.58} \\
 & AURC+TS & 6.64 & 8.66 & 18.69 & 7.24 & \textit{6.55} & \textit{6.55} & 11.11 & \textbf{5.57} \\
\addlinespace[2pt]
CIFAR-10 / WRN & AUROC & 0.9343 & 0.9290 & 0.8951 & \textbf{0.9393} & 0.9379 & 0.9350 & 0.9340 & \textit{0.9384} \\
 & AUROC+TS & 0.9316 & 0.9295 & 0.8957 & \textbf{0.9391} & 0.9362 & 0.9322 & 0.9331 & \textit{0.9370} \\
 & AURC & \textit{5.18} & 6.58 & 8.89 & 5.64 & 5.51 & 5.46 & 5.69 & \textbf{4.92} \\
 & AURC+TS & \textit{5.41} & 6.53 & 8.83 & 5.64 & 5.67 & 5.73 & 5.77 & \textbf{5.04} \\
\addlinespace[2pt]
CIFAR-100 / R50 & AUROC & 0.8695 & 0.8633 & 0.8594 & \textit{0.8709} & 0.8693 & 0.8697 & 0.8677 & \textbf{0.8764} \\
 & AUROC+TS & 0.8699 & 0.8637 & 0.8601 & \textit{0.8713} & 0.8694 & 0.8700 & 0.8674 & \textbf{0.8763} \\
 & AURC & 69.84 & \textit{69.26} & 73.31 & 70.60 & 71.05 & 69.78 & 69.74 & \textbf{60.41} \\
 & AURC+TS & 69.75 & \textit{69.14} & 72.95 & 70.54 & 71.03 & 69.71 & 69.91 & \textbf{60.44} \\
\addlinespace[2pt]
CIFAR-100 / WRN & AUROC & 0.8771 & 0.8688 & 0.8744 & 0.8761 & 0.8755 & \textit{0.8777} & 0.8747 & \textbf{0.8826} \\
 & AUROC+TS & 0.8773 & 0.8690 & 0.8742 & 0.8763 & 0.8759 & \textit{0.8779} & 0.8741 & \textbf{0.8817} \\
 & AURC & 61.29 & 57.50 & \textbf{52.95} & 64.97 & 62.47 & 62.93 & 60.19 & \textit{54.83} \\
 & AURC+TS & 61.17 & 57.44 & \textbf{53.05} & 64.92 & 62.39 & 62.91 & 60.40 & \textit{55.07} \\
\addlinespace[2pt]
Tiny-ImageNet & AUROC & 0.8594 & 0.8441 & 0.8506 & 0.8567 & \textit{0.8614} & \textbf{0.8644} & 0.8551 & 0.8611 \\
 & AUROC+TS & 0.8587 & 0.8443 & 0.8537 & 0.8557 & 0.8606 & \textbf{0.8637} & 0.8564 & \textit{0.8608} \\
 & AURC & 128.52 & 147.03 & 130.41 & 132.39 & \textit{123.87} & \textbf{122.88} & 125.62 & 126.24 \\
 & AURC+TS & 128.86 & 146.95 & 128.97 & 132.82 & \textit{124.25} & \textbf{123.23} & 124.98 & 126.36 \\
\bottomrule
\end{tabular}
\endgroup
\end{table}

\subsection{Robustness}\label{app:robustness}
Table~\ref{tab:app-robustness} reports corruption robustness and SVHN detection. \model improves corrupted NLL and ECE over CE in four of five image settings and accuracy in all five, attaining the highest corrupted accuracy among the methods in Table~\ref{tab:app-robustness} in four settings. OOD detection results are mixed: AUROC improves over CE on both CIFAR-100 settings and Tiny-ImageNet, but decreases on both CIFAR-10 settings. On Tiny-ImageNet, \model attains the highest reported SVHN AUROC, $0.9221$.

\begin{table}[!htbp]\centering
\caption{Robustness: corruption metrics averaged over five severity levels and 15 corruption types for CIFAR-10/100-C or 14 for Tiny-ImageNet-C; SVHN detection AUROC uses maximum softmax probability.}
\label{tab:app-robustness}
\begingroup
\scriptsize
\setlength{\tabcolsep}{1.6pt}
\renewcommand{\arraystretch}{1.03}
\begin{tabular*}{\linewidth}
  {@{}l@{\extracolsep{\fill}}rrrrrrrr@{}}
\toprule
Setting & CE & FL & LS & MMCE & MDCA & s-AvUC & MbLS & NeSCal \\
\midrule
\multicolumn{9}{@{}l}{\emph{Corrupted NLL}} \\
CIFAR-10 / R50  & 1.2363 & 0.9956 & 1.0565 & \textbf{0.9828} & 1.2799 & 1.2882 & 1.0554 & 1.2483 \\
CIFAR-10 / WRN  & 1.3416 & 1.0069 & 1.0651 & \textbf{0.9919} & 1.3808 & 1.4140 & 1.1209 & 1.2342 \\
CIFAR-100 / R50 & 2.6843 & \textbf{2.4157} & 2.4532 & 2.6950 & 2.7057 & 2.6799 & 2.5766 & 2.4443 \\
CIFAR-100 / WRN & 3.1114 & \textbf{2.5448} & 2.6913 & 2.9448 & 3.1199 & 2.9370 & 2.6783 & 2.5604 \\
Tiny-ImageNet & 4.1629 & 4.0258 & 3.8927 & 4.1066 & 4.1212 & 4.1864 & \textbf{3.8153} & 3.9413 \\
\midrule
\multicolumn{9}{@{}l}{\emph{Corrupted ECE (\%)}} \\
CIFAR-10 / R50  & 15.89 & 8.89 & 13.83 & \textbf{8.80} & 16.25 & 16.45 & 14.50 & 15.99 \\
CIFAR-10 / WRN  & 17.43 & 10.68 & 14.35 & \textbf{9.48} & 17.68 & 18.11 & 15.51 & 16.07 \\
CIFAR-100 / R50 & 21.36 & \textbf{12.83} & 16.48 & 21.44 & 21.53 & 21.18 & 19.79 & 18.58 \\
CIFAR-100 / WRN & 25.46 & \textbf{16.13} & 17.77 & 24.04 & 25.77 & 23.79 & 20.45 & 18.98 \\
Tiny-ImageNet & 22.21 & \textbf{13.88} & 15.24 & 20.41 & 22.13 & 22.83 & 14.08 & 17.17 \\
\midrule
\multicolumn{9}{@{}l}{\emph{Corrupted accuracy (\%)}} \\
CIFAR-10 / R50  & 71.66 & \textbf{72.79} & 72.58 & 70.94 & 71.46 & 71.29 & 72.35 & 72.43 \\
CIFAR-10 / WRN  & 73.85 & 73.70 & 73.05 & 72.24 & 72.14 & 74.20 & 73.11 & \textbf{74.55} \\
CIFAR-100 / R50 & 47.60 & 47.76 & 47.44 & 47.26 & 47.26 & 47.47 & 47.47 & \textbf{49.26} \\
CIFAR-100 / WRN & 47.45 & 47.65 & 47.45 & 47.31 & 47.39 & 48.05 & 47.90 & \textbf{49.27} \\
Tiny-ImageNet & 24.73 & 22.47 & 24.55 & 24.07 & 24.88 & 24.71 & 24.99 & \textbf{25.45} \\
\midrule
\multicolumn{9}{@{}l}{\emph{SVHN OOD AUROC}} \\
CIFAR-10 / R50  & 0.9355 & 0.8996 & 0.8592 & \textbf{0.9462} & 0.9267 & 0.9338 & 0.9064 & 0.9282 \\
CIFAR-10 / WRN  & 0.8996 & 0.8715 & 0.9208 & \textbf{0.9761} & 0.9150 & 0.9190 & 0.9397 & 0.8765 \\
CIFAR-100 / R50 & 0.7463 & \textbf{0.8078} & 0.7621 & 0.7259 & 0.7749 & 0.7357 & 0.7384 & 0.7692 \\
CIFAR-100 / WRN & 0.7340 & 0.7315 & 0.7388 & 0.7385 & 0.7096 & 0.7263 & \textbf{0.7524} & 0.7448 \\
Tiny-ImageNet & 0.9018 & 0.9085 & 0.8463 & 0.8875 & 0.8987 & 0.8967 & 0.8416 & \textbf{0.9221} \\
\bottomrule
\end{tabular*}
\endgroup
\end{table}

\subsection{Ablations}\label{app:ablations}
Component ablations use seeds $18$ and $42$, the cosine vs constant schedule comparison matches the headline in seed numbers, with activation epoch and regularization strength selected separately for each schedule. The Fisher-weighted arm uses $ D_{\mathrm{F}}(p_i,p_j) = \frac{1}{8\ln 2} \sum_k \frac{(p_{i,k}-p_{j,k})^2}{\max(p_{i,k},10^{-2})}$. The floor applies only to the source probabilities. This is a stabilized version of the local quadratic weighting in Proposition~\ref{lem:laplacian}.

\paragraph{Graph construction.} Projection-based graphs retain gains over CE in some of the evaluated settings, without a clear, consistent winner emerging. The choice of feature transformation is therefore not uniformly neutral. Table~\ref{tab:graph-control} displays the random-graph results discussed in Section~\ref{sec:res-design}.

\begin{table}[!htbp]
\centering
\caption{Test NLL for neighborhood controls; lower is better. Random uses headline $\lambda$; Random (scaled) matches penalty magnitude.}
\label{tab:graph-control}
\begingroup
\small
\setlength{\tabcolsep}{4pt}
\begin{tabular*}{\linewidth}{@{\extracolsep{\fill}}lrrrr@{}}
\toprule
Setting & CE & kNN & Random & Random (scaled) \\
\midrule
CIFAR-10 / R50  & 0.2065 & \textbf{0.1843} & 0.2156 & \textit{0.2058} \\
CIFAR-100 / R50 & \textit{0.8782} & \textbf{0.8026} & 1.0386 & 0.8785 \\
Helena         & 2.5651 & \textbf{2.5534} & 2.5890 & \textit{2.5639} \\
\bottomrule
\end{tabular*}
\endgroup
\end{table}

\paragraph{Constant versus cosine regularization.}
With each schedule selected by validation NLL, cosine decay improves NLL, Brier score, and accuracy in all four CIFAR settings; ECE improves in three (Table~\ref{tab:app-schedule}). On the two ResNet-50 settings with graph diagnostics, constant regularization produces lower final edge JS, whereas cosine decay produces lower across-seed error dispersion (Table~\ref{tab:app-theory}). These observations support relaxing the consistency constraint during training: stronger agreement at the selected weights does not necessarily imply better predictive quality. They motivate cosine decay as the default without implying uniform improvement across calibration metrics. 

\begin{table}[t]
\centering
\caption{Constant versus cosine regularization on CIFAR.}
\label{tab:app-schedule}
\begingroup
\scriptsize
\setlength{\tabcolsep}{4pt}
\renewcommand{\arraystretch}{1.05}
\begin{tabular*}{\linewidth}{@{\extracolsep{\fill}}llrrrr@{}}
\toprule
Setting & Schedule & NLL $\downarrow$ & Brier $\downarrow$ & Accuracy $\uparrow$ & ECE $\downarrow$ \\
\midrule
CIFAR-10 / R50 & Constant
& \nescalResultCell{0.1917}{0.0068}
& \nescalResultCell{0.0941}{0.0033}
& \nescalResultCell{0.9358}{0.0026}
& \textbf{\nescalResultCell{0.0155}{0.0012}} \\
& Cosine
& \textbf{\nescalResultCell{0.1843}{0.0058}}
& \textbf{\nescalResultCell{0.0872}{0.0015}}
& \textbf{\nescalResultCell{0.9428}{0.0015}}
& \nescalResultCell{0.0229}{0.0011} \\
\addlinespace[2pt]
CIFAR-10 / WRN & Constant
& \nescalResultCell{0.1966}{0.0092}
& \nescalResultCell{0.0897}{0.0037}
& \nescalResultCell{0.9413}{0.0028}
& \nescalResultCell{0.0252}{0.0021} \\
& Cosine
& \textbf{\nescalResultCell{0.1752}{0.0028}}
& \textbf{\nescalResultCell{0.0785}{0.0018}}
& \textbf{\nescalResultCell{0.9494}{0.0011}}
& \textbf{\nescalResultCell{0.0239}{0.0018}} \\
\addlinespace[2pt]
CIFAR-100 / R50 & Constant
& \nescalResultCell{0.8728}{0.0238}
& \nescalResultCell{0.3423}{0.0062}
& \nescalResultCell{0.7541}{0.0052}
& \nescalResultCell{0.0539}{0.0067} \\
& Cosine
& \textbf{\nescalResultCell{0.8026}{0.0155}}
& \textbf{\nescalResultCell{0.3178}{0.0050}}
& \textbf{\nescalResultCell{0.7719}{0.0052}}
& \textbf{\nescalResultCell{0.0395}{0.0067}} \\
\addlinespace[2pt]
CIFAR-100 / WRN & Constant
& \nescalResultCell{0.8319}{0.0109}
& \nescalResultCell{0.3230}{0.0068}
& \nescalResultCell{0.7713}{0.0076}
& \nescalResultCell{0.0636}{0.0057} \\
& Cosine
& \textbf{\nescalResultCell{0.7760}{0.0173}}
& \textbf{\nescalResultCell{0.3057}{0.0061}}
& \textbf{\nescalResultCell{0.7810}{0.0065}}
& \textbf{\nescalResultCell{0.0460}{0.0062}} \\
\bottomrule
\end{tabular*}
\endgroup
\end{table}

\paragraph{Neighborhood structure.}
At matched initial penalty magnitude, random graphs provide little improvement over CE in the three evaluated settings. Using the headline $\lambda$ produces worse NLL than CE in all three (Table~\ref{tab:app-ablations}), supporting the importance of which samples the graph connects.

\paragraph{Divergence sensitivity.} Divergence arms are at matched initial penalty magnitude, with no divergence uniformly dominating among the stable, scale-matched runs. The common penalty target over-regularizes some settings, so cells marked $\dagger$ and divergent runs are excluded from the comparison of divergence quality.

\begin{table}[!htbp]
\centering
\caption{Ablations: test NLL, lower is better. -- : not run. $^{\dagger}$: the common penalty target over-regularizes this setting; div.: diverged. }
\label{tab:app-ablations}
\begingroup\scriptsize\setlength{\tabcolsep}{3pt}
\resizebox{\linewidth}{!}{%
\begin{tabular}{@{}lcccccccc@{}}
\toprule
Arm & C10/R50 & C10/WRN & C100/R50 & C100/WRN & Helena & Jannis & Covertype & 20NG \\
\midrule
Headline ($k{=}10$, JS, raw features) & 0.1843 & 0.1752 & 0.8026 & 0.7760 & 2.5534 & 0.7066 & 0.2499 & 0.8023 \\
Cross-entropy (no regularizer)        & 0.2065 & 0.1972 & 0.8782 & 0.8785 & 2.5651 & 0.7125 & 0.2516 & 0.8127 \\
\addlinespace[2pt]
Random graph, same $\lambda$          & 0.2156 & --     & 1.0386 & --     & 2.5890 & --     & --     & --     \\
Random graph, matched penalty         & 0.2058 & --     & 0.8785 & --     & 2.5639 & --     & --     & --     \\
\addlinespace[2pt]
$k=2$                                 & 0.1830 & 0.1736 & 0.8368 & 0.7936 & 2.5554 & 0.7069 & 0.2487 & 0.7466 \\
$k=5$                                 & 0.1789 & 0.1702 & 0.8148 & 0.7985 & 2.5553 & 0.7072 & 0.2531 & 0.7650 \\
$k=20$                                & 0.1777 & 0.1765 & 0.8106 & 0.7993 & 2.5561 & 0.7072 & 0.2514 & 0.7726 \\
$k=50$                                & 0.1781 & 0.1754 & 0.8204 & 0.7974 & 2.5573 & 0.7079 & 0.2551 & 0.7795 \\
\addlinespace[2pt]
Graph on PCA-50                       & 0.1777 & 0.1774 & 0.8355 & 0.8440 & 2.5549 & 0.7053 & 0.2505 & 0.7608 \\
Graph on PCA-16                       & 0.1822 & 0.1702 & 0.8581 & 0.8354 & 2.5553 & 0.7051 & 0.2516 & 0.7580 \\
Graph on UMAP-30                      & 0.1801 & 0.1748 & 0.8291 & 0.7758 & 2.5535 & 0.7050 & 0.2498 & 0.7586 \\
\addlinespace[2pt]
Divergence: JS                        & 1.1048$^{\dagger}$ & --     & 0.8167 & 0.7642 & 2.5604 & 0.7128 & 0.3903$^{\dagger}$ & 0.8382 \\
Divergence: squared $L_2$             & 0.2548$^{\dagger}$ & --     & 0.8084 & 0.7498 & 2.6115 & 0.7182 & 0.3905$^{\dagger}$ & 0.8263 \\
Divergence: KL                        & 0.2749$^{\dagger}$ & --     & 0.8383 & 0.8154 & 2.5806 & 0.7145 & 0.4239$^{\dagger}$ & 0.8869 \\
Divergence: Fisher-weighted           & div.   & div.   & 0.8754 & 0.9818 & 2.6091 & 0.7057 & 0.3933$^{\dagger}$ & 0.7666 \\
\bottomrule
\end{tabular}}
\endgroup
\end{table}
 
\subsection{Activation rules}\label{app:activation}
If the activation epoch could be placed from a quantity observable during the warm-up, \model would lose one swept hyperparameter. We tested three such rules on the cross-entropy trajectory:
\begin{itemize}
  \item \emph{entropy drop}: mean predictive entropy falls below a fraction of its initial value;
  \item \emph{CWCC}: class-weighted cross-class connectivity falls below a fraction of its initial value;
  \item \emph{probe}: kNN-probe accuracy reaches a fraction of its maximum over the warm-up.
\end{itemize}
For these rule experiments, diagnostics are logged every $10$ epochs and activation gates are rounded down to a separate $20$-epoch checkpoint grid. Table~\ref{tab:app-activation} reports each rule's regret, the test-NLL difference against the grid-selected onset at the same $\lambda$.

No rule transfers consistently across settings. Probe-rule regret ranges from $-0.039$ to $+0.072$ nats. Because this rule uses the maximum probe accuracy over the warm-up trajectory, it is a retrospective diagnostic rather than an online activation rule.

\begin{table}[!htbp]\centering
\caption{Regret of activation rules: test NLL relative to the grid-selected onset at the same $\lambda$; positive is
worse. --: rule not evaluated.}
\label{tab:app-activation}
\begingroup\scriptsize\setlength{\tabcolsep}{4pt}
\begin{tabular}{@{}lccc@{}}
\toprule
Setting & Entropy drop & CWCC & Probe (hindsight) \\
\midrule
CIFAR-10 / R50  & $+0.0222$ & $-0.0018$ & $+0.0178$ \\
CIFAR-10 / WRN  & $-0.0012$ & $+0.0298$ & $+0.0170$ \\
CIFAR-100 / R50 & $+0.0940$ & $+0.2987$ & $+0.0123$ \\
CIFAR-100 / WRN & $+0.0439$ & $-0.0427$ & $-0.0388$ \\
Tiny-ImageNet & $+0.0686$ & -- & $+0.0718$ \\
Helena          & --        & --        & $+0.0075$ \\
Jannis          & $+0.4471$ & --        & $+0.0005$ \\
Covertype       & --        & --        & $+0.0000$ \\
20NG   & $+0.0393$ & $+0.0393$ & $-0.0059$ \\
\bottomrule
\end{tabular}
\endgroup
\end{table}
 
\subsection{Selection criterion}\label{app:selection}
Every train-time method is selected by validation NLL across checkpoints and hyperparameters. Selecting by ECE can favor less informative predictions. Across the eleven settings, ECE-based selection improves \modelNoSpace's test ECE by $0.85$ percentage points on average but worsens NLL by $0.017$ nats. On CIFAR-100/R50, for example, ECE improves from $3.95\%$ to $2.47\%$ while NLL worsens from $0.8026$ to $0.871$. Similar trade-offs occur among the baselines. These results support the shared proper-score criterion while confirming that it does not optimize each method's ECE.

\subsection{Training cost}\label{app:cost}
Table~\ref{tab:app-cost} reports training-loop time, one-off graph-construction time, and peak GPU memory. Peak GPU memory is the maximum allocated at any point in the run, including the one-off graph construction. Where \model's peak clearly exceeds CE's (Tiny-ImageNet, CIFAR-100/WRN and the tabular settings), the maximum occurs during the chunked similarity computation of graph construction; differences of a few hundredths of a GB are within run-to-run allocator variation. Relative overhead is small on the image models and larger on the smaller tabular and text models, where the absolute increase in the reported training time is at most $27.4$ seconds. Peak memory increases vary across settings. Inference is unchanged.

\begin{table}[!htbp]\centering
\caption{Training cost: wall-clock training time (s), relative overhead, one-off graph construction (s), and peak GPU memory (GB). Means follow headline seeds except CIFAR-10/ResNet-50 \model (one seed).}
\label{tab:app-cost}
\begingroup\scriptsize\setlength{\tabcolsep}{4pt}
\begin{tabular}{@{}lrrrrrr@{}}
\toprule
Setting & CE time & \model time & Overhead & Graph build & CE memory & \model memory \\
\midrule
CIFAR-10 / R50    & 3430 & 3456 & $+0.7\%$  & 20.4 & 2.05 & 2.05 \\
CIFAR-10 / WRN    & 5028 & 5053 & $+0.5\%$  & 26.9 & 2.29 & 2.26 \\
CIFAR-100 / R50   & 2702 & 2725 & $+0.9\%$  & 17.3 & 2.11 & 2.15 \\
CIFAR-100 / WRN   & 5068 & 5094 & $+0.5\%$  & 27.1 & 2.59 & 3.05 \\
Tiny-ImageNet & 11036 & 11085 & $+0.4\%$ & 146.2 & 3.89 & 6.89 \\
\addlinespace[2pt]
Helena            & 36.5 & 50.1 & $+37\%$   & 1.49 & 0.04 & 1.44 \\
Jannis            & 148.9 & 168.7 & $+13\%$ & 2.72 & 0.03 & 0.24 \\
Covertype         & 71.6 & 99.0 & $+38\%$   & 2.42 & 0.03 & 0.29 \\
\addlinespace[2pt]
20NG     & 61.0 & 66.9 & $+10\%$   & 1.65 & 0.61 & 0.62 \\
Amazon (5-class)  & 3.37 & 4.63 & $+38\%$   & 0.56 & 0.02 & 0.14 \\
Amazon (binary)   & 1.38 & 2.73 & $+97\%$   & 0.39 & 0.02 & 0.08 \\
\bottomrule
\end{tabular}
\endgroup
\end{table}

\subsection{Theory diagnostics}\label{app:theory-audit}
We evaluate bound tightness and related diagnostics on the frozen training graphs for CIFAR-10 and CIFAR-100 with ResNet-50, using CE and both \model schedules over four seeds (Table~\ref{tab:app-theory}).

All evaluated bounds hold on every edge in all evaluated configurations. As expected of worst-case inequalities, they are conservative: median confidence-bound tightness is $0.02$--$0.04$ on CIFAR-10 and $0.10$--$0.13$ on CIFAR-100, where neighboring disagreement is larger. We therefore use them to identify which quantities govern calibration propagation rather than to predict its magnitude. The Markov bound of Proposition~\ref{lem:markov-bound} is informative when the penalty is small relative to the threshold: on CIFAR-10 it bounds the fraction of edges with $\mathrm{JS}\ge 0.25$ by $0.52$, against an observed $0.15$; on CIFAR-100, where mean disagreement is higher, it is not informative at this threshold.

The diagnostics bearing on Proposition~\ref{prop:threshold} are consistent with its mechanism. Cosine-scheduled \model reduces the across-seed dispersion of neighboring errors and of individual confidences on both settings, and both schedules reduce the NLL gap to a test-fitted temperature. Final edge agreement behaves differently across schedules: the constant schedule lowers mean edge JS relative to CE on both settings, whereas under cosine decay $\lambda_t$ approaches zero by the end of training, and edge JS at the selected checkpoint is close to CE's. Better test NLL therefore need not coincide with the smallest final penalty, consistent with the regularizer acting on the dispersion of neighboring errors rather than on end-state agreement alone. Finally, the quadratic expansion of Corollary~\ref{lem:js-laplacian} describes the geometry near agreement; at the trained weights, neighboring predictions lie outside this regime on nearly all edges, so we treat the Fisher-type weighting as an interpretation of the penalty's local geometry rather than a description of training, in line with the ablations, where Fisher weighting shows no consistent advantage.

\begin{table}[!htbp]\centering
\caption{Theory diagnostics at the selected weights on the frozen training graph. Tightness concerns the observable predictive and confidence bounds. The dispersion proxy corresponds to the equal-bias, equal-variance specialization of Proposition~\ref{prop:threshold}, not the full expected squared error difference.}
\label{tab:app-theory}
\begingroup\scriptsize\setlength{\tabcolsep}{4pt}
\begin{tabular}{@{}lcccccc@{}}
\toprule
& \multicolumn{3}{c}{CIFAR-10 / R50} & \multicolumn{3}{c}{CIFAR-100 / R50} \\
\cmidrule(lr){2-4}\cmidrule(lr){5-7}
Quantity & CE & \model (const) & \model (cos) & CE & \model (const) & \model (cos) \\
\midrule
Bound violations (L1, confidence) & 0 & 0 & 0 & 0 & 0 & 0 \\
Confidence-bound tightness, median & 0.019 & 0.038 & 0.019 & 0.131 & 0.117 & 0.101 \\
Edges within the expansion's regime & $0\%$ & $0\%$ & $0\%$ & $0.005\%$ & $0.005\%$ & $0.005\%$ \\
Across-seed dispersion proxy $\sigma\sqrt{2(1-\rho)}$ & 0.055 & 0.061 & 0.038 & 0.102 & 0.102 & 0.088 \\
Across-seed confidence spread $\sigma_c$ & 0.059 & 0.059 & 0.035 & 0.103 & 0.112 & 0.076 \\
NLL gap to the test-fitted oracle temperature & 0.020 & 0.005 & 0.015 & 0.030 & 0.025 & 0.008 \\
Mean edge JS at the selected weights & 0.134 & 0.102 & 0.129 & 0.317 & 0.307 & 0.323 \\
Reciprocal edges & -- & 0.435 & 0.435 & -- & 0.439 & 0.439 \\
\bottomrule
\end{tabular}
\endgroup
\end{table}

\paragraph{Synthetic setup.} Each scenario draws equiprobable Gaussian classes in $\mathbb{R}^{20}$ with identity covariance, whose centers differ only in the first two coordinates. In the label-noise scenario, the centers lie on a circle of radius $8$. In the geometric scenario, half lie on that circle and half on a circle of radius $2.8$ centered at $(12.8,0)$, so the second group's classes are more closely spaced. In both scenarios, the labels of half the classes are redrawn uniformly at random with probability $H\in\{0,0.2,0.4,0.6\}$. Data are split $60/20/20$ into training, validation and test sets, and the model is an MLP with two hidden layers of width $64$ and a $16$-dimensional representation. \model uses $k=10$ and constant regularization, with activation epoch and strength selected by validation NLL and held fixed across the heterogeneity sweep.

Evaluation compares predictions with pointwise reference distributions $q_i$ and reports the top-label reference error $\frac{1}{n}\sum_i\bigl(c_i-q_{i,\hat y_i}\bigr)^2$, the mean squared pointwise calibration error of Section~\ref{sec:background} with $a_i=q_{i,\hat y_i}$. In the label-noise scenario, $q_i$ is the analytical noisy-label posterior. In the geometric scenario, $q_i$ mixes the clean posterior with the uniform distribution at weight $H$ when sample $i$'s generating class is noisy, and equals the clean posterior otherwise. This coincides with the noisy-label posterior wherever the class posterior is concentrated within either the noisy or the clean group, and differs from it only where classes from the two groups overlap. The comparator labeled ``oracle'' chooses, on the test set, the temperature minimizing this error among $47$ candidates (six equally spaced in $[0.5,1)$ and $41$ in $[1,5]$); it is a test-informed diagnostic rather than a deployable calibrator.

\begin{figure}[!htbp]\centering
\includegraphics[width=\linewidth]{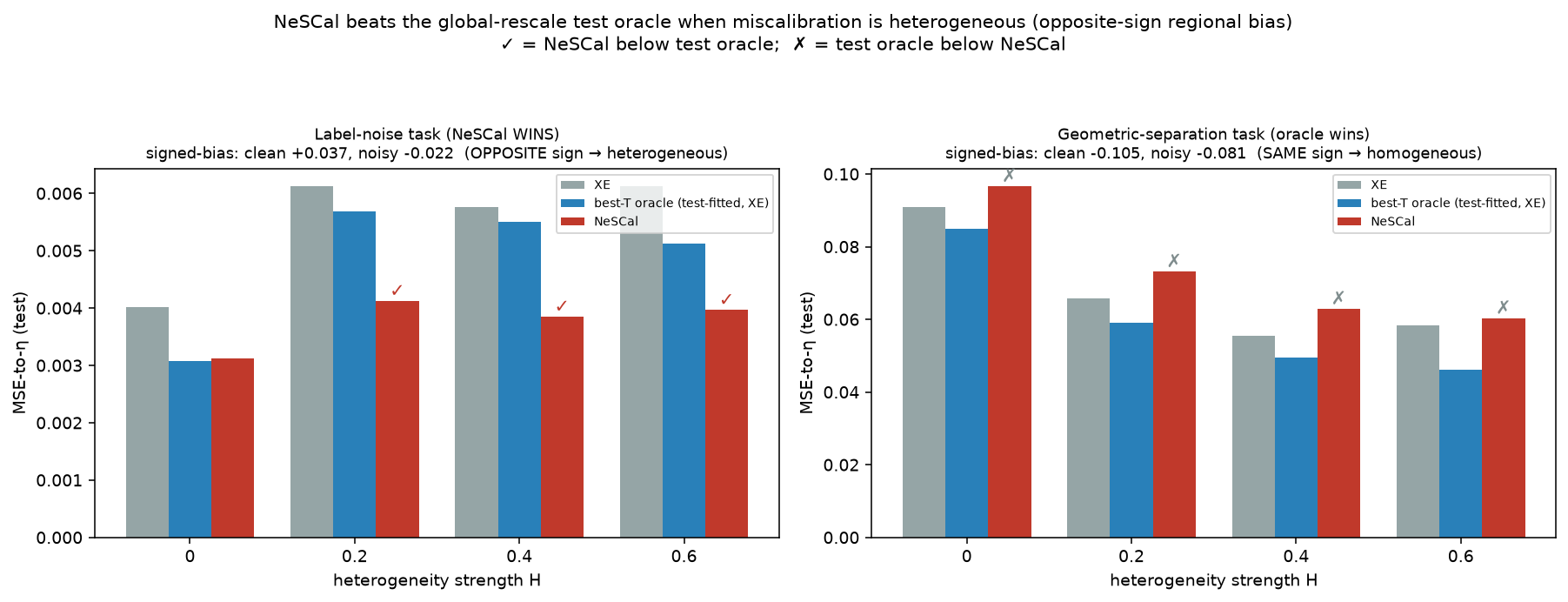}
\caption{Mean squared top-label reference error across heterogeneity strengths $H$, per scenario, on CE, CE with the test-informed temperature (\emph{oracle}), and \modelNoSpace.}
\label{fig:synthetic-oracle}
\end{figure}

\paragraph{The synthetic regimes.} Figure~\ref{fig:synthetic-oracle} compares \model with CE and the test-informed temperature search. In the label-noise scenario, the regional confidence biases have opposite signs, which no single temperature can correct together, and \model attains lower top-label reference error than the searched temperatures at every evaluated $H>0$. In the geometric scenario, the biases share a sign: the temperature search is best, and \model has higher reference error than CE. Both outcomes are consistent with the idealized analysis of Proposition~\ref{prop:threshold}. Averaging an edge changes its summed expected squared error by $\tfrac12(a_i-a_j)^2-\tfrac12\mathbb{E}[(e_i-e_j)^2]$ (Appendix~\ref{app:gs:threshold}), so smoothing helps only when neighboring errors differ more than neighboring correctness probabilities do. When neighbors share their error, as under a common same-sign bias, the second term is small, and any correctness gap, which the more closely spaced classes of the geometric scenario plausibly create, makes smoothing increase error. A global temperature acts on the shared component and is therefore the appropriate correction. The geometric scenario therefore exhibits the failure mode the analysis predicts, and the label-noise scenario the regime in which smoothing and rescaling are not interchangeable.

\subsection{Perturbation studies}\label{app:perturb}
These studies test neighborhood smoothing under class imbalance, which can make neighborhoods less representative for rare classes and label noise, which weakens the supervised anchor. Both studies use CIFAR-100 with the clean-selected configurations of R50 and WRN (Appendix~\ref{app:implementation:robust}).

\paragraph{Label noise.} \model lowers ECE relative to CE in all four architecture--noise conditions, but improves NLL and accuracy in only one (Table~\ref{tab:app-noise}). These results show that improved aggregate confidence calibration under label noise need not accompany improved predictive quality.

\begin{table}[!htbp]\centering
\caption{Symmetric label noise on CIFAR-100, without re-tuning $\lambda$.}
\label{tab:app-noise}
\begingroup\scriptsize\setlength{\tabcolsep}{4pt}
\begin{tabular}{@{}llcccccc@{}}
\toprule
& & \multicolumn{2}{c}{NLL $\downarrow$} & \multicolumn{2}{c}{ECE (\%) $\downarrow$} & \multicolumn{2}{c}{Accuracy (\%) $\uparrow$} \\
\cmidrule(lr){3-4}\cmidrule(lr){5-6}\cmidrule(lr){7-8}
Architecture & Noise & CE & \model & CE & \model & CE & \model \\
\midrule
ResNet-50 & $20\%$ & \textbf{1.2521} & 1.3123 & 6.42 & \textbf{4.32} & \textbf{68.20} & 67.64 \\
          & $40\%$ & \textbf{1.7125} & 1.9218 & 15.96 & \textbf{9.66} & \textbf{61.01} & 56.95 \\
WRN-28-10 & $20\%$ & 1.1743 & \textbf{1.1528} & 2.92 & \textbf{1.70} & 70.19 & \textbf{70.44} \\
          & $40\%$ & \textbf{1.5154} & 1.6212 & 9.58 & \textbf{4.85} & \textbf{64.35} & 61.74 \\
\bottomrule
\end{tabular}
\endgroup
\end{table}

\paragraph{Class imbalance.} Under long-tailed training, the JS penalty improves NLL over CE in three of six settings. Squared $L_2$ attains the lowest NLL in all six. Because these arms use equal $\lambda$ rather than matched penalty magnitude, the effect of divergence weighting (Table~\ref{tab:app-perturb}) is not isolated.

\begin{table}[!htbp]\centering
\caption{Long-tailed CIFAR-100: test NLL for CE and three divergences at equal $\lambda$.}
\label{tab:app-perturb}
\begingroup\scriptsize\setlength{\tabcolsep}{4pt}
\begin{tabular}{@{}llcccc@{}}
\toprule
Architecture & Imbalance & Cross-entropy & JS & Fisher-weighted & Uniform ($L_2$) \\
\midrule
ResNet-50  & $10$  & 1.8104 & 1.8018 & 1.8403 & \textbf{1.7629} \\
           & $50$  & 2.9898 & 3.0154 & 3.0717 & \textbf{2.8144} \\
           & $100$ & 3.4013 & 3.4364 & 3.5764 & \textbf{3.2381} \\
WRN-28-10  & $10$  & 1.7965 & 1.6838 & 1.6325 & \textbf{1.5680} \\
           & $50$  & 2.5238 & 2.4804 & 2.6973 & \textbf{2.3800} \\
           & $100$ & 2.9397 & 2.9967 & 3.2001 & \textbf{2.8844} \\
\bottomrule
\end{tabular}
\endgroup
\end{table}

\end{document}